\documentclass[a4paper,fleqn]{cas-dc}

\usepackage[authoryear,longnamesfirst]{natbib}

\usepackage{graphicx}%
\usepackage{multirow}%
\usepackage{amssymb,amsfonts}%
\usepackage{amsmath}  % Enable left-aligned equations
\usepackage{mathrsfs}%
\usepackage[title]{appendix}%
\usepackage{xcolor}%
\usepackage{manyfoot}%
\usepackage{booktabs}%
\usepackage{algorithm}%
\usepackage{algorithmicx}%
\usepackage{algpseudocode}%
\usepackage{listings}%

\newtheorem{definition}{Definition}%

\DeclareMathOperator*{\argmax}{arg\,max}

\begin{document}
\let\WriteBookmarks\relax
\def\floatpagepagefraction{1}
\def\textpagefraction{.001}

%\shorttitle{Towards Sample-Efficiency and Generalization of T-IRL: A Comprehensive Literature Review}

\shortauthors{Hassani et~al.}

\title [mode = title]{Towards Sample-Efficiency and Generalization of Transfer and Inverse Reinforcement Learning: A Comprehensive Literature Review}                      
%\tnotemark[1,2]

%\tnotetext[1]{This document is the results of the research
%   project funded by the National Science Foundation.}

%\tnotetext[2]{The second title footnote which is a longer text matter
%   to fill through the whole text width and overflow into
%   another line in the footnotes area of the first page.}

\author[1]{Hossein Hassani}[orcid=https://orcid.org/0000-0003-3979-9166]

\cormark[1]

\ead{hhassa52@uwo.ca}

\author[2]{Ehsan Hallaji}
\ead{hallaji@uwindsor.ca}
\author[3]{Roozbeh Razavi-Far}
\ead{roozbeh.razavi-far@unb.ca}

\author[2]{Mehrdad Saif}
\ead{msaif@uwindsor.ca}

\author[4]{Liang Lin}
\ead{linliang@ieee.org}

\affiliation[1]{organization={Department of Electrical and Computer Engineering, Western University},
    addressline={}, 
    city={London},
    postcode={ON N6A 3K7}, 
    country={Canada}}
    
\affiliation[2]{organization={Department of Electrical and Computer Engineering, University of Windsor},
    addressline={}, 
    city={Windsor},
    postcode={ON N9B 3P4}, 
    country={Canada}}

\affiliation[3]{organization={Faculty of Computer Science and Canadian Institute for Cybersecurity, University of New Brunswick},
    addressline={}, 
    city={Fredericton},
    postcode={NB E3B 5A3}, 
    country={Canada}}
    
\affiliation[4]{organization={School of Data and Computer Science, Sun Yat-sen University},
    addressline={}, 
    city={Guangzhou},
    postcode={510006}, 
    country={China}}

\cortext[cor1]{Corresponding author}

\begin{abstract}
Reinforcement learning (RL) is a sub-domain of machine learning, mainly concerned with solving sequential decision-making problems by a learning agent that interacts with the decision environment to improve its behavior through the reward it receives from the environment. This learning paradigm is, however, well-known for being time-consuming due to the necessity of collecting a large amount of data, making RL suffer from sample inefficiency and difficult generalization. Furthermore, the construction of an explicit reward function that accounts for the trade-off between multiple desiderata of a decision problem is often a laborious task. These challenges have been recently addressed utilizing transfer and inverse reinforcement learning (T-IRL). In this regard, this paper is devoted to a comprehensive review of realizing the sample efficiency and generalization of RL algorithms through T-IRL. Following a brief introduction to RL, the fundamental T-IRL methods are presented and the most recent advancements in each research field have been extensively reviewed. Our findings denote that a majority of recent research works have dealt with the aforementioned challenges by utilizing human-in-the-loop and sim-to-real strategies for the efficient transfer of knowledge from source domains to the target domain under the transfer learning scheme. Under the IRL structure, training schemes that require a low number of experience transitions and extension of such frameworks to multi-agent and multi-intention problems have been the priority of researchers in recent years.
\end{abstract}

% \includegraphics{figs/grabs.pdf}

%\item Research highlights item 1
%\item Research highlights item 2
%\item Research highlights item 3

% Keywords

\maketitle

\section{Introduction}
\label{sec1}
The ability to self-learn along with the expeditious advancements in computer hardware and data repository has made artificial intelligent (AI) become the foremost solution to computer vision, natural language processing (NLP), knowledge processing, and planning problems \cite{NIAN2020106886, qiang2023natural}. AI encompasses a wide range of concepts to address the aforementioned problems. Among them, machine learning (ML) could be referred to as the most influential field \cite{russell2010artificial, tiddi2022knowledge}, aimed at studying and developing learning algorithms and statistical models to predict outcomes without having to be explicitly programmed. 

ML can be decomposed into three fields including supervised learning (SL), unsupervised learning (UL), and reinforcement learning (RL). SL relies on a training data set that encompasses explicit examples of what a correct output should be for a given input. Outputs are generally provided by subject matter experts, where SL algorithms attempt to generalize through the labeled data examples to construct a model. The constructed models are ultimately used to predict the label of unseen data examples \cite{sun2022low, su2021enhanced}. UL, however, is not built based on a data set with explicit outputs and data examples in a training set do not include the correct outputs. The general aim of UL is then to find hidden patterns and structures within a given data set that has not been categorized or labeled. In this regard, UL has found a wide range of applications in dimensionality reduction \cite{hassani2021unsupervised}, feature extraction \cite{liu2019flexible}, and clustering \cite{zhou2022clustering}. By combining the ideas of SL and UL, the concept of semi-supervised learning (SSL) has emerged, for which the training of the predictive model is enabled by a few labeled data examples, while the remaining data examples are unlabeled \cite{bahrami2021joint}. In contrast to SL, UL, and SSL, the aim of RL is neither finding the hidden structures of data nor labeling unlabeled data, but it is to learn the best sequence of actions leading to a desired output \cite{hassani2022real}. Indeed, RL is learning how to map situations to actions to maximize a numerical reward. This mapping is called policy in the RL nomenclature, which is a function of state observations. The parameters of this function are adjustable, and learning in RL refers to systematically adjusting these parameters. This is where RL algorithms emerge. RL algorithms and policy are augmented into a learner which is called the agent \cite{candela2023risk}. The agent aims to learn the optimal policy, meaning discovering the actions that yield to the most reward through a trial-and-error search of the action space. The learning process is enabled through data from a dynamic environment, in which the current actions of the agent might affect the subsequent states of the environment and all the subsequent rewards, known as the delayed reward. Trial-and-error search and delayed reward are two of the characteristics that distinguish RL from SL and UL \cite{sutton2018reinforcement,9928222}. 

One way to realize RL is through Markov decision processes (MDPs), which are known as the classical formulation of sequential decision-making \cite{hasanbeig2023certified}. The theoretical statements of RL have been developed using MDPs, which are ideal forms of RL problems \cite{9404328}. Moreover, MDPs can be used to formulate many real-world problems, meaning that RL can be used to solve them as well. RL has thus gained the attention of many researchers and has found real-world applications in many fields. These applications include but are not limited to transportation (autonomous vehicles, adaptive signal traffic), health care (dynamic treatment regimes), education (recommendation, autonomous teacher), finance (pricing, trading, risk management), energy (control of smart grids, adaptive decision control, diagnostics), NLP (translation, chatbot), computer vision (image/video recognition), robotics (navigation, mapping, localization, control), and games \cite{du2021survey,9537641,7902130, gu2023safe}.

Training of an RL agent, however, involves a trade-off problem, known as the exploration-exploitation dilemma \cite{sutton2018reinforcement}. This dilemma originates from the trial-and-error interactions between the agent and environment that conducts the fact that learning is online in RL. In other words, the agent's action determines the information to be returned from the environment and the agent's choices determine the data to learn from. The exploration-exploitation dilemma states that there should be a trade-off between exploiting actions that collect the most reward from the environment and the agent is already aware of them, or exploring other parts of the action space that are unknown to the agent \cite{9679819}. Neither pure exploitation nor pure exploration is practical. The former makes the agent ignore other states of the environment and the information of states beyond the low-rewarded areas that consequently increases the required time for the agent to learn an optimal policy, and even, the agent might end up learning a sub-optimal policy. The latter is also not a good approach for training on physical hardware and is not efficient because it is more likely for the agent to spend much time covering a big portion of the state space. This dilemma is therefore a critical concept in the training of RL agents \cite{zhu2020transfer}. However, because the environment dynamics are not usually available, the agent requires to collect sufficient interaction experiences to learn the policy. This can be restrictive, especially in practical problems, where the observations might not be complete, the reward might be sparse, and the agent deals with complex state and action spaces. To tackle such issues, the concept of transfer reinforcement learning  (TRL) has been widely employed, where the ultimate goal is to guide the agent with external expertise to efficiently learn a target task. 

Even though RL has shown encouraging results in numerous applications, however, it is often a laborious task to end up with an explicit reward function that meaningfully accounts for the trade-off between different desiderata of a given problem \cite{9537731}. For a self-driving car, for example, there should be a trade-off between different features such as lane preference, distance to other cars, speed, lane changes, and so on. This requires a designer to explicitly assign proper weight to each feature in the construction of the reward by considering their trade-off. This adjustment can either be time-consuming or might not lead to the best reward function for the efficient training of an RL agent. Besides, the design of the reward function is crucial in advantage-based RL algorithms to deal with the scalarization issue as it directly influences the estimation of the advantage function, which guides the agent's decision-making process \cite{galatolo2021solving}. A well-designed reward function can provide informative feedback to the agent, facilitating more efficient learning and better performance in complex environments. Conversely, a poorly designed reward function can lead to challenges such as sparse rewards, scaling issues, or non-stationarity, which can prevent the accurate estimation of advantages and hinder the agent's learning progress. This issue has been well-addressed by employing inverse reinforcement learning (IRL) algorithms that benefit from expert demonstrations to implicitly encode the reward function \cite{olson2021counterfactual}. IRL could generally be thought of as a technique for estimating the reward function through available demonstrations of the optimal behavior \cite{9715172}. In recent years, there has been a substantial interest in IRL due to its significant features such as removing the need for manual characterization of reward function which broadens the applicability of RL. Moreover, IRL benefits from a more generalization capability in dealing with environment changes (e.g., noisy observations) since the reward function is more transferable compared with the learned policy by the agent \cite{ARORA2021103500}.  

Transfer and inverse RL (T-IRL) has gained much attention and a notable number of research studies have been conducted recently. This work is devoted to the study of fundamental T-IRL techniques along with a comprehensive review of the most recent advancements in these research fields. In contrast to the recent surveys on T-IRL \cite{zhu2020transfer,ARORA2021103500}, this survey not only reviews the fundamental models of T-IRL but also it is mainly focused on the developed strategies to deal with existing problems in RL, i.e., sample efficiency and generalization, under the T-IRL framework. 

The rest of this paper is organized as follows. In Section \ref{sec2}, a brief introduction is given to the RL algorithms and fundamental TRL and IRL models are studied. Section \ref{sec3} reviews the most recent advancements in TRL, which is then followed by a review of IRL techniques in Section \ref{sec4}. Applications, challenges, and open problems are discussed in Section \ref{sec5} and Section \ref{sec6} concludes the paper. 

\section{Background}\label{sec2}
This section gives a brief introduction to RL and its formulation along with studying the fundamental T-IRL models. It is worth noting that Section \ref{secTRL} delves into the foundational TRL models, while Section \ref{sec3} provides a comprehensive examination of the prevalent trends in this domain. Similarly, Section \ref{secIRL} offers a concise overview of basic IRL models, followed by an in-depth exploration of contemporary techniques in IRL in Section \ref{sec4}.

\begin{figure}
\centering
\includegraphics[width=0.5\columnwidth]{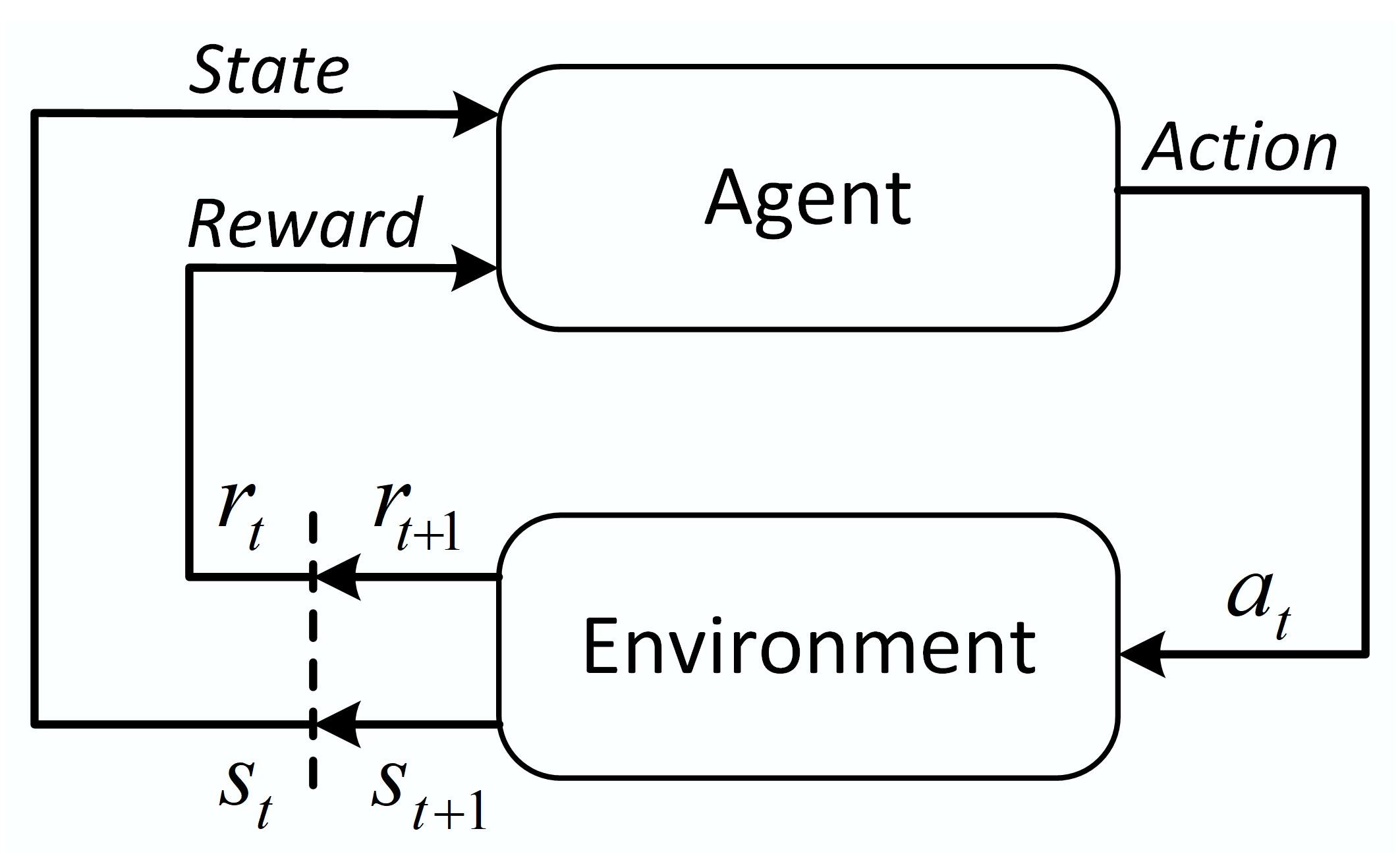}
\caption{Interactions between an RL agent and its environment.}\label{fig1}
\end{figure}

\subsection{Reinforcement Learning}
RL refers to the problems, in which an agent is assigned to a sequential decision-making task. Such problems can be classically formulated using MDPs that are usually defined utilizing a tuple shown by $\langle S, A, T, R, \gamma\rangle$, where $S$ denotes the state-space and $A$ stands for the set of possible actions that agent can choose from. In addition, $T:S\times A\rightarrow P(s'|s,a)$ is a state-transition probability function and defines the probability, by which the environment transits from state $s$ to state $s'$ when action $a$ is executed. Furthermore, $R:S\times A\rightarrow \mathbb{R}$ is a reward function and $r(s,a)$ is the received reward by executing action $a$ in state $s$. $\gamma\in[0,1]$ is called the discount factor (or the discount rate) and it is associated with time horizons in the construction of the discounted return $G_t$ as given below:
\begin{align}
G_t\doteq r_{t+1}+\gamma r_{t+2}+\gamma^2r_{t+3}+\ldots=\sum_{k=1}^{\infty} \gamma^kr_{t+k+1}.
\end{align}
The agent interacts with the environment at each time step $t$ ($t=0,1,\ldots$) in a way that it firstly observes the current state of the environment $s_t$, and, then, decides to take an action $a_t$ based on its observation. The agent receives the consequence of its action through a reward of value $r_{t+1}$ and finds itself in the next state of the environment $s_{t+1}$. This interaction is illustrated in Fig. \ref{fig1}. Through this interaction, the behavior of the agent is represented by the policy $\pi(a|s)$, denoting the probability by which the agent selects action $A_t=a$ given the state $S_t=s$. Indeed, the policy is a mapping from states to probabilities of selecting each possible action. The goodness of a policy is evaluated by the expected return, which can be estimated by either a state-value function or an action-value function. The former can be written as follows:
\begin{align}
V_{\pi}(s) & \doteq \mathbb{E}_{\pi}\Big[G_t|S_t=s\Big] \nonumber\\
& =\sum_a\pi(a|s)\sum_{s',r}P(s',r|s,a)\Big[r(s,a)+\gamma V_{\pi}(s')\Big],\nonumber
\end{align}
and it is defined as the value function of state $s$ under policy $\pi$. The latter shows the values of taking an action $a$ in a state $s$ following a policy $\pi$ and is defined as follows:
\begin{align}
Q_{\pi} & (s,a) \doteq \mathbb{E}_{\pi}\left[G_t|S_t=s,A_t=a\right]\nonumber\\
& =\sum_{s'}P(s'|s,a)\Big[r(s,a,s')+\gamma\sum_{a}\pi(a'|s')Q_{\pi}(s',a')\Big].\nonumber
\end{align}

\begin{figure}
\centering
\includegraphics[width=0.85\columnwidth]{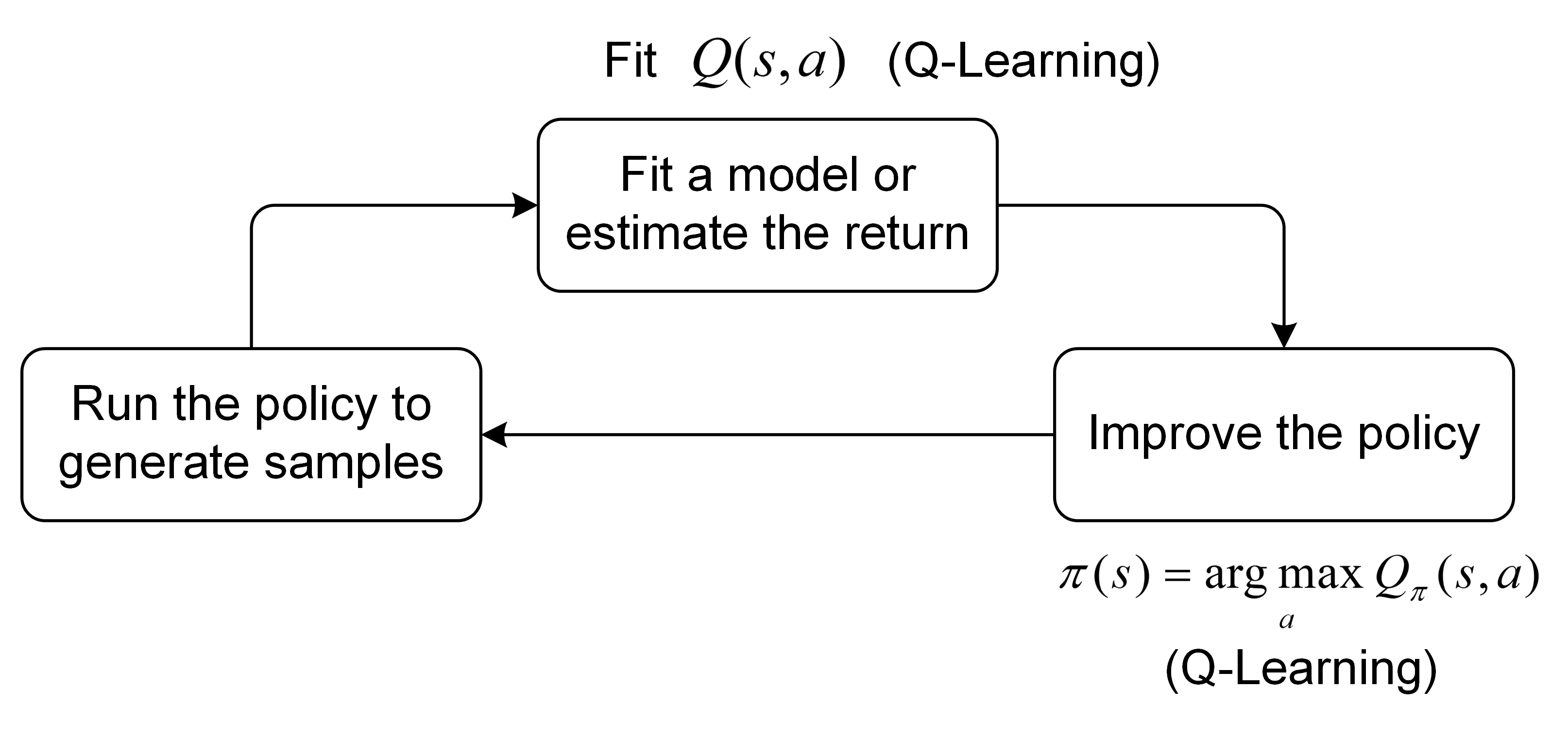}
\caption{The anatomy of an RL algorithm.}\label{fig2}
\end{figure}

The ultimate goal of the agent is to find an optimal policy $\pi_*$, for which the value of a state must equal the expected return for the best action from a given state (the Bellman optimality equation) \cite{sutton2018reinforcement}:
\begin{align}
V_*(s) & \doteq \max_{\pi}V_{\pi}(s) =\max_a\sum_{s',r}P(s',r|s,a)\Big[r(s,a)+ \gamma V_*(s')\Big],\nonumber
\end{align}
and, 
\begin{align}
Q_*(s,a)=\sum_{s',r}P(s',r|s,a)\Big[r(s,a)+\gamma \max_{a'}Q_*(s',a')\Big].\nonumber
\end{align}
Bellman optimality equations are nonlinear where the solution could be found through iterative algorithms. This could be achieved by means of value-based RL algorithms. Other developed RL algorithms can be categorized into either policy-based or actor-critic (AC) algorithms \cite{najar2021reinforcement}. In this regard, the anatomy of RL algorithms is illustrated in Fig. \ref{fig2}. The general idea is to either fit a model or estimate the return by either updating the policy or by generating new training samples. 

Value-based algorithms involve the iterative optimization of the value function to obtain the optimal policy. Q-Learning \cite{watkins1992q} and SARSA \cite{sutton1995generalization} are two of the well-known value-based algorithms. As it can be observed from Fig. \ref{fig2}, the Q-Learning algorithm involves the fitting of the action-value function $Q(s,a)$ and tries to improve the policy through $\underset{a}{\argmax}Q_{\pi}(s,a)$. In other words, Q-Learning attempts to compute the action-value function of the optimal policy iteratively. In this regard, the taken action $a_t$ by the agent at time step $t$ within state $s_t$, transits the environment state to $s_{t+1}$ and produces a reward of value $r_t$, for which the action-value function can be updated employing the following update rule:
\begin{align}\label{eq6}
Q(s_t,a_t)\leftarrow & Q(s_t,a_t)+  \alpha\left[r_t+\gamma \max_{a'\in A}Q(s_{t+1},a')-Q(s_t,a_t)\right],
\end{align}   
where $\alpha\in[0,1]$ is the learning rate. Eq. \ref{eq6} denotes that the new best estimate of the action-value function, i.e., $r_t+\gamma \max_{a'\in A}Q(s_{t+1},a')$, is compared with the value of the previous time step, and, then, it is multiplied by the learning rate to be added to the old estimate of the value to update the Q-Table. In contrast to the Q-Learning algorithm that is independent of the policy being followed, the SARSA algorithm learns the action-value function by involving the taken action at time step $t+1$, where the update rule is as follows:
\begin{align}
Q(s_t,a_t)\leftarrow & Q(s_t,a_t)+ \alpha\left[r_t+\gamma Q(s_{t+1},a_{t+1})-Q(s_t,a_t)\right].
\end{align}

Tabular value updates are not a suitable solution when dealing with a large number of state-action pairs and when the state/action space is continuous. This could be addressed more efficiently using deep reinforcement learning (DRL). In value-based methods, one notable effort is deep Q-networks (DQN) \cite{mnih2015human}, where a deep network $Q(s_t,a_t|\theta)$ is used to estimate the Q-function $Q^*(s_t,a_t)$, with $\theta$ being the network's parameters to be trained given the following target:
\begin{align}\label{eq1_rev}
Y_t^{\text{DQN}} \equiv r_{t+1} + \gamma \max_{a} Q(s_{t+1}, a|\theta'_t),
\end{align}
where $\theta'$ are the parameters of the target network used for stabilizing the training. Original DQN uses a uniform sampling from a replay memory to extract a minibatch of transitions to train the network on. However, it is well-studied that uniform sampling is not sample-efficient, and techniques such as prioritized experience replay (PER) \cite{schaul2015prioritized} are developed to speed up training by a targeted sampling, where transitions with higher TD errors are sampled more frequently to train the network's parameter. Besides, there are other variants of DQN, such as double DQN (DDQN) \cite{van2016deep}, to cope with overoptimistic value estimates in DQN. The issues goes back to the $\max$ operator used in (\ref{eq1_rev}) that uses the same values both to select and evaluate actions. DDQN makes use of an extra network for each $Q$ and its target is as follows:
\begin{align}
Y_t^{\text{DDQN}} \equiv r_{t+1} + \gamma Q(s_{t+1},\argmax_aQ(s_{t+1},a|\theta_t)|\theta_t^-),
\end{align}
where $\theta_t^-$ is the target network's parameters. Dueling DQN \cite{wang2016dueling} is another variation of DQN that incorporates the advantage value alongside the state value within the Q-function. This is achieved by splitting the network into two streams: one estimates state values, and the other calculates state-dependent action advantages. The final module of the network then merges these outputs. This approach explicitly separates the representation of state values and action advantages, allowing the network to discern valuable states without the need to evaluate every action's effect at each timestep.

Policy-based methods \cite{williams1992simple}, in contrast to the value-based techniques, do not aim at estimating the value function, but they directly parameterize the policy $\pi_{\theta}(a|s)$ with a set of parameters $\theta$. These techniques show better convergence properties compared with the value-based methods, are more effective in high-dimensional action spaces, and could learn stochastic policies as well. However, they suffer from high variance and get stuck in the local optimum. In these techniques, the general idea is to find the set of parameters $\theta$ for a policy that optimizes an objective function $J(\theta)$. For instance, for policy gradient (PG) agents, this objective function is the expected return and the optimization is done through the gradient descent algorithm. REINFORCE (Monte Carlo PG) is another policy-based algorithm, for which the stochastic gradient descent (SGD) is employed to optimize the set of parameters. In particular, the set of policy parameters is initialized randomly and an episode experience is generated by following the policy $s_0,a_0,r_1,s_1,\ldots,s_{T-1},a_{T-1},r_T,s_T$, where $s_T$ denotes the terminal state. For states in this episode sequence, the expected return $G_t=\sum_{k=t}^T\gamma^{k-t}r_k$ is calculated, and, by accumulating the gradients for the PG to maximize the expected return, the set of parameters can be updated through the following update rule:
\begin{align}
\theta\leftarrow\theta+\alpha\sum_{t=1}^{T-1}G_t\nabla_{\theta}\ln\pi(s_t|\theta).
\end{align} 

AC is known as a hybrid method that involves the evaluation and calculation of the policy (the actor) and value function (the critic) \cite{barto1983neuronlike}. This technique maintains two sets of parameters, where the actor updates the policy parameters $\theta$ employing the PG algorithm, and the critic updates parameters $w$ of the value function for the sake of function approximation. In this regard, parameters of the actor $\pi$ and the critic $V$ are arbitrarily initialized by $\theta$ and $w$, and, then, the taken action within state $s$, i.e., $a\sim\pi_{\theta}(s)$, is evaluated through the temporal different (TD) error, $A(s)=r+\gamma V_w(s')-V_w(s)$, for which $\theta$ and $w$ can then be updated accordingly:
\begin{align}
w & \leftarrow w-\alpha\nabla_w\|A(s)\|^2\\
\theta & \leftarrow \theta+\beta \nabla_{\theta}\log\pi_{\theta}(s)A(s).
\end{align}

A key challenge in the design of PG algorithms such as REINFORCE is drastic policy updates that makes it difficult to ensure stable training. Proximal policy optimization (PPO) \cite{schulman2017proximal} handles this limitation by assigning a constraint to the policy update rule to make sure the updates will remain within a proximal region. In this context, the objective function of PPO is to maximize the expected return while ensuring the policy update remains close to the current policy, as given below:
\begin{align}
\mathcal{L}(\theta) = \mathbb{E}_t \left[ \frac{\pi_{\theta}(a_t|s_t)}{\pi_{\theta_{\text{old}}}(a_t|s_t)} A_t - \beta \text{KL}[\pi_{\theta_{\text{old}}}(\cdot|s_t), \pi_{\theta}(\cdot|s_t)] \right],
\end{align}
where $\pi_{\theta}(a_t|s_t)$ is the probability of taking action $a_t$ in state $s_t$ under the policy parameterized by $\theta$, and $\pi_{\theta_{\text{old}}}(a_t|s_t)$ is the probability under the old policy. $A_t$ represents the advantage function, and the term $\text{KL}[\pi_{\theta_{\text{old}}}(\cdot|s_t), \pi_{\theta}(\cdot|s_t)]$ denotes the Kullback-Leibler (KL) divergence between the old and updated policies. TRPO \cite{schulman2015trust} is another notable PG algorithm developed for ensuring the stability of policy updates. TRPO also assigns a constrain to the policy updates to make sure updates remain within a trust region through a surrogate objective function given below:
\begin{align}
\theta^* = \arg \max_{\theta} \mathbb{E}_t \left[ \frac{\pi_{\theta}(a_t|s_t)}{\pi_{\theta_{\text{old}}}(a_t|s_t)} A_t \right],
\end{align}
subject to the constraint:
\begin{align}
\mathbb{E}_t \left[ \text{KL}[\pi_{\theta_{\text{old}}}(\cdot|s_t), \pi_{\theta}(\cdot|s_t)] \right] \leq \delta
\end{align}
where $\delta$ is a hyperparameter that controls the size of the trust region.

\begin{figure}
\centering
\includegraphics[width=0.8\columnwidth]{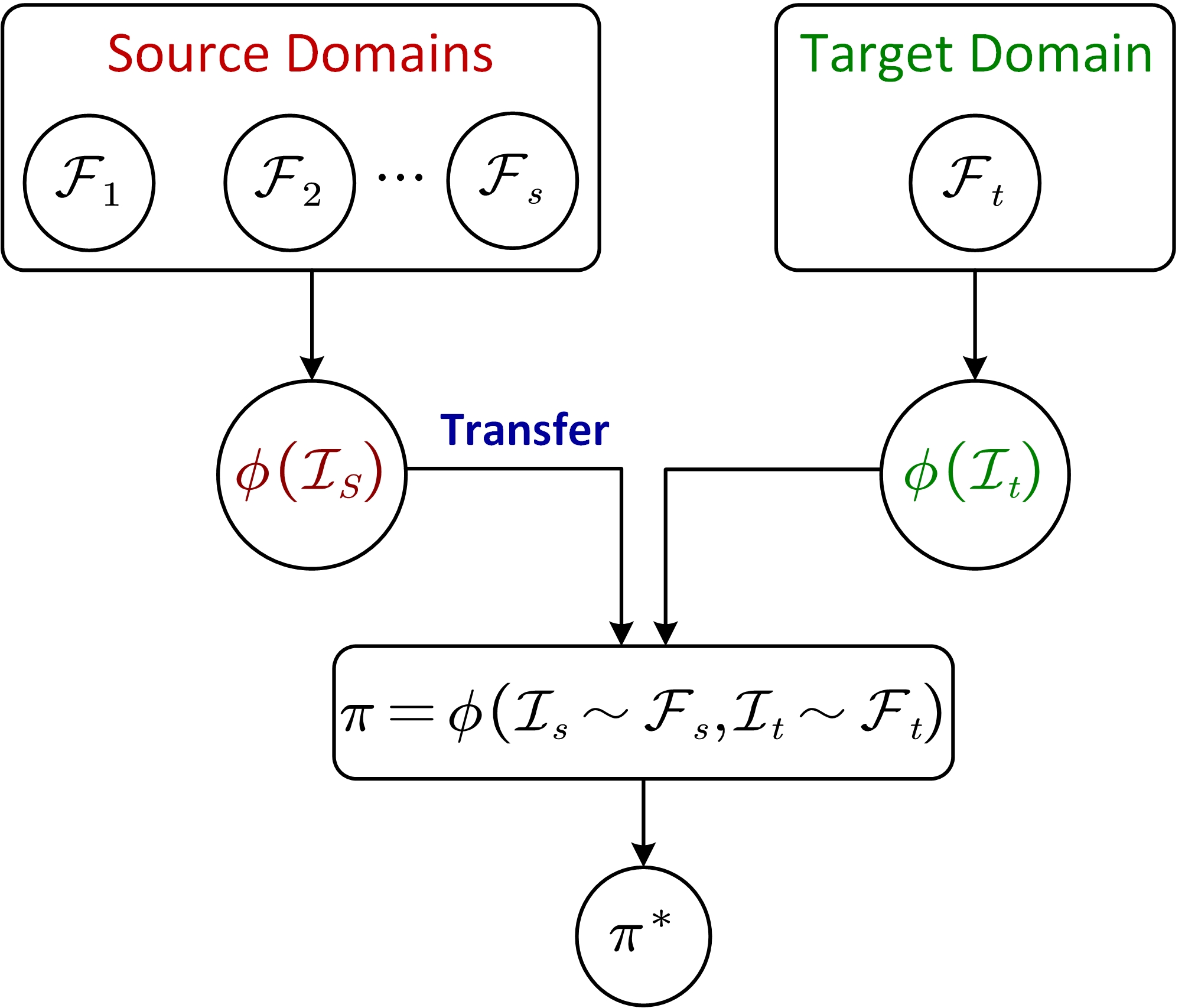}
\caption{Transfer learning in RL.}\label{fig4}
\end{figure}

\subsection{Transfer Reinforcement Learning}\label{secTRL}
TL refers to a learning scheme that leverages the learned knowledge from external expertise and domains to improve the learning process in the target domain of interest. TRL can be defined as the application of TL under the RL framework \cite{zhu2020transfer}. As it can be observed from Fig. \ref{fig4}, TRL could be thought of as finding an optimal policy $\pi^*$ for a target domain given several source domains $\mathcal{\mathbf{M}}_s=\{\mathcal{M}_s|\mathcal{M}_s\in \mathcal{\mathbf{M}}_s\}$ and a target domain $\mathcal{M}_t$, employing the extracted knowledge from external domains $\mathcal{I}_s$ and the interior domain $\mathcal{I}_t$. The extracted knowledge from source domains $\mathcal{I}_S$ is transferred and augmented with that learned from the target domain $\mathcal{I}_t$ to construct the policy of the target domain shown by $\pi=\phi(\mathcal{I}_s\sim\mathcal{M}_s,\mathcal{I}_t\sim\mathcal{M}_t):\mathcal{S}^t\rightarrow\mathcal{A}^t$. Having $\pi$ constructed, the TRL problem could be defined as finding the optimal policy:
\begin{align}
\pi^*=\argmax_{\pi}\mathbb{E}_{s\in\mathcal{S},\pi\sim\pi}\left[Q^{\pi}_{\mathcal{M}}(s,a)\right].
\end{align}
A common way by which TRL algorithms can be distinguished and categorized is based on the type of transferred knowledge. In this regard, TRL algorithms can be categorized into algorithms developed based on reward shaping (RS), learning from demonstration (LfD), and policy transfer (PT). 

\begin{figure}
\centering
\includegraphics[width=0.8\columnwidth]{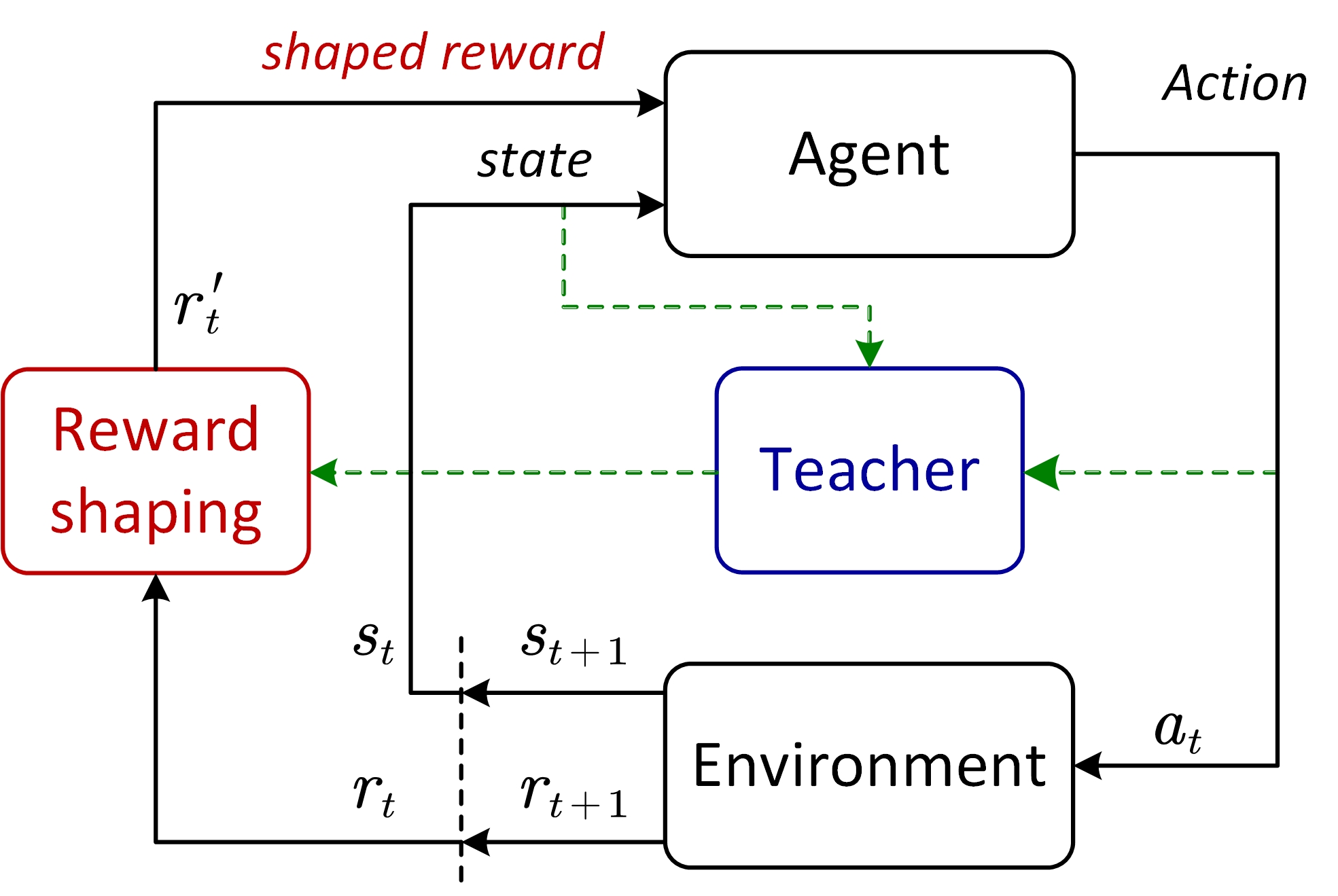}
\caption{Reward shaping in RL.}\label{fig5}
\end{figure}

\subsubsection{Reward Shaping}
RS refers to a category of techniques, in which an RL agent is provided with an extra reward signal $R_F$ in addition to what it receives from environment, i.e., $R$, to improve the learning process. As it can be observed from Fig. \ref{fig5}, the agent is thus supposed to learn a composite reward $R'=R_F+R$, where the shaped reward $R'$ encodes the extra knowledge that could deal with the issue of sparse reward signals \cite{DEMOOR2022535}. Even though RS can help with improving the learning process, however, it might cause changes in the designated task of an RL agent since rewards define the task of an RL agent. To address this problem, the authors in \cite{ng1999policy} proved that one way to shape the reward without changing the underlying task of the agent can be realized through potential-based shaping. In this approach, there is a need to define a potential function $\Phi$ over the state space $\mathcal{S}$ and define the shaped reward $R_F$ as follows:
\begin{align}
R_F(s,a,s')=\gamma \Phi(s')-\Phi(s),
\end{align}
where it stands for the deviation between the potential of states $s$ and $s'$. This technique was also extended to the RS over state-action pairs as given below \cite{wiewiora2003principled}:
\begin{align}
R_F(s,a,s',a')=\gamma\Phi(s',a')-\Phi(s,a),
\end{align}
where it involves more information pertaining to the states and actions. 

These potential-based RS methods, however, are static and are built based upon the assumption that potential of a state does not change dynamically \cite{badnava2023new}. This assumption is often broken, where dynamic potential-based RS methods have been developed to cope with this shortcoming \cite{devlin2012dynamic}. Dynamic methods incorporate time-varying potential functions that instead of solely relaying on the states and actions to shape the reward, they also consider the temporal aspects of the interactions the agent makes with the environment. In this respect, the potential function evolves over time based on factors such as experienced trajectories by the agent, transitions of states and the progress of the underlying task. For a dynamic potential function $\Phi(s,t)$, where $t$ represents time, the shaped reward can be of the following form:
\begin{align}
RF(s, a, s', t) = \gamma\Phi(s', t+1) - \Phi(s, t),
\end{align}
to capture the change in potential between the current state $s$ at time $t$ and the next state $s'$ at time $t+1$ in response to the action $a$ taken by the agent. The incorporation of time into the potential function makes the shaped reward adapt to the evolving dynamics of the environment and provides the agent with timely feedback to guide the learning process.

\subsubsection{Learning From Demonstrations}
The general idea in LfD is to provide an RL agent with an external set of demonstrations $\mathcal{F}$ to guide the agent towards exploring more beneficial states for the sake of learning the optimal policy more efficiently. The provided demonstrations can either be collected from different source domains $\{\mathcal{F}_1,\ldots,\mathcal{F}_s\}$ or the source domain which is built based upon the same MDP as that of the target source, i.e., $\mathcal{F}_s=\mathcal{F}_t$. In either case, the source domain consists of tuples of the form $(s,a,s',r)$, where it can be prepared by an expert or a previously learned optimal or sub-optimal policy. 

\subsubsection{Policy Transfer}
In PT techniques, the external knowledge to be fed into the learning process of a target domain is a set of pre-trained policies that are learned using either one or multiple domain resources. In other words, PT techniques make use of a set of teacher policies $\pi_E=\{\pi_{E_1},\pi_{E_2},\ldots,\pi_{E_s}\}$, which are trained on source domains $\{\mathcal{F}_1,\mathcal{F}_2,\ldots,\mathcal{F}_s\}$. A student policy $\pi$ can then be learned for the target domain $\mathcal{F}_t$ by the transferred knowledge from the set of pre-trained policies $\pi_E$. 

\begin{figure}
\centering
\includegraphics[width=0.8\columnwidth]{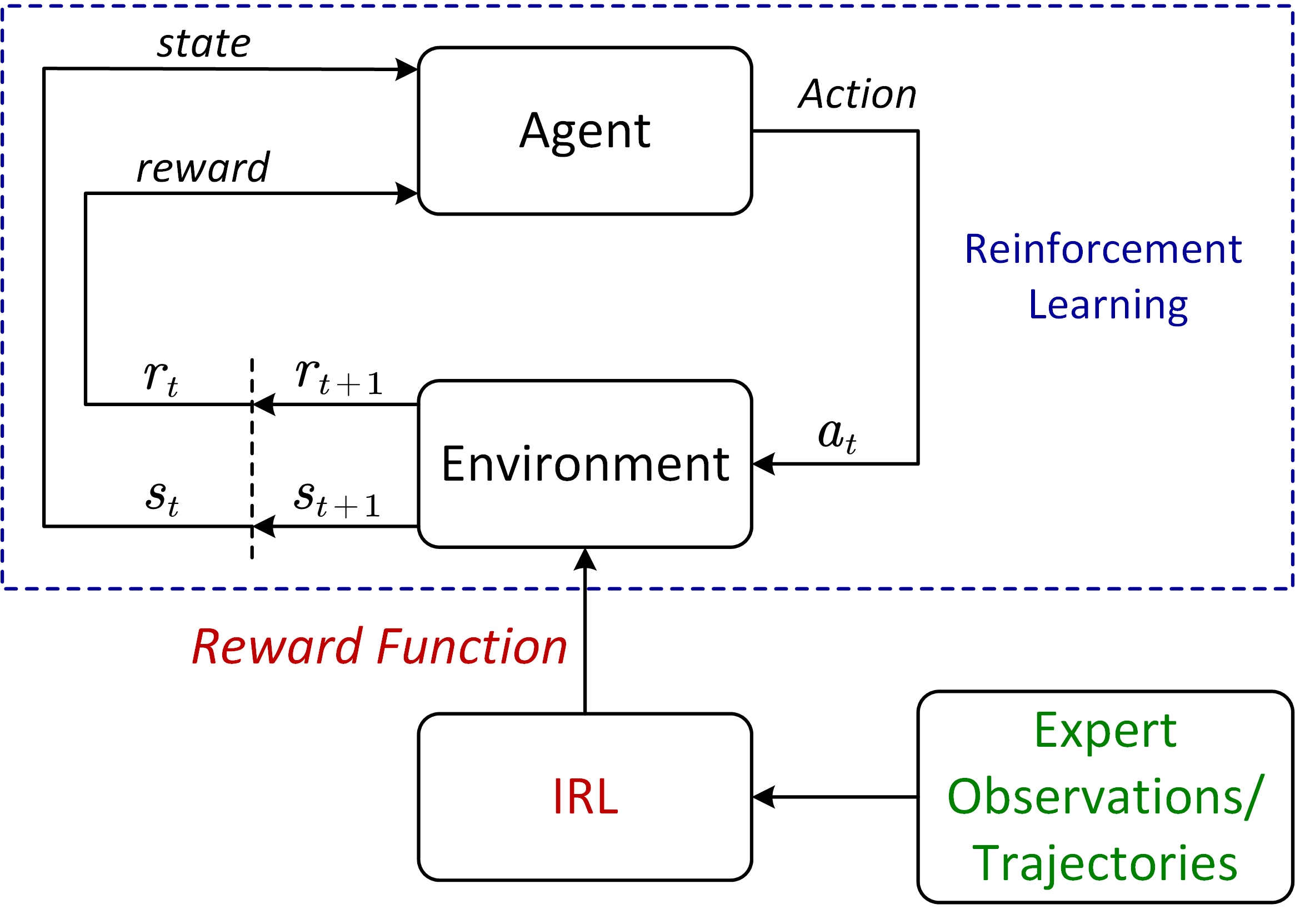}
\caption{The general structure of IRL.}\label{fig3}
\end{figure}

\subsection{Inverse Reinforcement Learning}\label{secIRL}
IRL is the problem of either learning or estimating the reward function using the provided demonstrations and trajectories from environment. Demonstrations were initially assumed to be given by an expert, however, this assumption has been further relaxed recently. IRL makes use of given demonstrations in terms of trajectories of action-state pairs:
\begin{align}
\mathcal{T}_m=\{(s_1,a_1),(s_2,a_2),\ldots,(s_{T_m},a_{T_m})\},
\end{align}
where $m=1,\ldots,M$ denotes the number of experts and $T_m$ stands for the number of time steps in the $m$th trajectory. We denote the set of all trajectories by $\mathcal{T}$. The general structure of IRL is illustrated in Fig. \ref{fig3}, where IRL could be thought of as an attempt towards recovering the reward function. It was first introduced by Russell in 1998 \cite{russell1998learning}, and it is defined as the problem of determining the reward function utilizing the agent's behavior over time and the environment model, where the MDP is the most popular model used in IRL. As mentioned earlier, an MDP can be represented utilizing a tuple $\mathcal{M}=\langle S, A, T, R, \gamma\rangle$. Then, IRL can be officially defined as given below.
\begin{definition}
\cite{ARORA2021103500} Given an MDP of the form $\mathcal{M}$, suppose that $\mathcal{M}/R_m$ represents an MDP with no reward function and it models the interactions of the $m$th agent with the environment. For the set of demonstrations $\mathcal{T}=\{\mathcal{T}_m|m=1,\ldots,M\}$ with $\mathcal{T}_m=\{(s_1,a_1),(s_2,a_2),\ldots,(s_{T_m},a_{T_m})\}$ being the $m$th expert trajectories, the IRL problem is determining $\hat{R}_m$ which best describes the observed behavior in the $m$th trajectory.
\end{definition}
The developed IRL models share some common steps and are generally built based upon the following procedure \cite{adams2022survey}: (1) Given the expert demonstrations $\mathcal{T}$, create an MDP without the reward function through the provided demonstrations; (2) Formulate the reward utilizing parameters such as states, state features, etc; (3) Realize the behavior of the MDP (e.g., concerning policy or state visitation frequency) under the formulated reward function; (4) By minimizing the deviation between the behavior of the current MDP and that of the experts given in demonstrations, update the parameters of the constructed reward function; (5) Repeat these steps until a criterion based on the aforementioned deviation is met. 

Following this procedure, many efforts have been devoted to addressing the IRL problem, where the developed techniques can be divided into techniques based on margin optimization, entropy optimization, and Bayesian update. 

\subsubsection{IRL-Margin Optimization}
These techniques inquire a reward function that explains the given demonstrations in $\mathcal{T}$ better than all the other possible policies by a pre-defined margin. One way to realize IRL using margin optimization is through linear programming by assuming the expert's policy is the optimal solution to an MDP \cite{ng2000algorithms}. Within this framework, the margin is constructed by quantifying the deviation between the action values of each state of a trajectory and that of the best trajectory among the remaining trajectories, where the agent's policy is utilized to initiate the policy in the margin. The aim is then to retrieve a linear reward function:
\begin{align}
R(s,a)& \doteq w.\phi(s,a)\nonumber\\
& =w_1\phi_1(s,a)+w_2\phi_2(s,a)+\ldots+w_K\phi_K(s,a)\nonumber\\
& =\sum_{k=1}^kw_k\phi_k(s,a),
\end{align}
where $\phi_k(s,a)$ are some vector of features for which $\phi(s,a):S\times A\rightarrow [0,1]^K$, and $w_k$ are the feature weights.

Another well-known technique is apprenticeship learning where in contrast to linear programming, it involves expert demonstrations \cite{abbeel2004apprenticeship}. This technique assumes that states of the environment are characterized by means of some features which pose real values. The vector of feature expectations $\mu(\pi)$ can be defined by resorting to the expected discounted accumulated feature value of a given policy:
\begin{align}
\mu(\pi)=\mathbb{E}\left[\sum_{t=0}^{\infty}\gamma^t\phi(s_t)|\pi\right].
\end{align}
Following this, by resorting to the set of demonstrations by some expert $\pi_E$, one can then define the expert's feature expectation that is required to be estimated as follows:
\begin{align}
\hat{\mu}_E=\hat{\mu}(\pi_E)=\mathbb{E}\left[\frac{1}{M}\sum_{i=1}^M\sum_{t=0}^{\infty}\gamma^t\phi(s_t^i)\right],
\end{align}
where the ultimate goal is to find an optimal policy $\tilde{\pi}$ such that $\|\mu(\tilde{\pi})-\mu_E\|_2\leq \epsilon$, which conducts the fact that policy $\tilde{\pi}$ should have a close feature expectation $\mu(\tilde{\pi})$ to $\mu_E$. 

\subsubsection{IRL-Entropy Optimization}
Margin maximization methods make IRL problems ill-posed due to the degeneracy problem. This problem comes from the fact that several policies, and, therefore, reward functions, might satisfy the feature expectation constraint, especially when the reward function produces zero for every state \cite{buning2022interactive}. Maximum margin methods are also built based on the assumption that the given demonstrations are the results of an optimal policy. Moreover, there always exists noise or sub-optimal behavior in the expert's demonstrations. To deal with such shortcomings, entropy optimization techniques have been developed by relying on the probabilistic maximum entropy approach \cite{jaynes1957information}.

This technique parameterizes the distribution over trajectories by means of reward weights $w$ in a way that the probability of a trajectory to be performed by an agent is exponentially dependent on its expected reward, as defined below:
\begin{align}
\mathbb{P}(\mathcal{T}_i|w)=\frac{1}{Z(w)}e^{\sum_{\langle a,s\rangle\in \mathcal{T}_i}w.\phi(s,a)},
\end{align}
where $Z(w)$ is called the partition function, which always converges for both finite and infinite horizon problems by the given reward weights. This probability distribution makes the trajectories with the same reward have equal probabilities, and trajectories with a higher reward are exponentially more preferred. Having the probability distribution of trajectories formulated in terms of the reward function, the goal of the maximum entropy methods is to find reward weights $w$ such that they maximize the log-likelihood of the probability distribution function as follows:
\begin{align}
w^*=\argmax_{w}L(w)=\argmax_w\log\prod_{\mathcal{T}_m\in\mathcal{T}}\mathbb{P}(\mathcal{T}_m|w).
\end{align}

\subsubsection{IRL-Bayesian Update}
The Bayesian update is a fundamental IRL technique that is constructed based on Bayes' theorem for updating a prior distribution by means of candidate reward functions \cite{9334410}. The Bayes' theorem deals with updating the probability of a hypothesis once more information becomes available \cite{schafer2022bayesian}. The Bayes rule can be stated as follows:
\begin{align}
\mathbb{P}(A|B)=\frac{\mathbb{P}(B|A)\mathbb{P}(A)}{\mathbb{P}(B)},
\end{align}
where $A$ denotes the event for which the probability is going to be found, and $B$ stands for the new information. Then, a posterior distribution over available reward functions is:
\begin{align}\label{eq18}
\mathbb{P}(R_E|T_m)=\frac{\mathbb{P}(T_m|R_E)\mathbb{P}(R_E)}{\mathbb{P}(T_m)},
\end{align}
where $\mathbb{P}(T_m|R_E)=\prod_{\langle s,a\rangle\in T_m}\mathbb{P}\left(\langle s,a\rangle|R_E\right)$. The update rule (\ref{eq18}) will be continued until all the provided trajectories in the set of demonstrations are examined.

\subsection{Sample complexity and convergence rate}
One research question in RL studies is to understand how many transitions or experiences are required for the agent to act well in dealing with an environment. This is known as the sample complexity of RL and refers to the number of time steps on which an RL algorithm may select an action whose value is not near-optimal \cite{dann2015sample}. In this context, RL algorithms whose sample complexity is a polynomial function of some domain parameters are called probably approximately correct (PAC) \cite{kearns1998finite,brafman2002r}. For fixed-horizon PAC RL algorithms, Kakade \cite{kakade2003sample} proved upper and lower PAC bounds for the case, in which the agent interacts indefinitely with the environment and showed that there are not more than $\tilde{O}\left(\frac{|S||A|H^6}{\epsilon^3}\ln\frac{1}{\delta}\right)$ time steps in which the agent performs $\epsilon$-suboptimal, where $|.|$ denotes the cardinality of the enclosed set, $H$ denotes the time horizon, and $\delta$ is the failure probability. This was later on improved by Strehl et al. \cite{strehl2006pac} for a delayed Q-learning algorithm to $\tilde{O}\left(\frac{|S||A|H^5}{\epsilon^4}\ln\frac{1}{\delta}\right)$ and to $\tilde{O}\left(\frac{|S|^2|A|H^3}{\epsilon^2}\ln\frac{1}{\delta}\right)$ by Jaksch et al. \cite{auer2008near} for the episodic case. More recently, it has been argued that the key feature enabling sample-efficient RL algorithms is the generalization, where without a generalization model, the sample complexity will be dependent to the size of action space \cite{mou2020sample, azar2017minimax}. Besides, there is currently a surge of interest in understanding the dependency of sample complexity to the time horizon, where PAC RL algorithms usually learn an $O(1)$-optimal policy using $\textit{polylog(H)}$ episodes of interactions with environment \cite{jiang2018open}, which has been more relaxed in \cite{li2022settling} by achieving the same PAC guarantee with only $O(1)$ episodes of interactions with the environment. The study of sample complexity has also been extended to IRL with finite state and action \cite{komanduru2021lower,komanduru2019correctness},  continuous state space with unknown transition dynamics \cite{dexter2021inverse}, and RL algorithms with myopic exploration such as $\epsilon$-greedy and softmax \cite{dann2022guarantees}.

In contrast to sample complexity that focuses on the number of samples or interactions needed to achieve a satisfactory level of performance, convergence rate refers to the speed at which an RL algorithm converges to a near-optimal or optimal policy or value function \cite{li2019convergence}. The question to be addressed by the convergence rate analysis is how quickly an agent can learn to make optimal decisions in an underlying environment \cite{even2003learning}. In this context, one common practice used in improving the convergence rate of RL algorithms is function approximation \cite{tosatto2017boosted}. The idea is enabling RL algorithms generalize well from observed experiences to unobserved states or actions. Other than function approximation, exploration strategy and learning rate scheduling are crucial factors influencing the convergence rate \cite{pathak2017curiosity}. Exploration mechanisms such as $\epsilon$-greedy and upper confidence bound balance the exploration-exploitation trade-off and learning rate scheduling dynamically adjusts the step size of parameter updates that helps with ensuring RL algorithms converge more quickly without overshooting optimal solutions. 

\section{Transfer Reinforcement Learning}\label{sec3}
This section resorts to the fundamental TRL techniques given in Section \ref{sec2} to study several major improvements of such models in terms of sample efficiency and generalization. 

\subsection{Human-in-the-loop}
Through the use of large amounts of training data, the human-in-the-loop (HITL) architecture improves the learning speed of RL algorithms, particularly in real-world applications where agent-hardware interactions are expensive, and agent actions might result in catastrophic hardware failure. Leveraging human knowledge in the learning process could be of utmost importance to reduce the risk of destructive actions and to shape the behavior of RL agents \cite{goecks2020human}. One way to categorize the learning models under the HITL concept is through the level of human involvement, by which these models can be categorized into active learning, interactive learning, and machine teaching \cite{holmberg2020feature}. In active learning, the learning process is under the control of the system and the human could just annotate the unlabeled data. Interactive learning is the case, in which there exist more frequent interactions between humans and the system in the learning process compared with active learning. Machine teaching refers to models that the learning process is under the control of the human so that they are capable of transferring their knowledge into the learner \cite{mosqueira2022human}. 

One of the HITL mechanisms is the situation that the human is queried by the agent for demonstrations in some states. This mechanism could be thought of as active learning, where the ultimate goal is to improve learning performance by imitating the human. One way to improve the learning process is by directly learning a reward function through the provided feedback by humans. TAMER \cite{knox2009interactively} is one of the well-known frameworks for the sake of learning a reward function through human feedback, where the agent refers to the human's reinforcement to exploit the desired action. TAMER was then used in \cite{christiano2017deep}, to firstly scale it to deep RL (DRL), and, secondly, improve the frequency of agent-human interactions, where it has been shown that the agent queries the human for the feedback on reward in less than 1\% of interactions. TAMER was also scaled up in \cite{arakawa2018dqn}, where not only a more precise model of the human observer is proposed, but also an algorithm called deep Q-network (DQN)-TAMER is devised that benefits from both the human feedback and additional rewards. Following the modeling of human behavior, it has been argued \cite{lindner2022humans} that the developed models should be more personal \cite{alamdari2020personalization}, dynamic \cite{chen2022asha}, and contextual \cite{hill2020human}. Other than the reward, the agent could also benefit from the human's feedback on the observation \cite{keramati2019value}, actions \cite{verma2022advice}, and state-action pairs \cite{guan2021widening}. 

Another way to realize HITL in RL is through modifying the training loss function so as to leverage the given demonstrations by a human in the learning process. One of the well-known techniques is the DQN from demonstrations (DQfD) \cite{hester2018deep}, where it is proposed to involve four loss functions including the one-step and $n$-step double Q-learning losses, maximum margin classification loss, and regularization of weights and bias in the network. DQfD is concerned with discrete action space, whereby removing the constraint on the classification loss, the authors in \cite{vecerik2017leveraging} built the deep deterministic policy gradient (DDPG) from demonstration (DDPGfD) on DQfD to deal with continuous action spaces. Later in \cite{pohlen2018observe}, the authors proposed an algorithm called Ape-X DQfD, which combines the Ape-X DQN \cite{horgan2018distributed} and DQfD in order to benefit from a prioritized replay buffer and multiple critic networks to make the agent learn human-level policies. %More recently, it was suggested in \cite{cruz2019jointly} to augment an unsupervised loss with the value function loss along with the aforementioned supervised classification loss to train the agent to learn more useful features from a lower number of demonstrations. 

Other than providing the agent with feedback on the reward, states, and actions, and modifying the training loss function, some other HITL techniques have also been recently developed. One of these techniques makes use of the teacher-student framework \cite{abel2017agent}, where the teacher is typically a human and the RL agent is the student \cite{navidi2021new}. In such a framework, several goals could be defined. The ultimate goal, however, could be to improve the learning performance of the student through interactions by the teacher. To fulfill this goal, the student should make the best of use of the provided advice by the teacher, try to minimize the amount of advice from the teacher, and ensure the training is not impaired by a sub-optimal teacher \cite{taylor2021improving}. %Other than receiving feedback from the student, it could also be beneficial for the teacher to receive feedback (e.g., advice conformance verification) from the student \cite{verma2022advice}.

\subsection{Sim-to-Real Transfer}
Simulation-to-reality (sim-to-real) transfer refers to deploying a trained RL agent on a simulator environment into the corresponding real environment \cite{zhao2020sim}. Due to the potential differences between the simulator the agent trained on, and the real environment in which the learned policy is evaluated, also known as the sim-to-reality gap or simply, the \textit{reality gap}, the policy trajectories of both environments differ considerably \cite{matas2018sim}. Therefore, there is a need to bridge this gap for deploying RL agents in real environments. The sim-to-real transfer is usually performed in two ways in the RL literature, namely, the zero-shot transfer and domain adaptation \cite{chen2022zero}. Zero-shot transfer or direct transfer is a straightforward technique that directly transfers the learned knowledge from the simulator to the real environment, where domain randomization is one of the well-known techniques in this area \cite{bi2021zero}. Domain adaptation is a sub-category of TL, in which the ultimate goal is to make use of the data from a source domain so as to improve the efficiency of the learner in the target domain \cite{zhang2019vr}. 

Domain randomization is the idea of randomizing the simulation environment with diverse scenarios with the hope of covering the distribution of data in the real environment to get a better generalization performance \cite{tobin2019real}. It could be achieved, as proposed in \cite{chaffre2020sim}, by splitting the RL task into several environments with an incremental environment complexity and by training the policy using these diverse environments. The policy could then be learned by maximizing the average expected reward across all data distributions, where the domain randomization is modeled by a parameterized distribution involved in the learning process of the policy with maximal performance. Domain randomization could also be realized by uniformly randomizing the environment parameters \cite{candela2022transferring,muratore2021data} and the set of observations and actions of the agent in the simulator \cite{lobos2020point}. However, such methodologies could be sample-inefficient. A more sample-efficient mechanism could be dividing the range of randomizing parameters into several sub-domains to be used for learning local policies w.r.t. each sub-domain and then augmenting all the learned policies into a global one for the sake of sim-to-real transfer \cite{kadokawa2022cyclic}. 

Domain adaptation refers to transferring an already learned policy in simulations to the real environment for further adaptation \cite{arndt2020meta}. One way to realize domain adaptation is through meta-learning paradigm \cite{yao2022learning}. The basic principle of meta-learning, or \textit{learning to learn}, is to leverage data from previously learned tasks to enable the learner to adapt to new tasks. The primary work on model-based meta-learning for continuous domain adaption is the work of Nagabandi et al. \cite{nagabandi2018learning}, which is employed for real-world robotic applications. The recent trends in meta-learning for domain adaptation could be considered as memory-based and gradient-based meta-learning, where in the former, the memory helps with quickly learning never-before-seen tasks, and the latter proposes a good starting point for the further adaptation \cite{duan2016rl,finn2017model}. Other than meta-learning, techniques for enabling the self-supervised adaptation \cite{jeong2020self}, retraining of neural networks with optimized parameters by RL \cite{lonvcarcvic2021accelerated}, adversarial RL \cite{jiang2021simgan}, ensemble learning \cite{exarchos2021policy}, and application-oriented domain adaption \cite{park2021sim} are of some recent developments in sim-to-real TL.

\subsection{Policy Transfer}
There are several ways for knowledge transfer in RL based on the type of transferred information. Among them, transferring the learned value function and policy in the source domain(s) to the target domain are widely-studied \cite{joshi2021adaptive}. Transferring the value function is usually referred to as the direct transfer, where the to-be-learned value function in the target domain is initially approximated by the transferred value function from the source domain \cite{fernandez2021probabilistic}. This approach, however, suffers from degrading the convergence speed in the target domain when the already learned value function is not close enough to the optimal one in the target domain \cite{barekatain2019multipolar}. Besides, such a direct method might also not be efficient for multi-source tasks, where it could become complicated to select the task from which the policy is going to be transferred \cite{yang2020efficient}. 

PT methodologies have been studied in \cite{zhu2020transfer}, where the developed models are categorized under either \textit{policy distillation} or \textit{policy reuse}. The former refers to an ensemble of available knowledge in the source domains to be used in the target domain through a teacher-student learning mechanism. The latter, however, stands for directly reusing the learned policies in the source domains to construct the policy in the target domain through a combination of source-domain policies. Since then, several advancements have been made to both categories of PT techniques. Towards this end, a category of models is devised to deal with multi-agent frameworks, in which agents are trained independently, where adding a new agent to the problem needs training from scratch. In this regard, scaling the PT methodologies up to the multi-agent RL could be of utmost importance for the sake of sample efficiency \cite{shi2022knowledge,shi2021lateral,zhu2022context}. In the context of knowledge transfer, it is also studied that along with transferring the learned policy, the transfer of model (state transition and reward function) could also improve the learning efficiency in the target domain, leading to the concept of multi-knowledge transfer in RL \cite{fernandez2021probabilistic,tao2021repaint}. Further to this, a common assumption in PT techniques is to consider the same dynamics in the source domains that limit their applications in a real-world setting, whereas a more general case could consider different dynamics in the source domains that require a mixture of experts for learning a more robust policy in the target domain \cite{gimelfarb2021contextual}. Table \ref{tab1} summarizes the recently-developed TRL algorithms. 

\begin{table*}
\scriptsize
\caption{An overview and comparison of recent TRL algorithms.}\label{tab1}
\begin{tabular}{llll}
\hline Category & Learner/Algorithm & Notable Features & Citation\\
\hline Human-in-the-loop & DQfD & Accelerating the learning process with \\
 & & a small set of demonstrations & \cite{hester2018deep}\\
 & DDPG & Prioritized replay buffer & \cite{vecerik2017leveraging}\\
 & Ape-X DQN & Prioritized replay buffer with decoupling \\
 & & acting from learning & \cite{horgan2018distributed}\\
 & A3C & Augmenting supervised and value losses in pre-training & \cite{cruz2019jointly}\\
 & TAMER & Policy shaping by a human trainer & \cite{knox2009interactively}\\
 & TAMER & Agent-agnostic & \cite{abel2017agent}\\
 & DQN-TAMER & The precise model of human observer & \cite{arakawa2018dqn}\\
 & A2C & Low rate of shaping feedback in agent interactions & \cite{christiano2017deep}\\
 & ASHA & Offline pre-training and online shaping feedback & \cite{chen2022asha}\\
 & DRoP & Dynamic reuse of prior knowledge & \cite{wang2018interactive}\\
 & DQN  & Human visual explanation and\\
 & &  text-based language model & \cite{guan2021widening,hill2020human}\\
Sim-to-Real & DDPG & Domain randomization using a \\
& & combination of DRL algorithms & \cite{matas2018sim}\\
 & SimTwin & Zero-shot transfer learning & \cite{chen2022zero,bi2021zero}\\
 & VR-Goggles & Parallel policy learning and transfer learning & \cite{zhang2019vr}\\
 & SAC & Incremental environment complexity  & \cite{chaffre2020sim}\\
 & MAPPO/CPD & Allocation of different levels of randomization & \cite{candela2022transferring,kadokawa2022cyclic}\\
 & BayRn & Different parameter distributions at the source level & \cite{muratore2021data}\\
 & RL$^2$ & Encoding the RL algorithm into the weights of a RNN & \cite{duan2016rl}\\
 & MAML & Model agnostic meta-learning for fast domain adaptation & \cite{finn2017model}\\
 & RWPL & Dimensionality reduction for domain adaptation & \cite{lonvcarcvic2021accelerated}\\
 & SimGAN & Adversarial domain adaptation mechanism & \cite{jiang2021simgan}\\
 & MPBO & Combination of domain randomization and adaptation & \cite{exarchos2021policy}\\
 & SRL & Domain adaptation for visual grasping tasks & \cite{park2021sim}\\
Policy Transfer & ATL & Learning from adaptation and exploration & \cite{joshi2021adaptive} \\
 & SEPT & Single episode policy transfer & \cite{yang2019single}\\
 & PTPM & Multiple knowledge policy transfer & \cite{fernandez2021probabilistic}\\
 & MULTIPOLAR & Knowledge transfer without access to \\
 & & the source environments & \cite{barekatain2019multipolar}\\
 & PTF-A3C & Direct optimization of the target policy & \cite{yang2020efficient}\\
 & MALT & Transferring features instead of policies or experiences & \cite{shi2021lateral}\\
 & DQN & Mixture of experts for tasks with different dynamics & \cite{gimelfarb2021contextual}\\
\hline
\end{tabular}
\end{table*}

\section{Inverse Reinforcement Learning}\label{sec4}
This section reviews the developed techniques for improving the sample-efficiency and generalization of IRL models. 

\subsection{Multi-Intention IRL}
In contrast to the typical IRL problem, where the ultimate goal is to recover a reward function from the given set of demonstrations generated by a single expert, multi-intention IRL is a more general scheme that deals with demonstrations coming from multiple experts, and the aim is recovering multiple reward functions. Concepts such as multi-task and multi-agent IRL could also fall under the multi-intention category of IRL techniques due to the fact that the general purpose is to recover reward functions by accessing demonstrations from distinct sources \cite{9525046}.

One of the earliest attempts toward multi-intention IRL is the work of Dimitrakakis and Rothkopf \cite{dimitrakakis2011bayesian}, in which the LfD problem is formulated as a multi-task learning problem and Bayesian IRL technique \cite{ramachandran2007bayesian} is employed under the assumption that each observed trajectory conducts a single reward function. Later in \cite{babes2011apprenticeship}, the authors stepped forward and proposed a novel viewpoint on multi-intention frameworks by suggesting to formulate the problem as a clustering task by means of IRL. The general idea was to deduce the reward functions by clustering the given trajectories, where the reward parameters could then be estimated using a maximum likelihood objective function. However, the limitation of this technique was that the number of clusters is required to be known \textit{a priori}. To remove this limitation, the authors in \cite{choi2012nonparametric} proposed to rely on a non-parametric Bayesian IRL to recover an unknown number of reward functions. %Under this structure, recent multi-intention works could be categorized into model-based \cite{nguyen2015inverse} and model-free \cite{lin2018acgail} categories.

More recently, by resorting to the maximum entropy IRL, a deep and adaptive model is proposed in \cite{bighashdel2021deep} to address the multi-intention IRL problem. The reward function is assumed to be non-linear and it is modeled by a deep network that includes a base reward model and intention-specific reward models. The multi-intention behaviors in the given trajectories are modeled as intention distributions utilizing a conditional maximum entropy approach and the reward network is adaptively updated via the expectation-maximization (EM) algorithm. However, the EM algorithm is known to be prone to get stuck in local minima, where a good initialization could be of utmost importance to deal with this issue \cite{balakrishnan2017statistical}. To this end, the authors in \cite{snoswell2021limiirl} proposed a lightweight multi-intent IRL (LiMIIRL) framework to effectively initialize EM by clustering the given demonstrations. For a more realistic scenario, called \textit{batch model-free}, in which the only available information to the agent is the given demonstrations and there are no further interactions with the environment, the authors in \cite{ramponi2020truly} proposed the $\Sigma$-Gradient IRL ($\Sigma$-GIRL) and extended it to the multi-intention IRL. The given experts are clustered by resorting to the likelihood model of $\Sigma$-GIRL and the reward parameters are recovered by the EM algorithm. %Different from clustering the given demonstrations, the choice of demonstrations could be parameterized and augmented in the multi-intention objective function through a factual vector sparsity to optimize the intention dictionary \cite{piao2019online}.

Concerning multi-agent IRL problems, it is well-studied \cite{bergerson2021multi} that such IRL models can be realized through either collective agent populations \cite{chen2021agent} or individual agents that have access to a shared feature space \cite{filos2021psiphi}, and under cooperative \cite{likmeta2021dealing} or competitive \cite{neumeyer2021general} dynamics.   Even though the multi-agent IRL framework could be of potential benefit for achieving a beneficial social outcome, however, such models need to account for complex rational behaviors. To address this issue, a primary work in the context of generative adversarial networks (GANs) was proposed in \cite{yu2019multi}, which was then extended to a more scalable and sample-efficient framework in \cite{jeon2020scalable} through modeling the reward signals by different discriminators. The use of a teacher-student learning scheme for students to learn the target task via a single demonstration \cite{melo2021teaching}, cooperative learning by considering the behavior of each individual agent \cite{fukumoto2020cooperative}, online IRL (I2RL) for which the demonstrations could be acquired incrementally during the learning process \cite{arora2021i2rl}, and context-aware multi-agent IRL \cite{nishi2020fine} are worthwhile research directions.% under the multi-agent IRL framework.

\subsection{Adversarial IRL}
It is well-studied that GANs benefit from two networks called generator $\mathcal{G}$ and discriminator $\mathcal{D}$ \cite{goodfellow2020generative}. The generator aims at generating new data points by learning the distribution of the input data, while the discriminator tries to distinguish whether a given data point is generated by means of the generator or it is from the real input dataset. This idea has been extended to IRL problems, too. It is basically used to deal with the case, in which a low number of expert trajectories are available. Generators are then trained to imitate the behavior of experts in the given trajectories, and the discriminator serves as the reward function, showing how similar are the generated behaviors to those of the experts.

This category of methods is mostly concerned with augmenting imitation learning using GANs. The idea originated in \cite{ho2016generative}, where the authors proposed a generative adversarial imitation learning (GAIL) algorithm in order to address a shortcoming of IRL. That is, IRL tries to first recover a cost function from which an optimal policy can then be extracted through RL. This is an indirect method of learning the policy which can in turn be slow, especially for complex behaviors in trajectories obtained from high-dimensional environments. GAIL addresses this issue through the augmentation of GANs and imitation learning. In the GAIL framework, an occupancy measure is defined for the learner, shown by $\rho_{\pi}$, which is indeed analogous to that of the generated data distribution by means of $\mathcal{G}$. Furthermore, the true data distribution is replaced with the term expert's occupancy measure. The occupancy measure $\rho_{\pi}:\mathcal{S}\times \mathcal{A}\rightarrow \mathbb{R}$ is defined as $\rho_{\pi}(s,a)=\pi(s|a)\sum_{t=0}^{\infty}\gamma^t\mathbb{P}(s_t=s|\pi)$. Given this occupancy measure, GAIL is concerned with the following optimization problem:
\begin{align}\label{eq20}
\min_{\pi}\hspace{0.25cm}\psi(\rho_{\pi}-\rho_{\pi_E})-\lambda H(\pi),
\end{align}
where $\psi$ is a cost regularizer and $H(\pi)\doteq\mathbb{E}_{\pi}\left[-\log\pi(a|s)\right]$, called $\gamma$-discounted causal entropy \cite{bloem2014infinite}, is a policy regularizer which is controlled by $\lambda$. Obviously, Eq. (\ref{eq20}) aims at learning a generator that imitates the expert policies.

Even though GAIL is known as a breakthrough in IRL, however, it does not attempt to recover the reward function, and it also fails to generalize when the dynamics of the environment are subject to changes. In this regard, the authors in \cite{finn2016connection} suggested that for the IRL problems solved through the maximum entropy technique, the optimization problem could also be viewed as that of a GAN optimization for the case that the discriminator takes on a special form:
\begin{align}\label{eq21}
\mathcal{D}_{\theta}(\mathcal{T})=\frac{e^{f_{\theta}(\mathcal{T})}}{e^{f_{\theta}(\mathcal{T})}+\pi(\mathcal{T})},
\end{align}
where $f_{\theta}(\mathcal{T})$ is a learned function, $\pi(\mathcal{T})$ is pre-computed, and the learned policy aims to maximize $R(\mathcal{T})=\log(1-\mathcal{D}(\mathcal{T})-\log\mathcal{D}(\mathcal{T}))$. Following this formulation, the discriminator enables the learning of the reward function, and updating the discriminator could be thought of as updating the reward function. To address the issue of GAIL with dynamic changes of environment in addition to providing the estimation of the reward function, adversarial IRL (AIRL) was then proposed in \cite{fu2017learning} by reformulation of $f_{\theta}$ in (\ref{eq21}) as below:
\begin{align} 
f_{\theta,\phi}(s,a,s')=g_{\theta}(s,a)+\gamma h_{\phi}(s')-h_{\phi}(s),
\end{align}
where $g_{\theta}$ is the reward approximation and $h_{\phi}$ is a shaping term, which originated from the concept of RS by preserving the optimal policy proposed in \cite{ng1999policy}.  

Recently, some efforts have been devoted to IRL using GANs that are either trying to extend GAIL applications or improving its performance. For instance, OptionGAN \cite{henderson2018optiongan} is built based upon GAIL and options framework in forward RL, where the aim is to simultaneously recover reward and policy options. Such a framework is vital in dealing with demonstrations arising from multiple implicit reward functions instead of just a single reward function. The authors in \cite{yu2019multi} extended the AIRL technique to a multi-agent framework through the concept of maximum entropy in IRL. Following this, the concept of situated GAIL (S-GAIL) was proposed in \cite{kobayashi2019situated}, in which a task variable is involved in the learning process of the generator and discriminator in order to enable the learning of different reward functions and policies gathered from different tasks. To deal with the issue of reward entanglement, which means the recovered reward function through some IRL techniques is highly entangled with the environment dynamics, a robust method called o-IRL is built based on the AIRL in \cite{venuto2020robust}. Later in 2020, the authors in \cite{arnob2020off} proposed off-policy-AIRL, where compared to the original AIRL, it is more sample efficient and can leverage TL into IRL in a way that it shows a better imitation performance. In contrast to GAIL, AIRL, and state-marginal-matching (SMM) \cite{ghasemipour2020divergence} that learn a non-stationary reward that cannot, later on, be used for training a new policy, the authors in \cite{ni2020f} proposed an algorithm called $f$-IRL that is capable of learning a stationary reward function by resorting to the $f$-divergence (w.r.t. reward parameters) between the state distribution of the agent and expert. More recently, to adapt the learned reward function and policy through AIRL to new tasks, meta-learning is integrated into AIRL in an algorithm called Meta-AIRL \cite{wang2021meta}, which is more sample-efficient compared to AIRL. %Moreover, the authors in \cite{chen2022hierarchical} proposed an algorithm called hierarchical AIRL to recover a hierarchical policy from highly-complex and long-horizontal tasks by augmenting the hierarchical imitation learning into AIRL. 

\subsection{Sample-Efficient IRL}
Sample efficiency in IRL refers to training mechanisms, in which agents require a fewer number of interactions with the environment to recover the reward function. On-policy techniques, e.g., AIRL, rely on sampling from the current policy that the agent is being trained on, and on the Monte Carlo estimation, that suffers from high gradient variance and intensive sampling due to discarding the experienced transitions. To this end, a category of IRL methods has been devoted to addressing this challenge through off-policy mechanisms and imitation learning. 

One of the first attempts towards sample-efficient IRL is called sample-efficient adversarial mimic (\textsc{Sam}) \cite{blonde2019sample}, which improves the sample efficiency of GAIL by resorting to sample-efficient AC with experience replay \cite{wang2016sample}. \textsc{Sam} is therefore an off-policy architecture in the context of imitation learning that relies on a replay buffer \cite{lillicrap2015continuous} to deal with the sample efficiency issue of IRL. It involves three modules including a reward module, a critic module, and a policy module, where it interacts with the environment by exploiting the current policy and storing the experienced transitions in a replay buffer, and, then, it updates the reward and critic module and the policy and critic modules using the sampled state-action pairs from the replay buffer and through the bilevel optimization of GANs. Following this, in order to make GAIL and its extensions even more sample-efficient, the authors in \cite{sasaki2018sample} proposed to adopt off-policy AC (Off-PAC) \cite{degris2012off} so as to optimize the learner policy and to reduce the number of agent-environment interactions. Besides, it is suggested to combine the objectives of reward learning and value function approximation in a novel objective that leads to learning the value function without the need to learn the reward function. Omitting the reward learning along with bounding the policy function has made this technique to be significantly sample-efficient. More recently, and in contrast to the proposed strategy in \cite{sasaki2018sample}, the authors in \cite{hoshino2022opirl} proposed an algorithm called off-policy IRL (OPIRL), which not only makes use of off-policy data distribution for the sake of sample efficiency, but also the recovered reward function is generalizable to the environment with changing dynamics. This is achieved thanks to the introduction of causal entropy as a regularization term in the modified off-policy objective function built based on the on-policy objective function of the AIRL. Soft Q-functioned meta IRL (SQUIRL) \cite{wu2020squirl} is also another recently-developed algorithm for efficient learning from video demonstrations using off-policy training of robots for manipulation tasks. 

Other than moving from on-policy to off-policy learning, some other application-based techniques have also been devised to deal with the sample-efficiency issue of IRL. In this regard, a technique called sampling-based maximum-entropy IRL (SMIRL) is proposed in \cite{wu2020efficient} for autonomous driving, where the sample efficiency is enabled through a path sampler. The sampler does perform path sampling to discover collision-free paths, and speed sampling to find time-optimal speed samples. Soft actor-critic (SAC) \cite{haarnoja2018soft} and discriminator-actor-critic (DAC) \cite{kostrikov2018discriminator}, two well-known sample-efficient RL algorithms, are employed in \cite{baghi2021sample} to construct ReplayIRL algorithm to deal with sample-efficiency of IRL in social navigation applications. Table \ref{tab2} summarizes the recently-developed IRL algorithms.%In the same vein, by replacing the probabilistic transition dynamics and probabilistic policy with deterministic ones, the authors in \cite{konar2021learning} proposed a sample-efficient algorithm called deterministic maximum entropy IRL for the social navigation application. 

\begin{table*}
\scriptsize
\caption{An overview and comparison of recent IRL algorithms.}\label{tab2}
\begin{tabular}{llll}
\hline Category & Learner/Algorithm & Notable Features & Citation\\
\hline Multi-Intention IRL & LiMIIRL & Warm-start strategy using clustering & \cite{snoswell2021limiirl,babes2011apprenticeship}\\
 & MI $\Sigma$-GIRL & Extension of $\Sigma$-GIRL to multi-intention tasks & \cite{ramponi2020truly}\\
 & MaxEnt MIRL & Multi-intention and multi-agent IRL & \cite{bergerson2021multi}\\
 & MF-AIRL & Extension of AIRL to the multi-intention tasks & \cite{chen2021agent,yu2019multi}\\
 & PsiPhi & Inverse temporal difference learning & \cite{filos2021psiphi}\\
 & GS-CIOC & Continuous inverse optimal control for \\
 & & multi-agent setting & \cite{neumeyer2021general}\\ 
 & AAC & Sample efficient and scalable IRL & \cite{jeon2020scalable}\\
 & I2RL & Common ground for online IRL & \cite{arora2021i2rl}\\
 & BIRL & Bayesian IRL & \cite{ramachandran2007bayesian}\\
 & NBIRL & Non-parametric BIRL & \cite{choi2012nonparametric}\\
 & ACGAIL & Auxiliary classification for label conditioning in GAIL & \cite{lin2018acgail}\\
Adversarial IRL & GAIL & Direct extraction of policy from data & \cite{ho2016generative}\\
 & MaxEntropy & Connecting GANs, IRL, and energy-based models & \cite{finn2016connection}\\
 & AIRL & Recovery of robust rewards against changes in dynamics & \cite{fu2017learning}\\
 & OptionGAN & Connecting options framework in RL with GANs & \cite{henderson2018optiongan}\\
 & Situated GAIL & Extending GAIL to multi-task frameworks & \cite{kobayashi2019situated}\\
 & $f$-MAX & $f$-divergence extension of AIRL & \cite{ghasemipour2020divergence}\\
 & $f$-IRL & Recovery of stationary rewards using gradient descent & \cite{ni2020f}\\
 & Meta-AIRL & Adaptable imitation learning using \\
 & & meta-learning and AIRL & \cite{wang2021meta}\\
 & HAIRL & Hierarchical framework based on AIRL & \cite{chen2022hierarchical}\\
Sample-Efficient IRL & SAM & Sample efficiency by leveraging \\
& & an off-policy AC architecture & \cite{blonde2019sample}\\
 & DDPG & Sample efficiency by improving the replay buffer & \cite{lillicrap2015continuous}\\
 & OPIRL & Recovering a generalizable reward w.r.t. \\
 & & changes in dynamics &  \cite{hoshino2022opirl}\\
 & SQUIRL & Sample efficient meta-learning for IRL & \cite{wu2020squirl}\\
 & SMIRL & Path sampling for autonomous driving & \cite{wu2020efficient}\\
 & Replay IRL & Combining SAC and DAC for sample efficiency of IRL & \cite{baghi2021sample}\\
\hline
\end{tabular}
\end{table*}

\section{T-IRL: Applications and Open Problems}\label{sec5}
The literature review of T-IRL signifies their encouraging results in improving the learning efficiency of RL algorithms that have paved the way of their applications in various sectors. Among them, the \textit{cybersecurity} of cyber-physical systems (CPSs) has been widely paid attention to in recent years. Due to the distributed setup of CPSs, the concept of TRL has shown promising features that fit the nature of CPSs, where the learned knowledge of local agents could be shared with, and transferred to other agents for securing the global system against jamming and denial-of-service cyber attacks, and stealthily compromising the sensor readings \cite{8742593,9705557,HU2021106967,9170648}. On the contrary, IRL models study the situation that both the learner and expert suffer from adversarial attacks in learning process and the aim is to design an agent that bypasses these attacks and avoids undesired and hazardous states \cite{9844128,9565156,elnaggar2018irl,chen2021protecting}. T-IRL has also been widely employed in self-driving or \textit{autonomous driving} (AD) systems \cite{HASSANI2025128836, HASSANI2024109147}. AD systems constantly deal with a sequential decision-making task for trajectory optimization and motion planning, where the classical SL models are no longer applicable due to the changes in the configuration of the environment and the involved uncertainties in the supervisory signals. To account for these shortcomings, T-IRL has offered promising solutions in each component of the AD systems including scene understanding \cite{he2022irlsot,DING2019243}, mapping and localization \cite{9889241}, driving policies and planning \cite{YOU20191}, and control \cite{gao2018car,9583858}. Besides, TRL has recently been a hot research topic in the NLP field for constructing \textit{large language models} (LLMs). In line with HITL, RL from human feedback (RLHF) \cite{christiano2017deep} has shown encouraging results in LLMs. The most recent model is the ChatGPT that has gained much attention in the NLP community. ChatGPT, which is a chatbot, is built based upon the InstructGPT \cite{ouyang2022training}, a recently developed model for fine-tuning the generative pre-trained transformer (GPT)-3 \cite{brown2020language} by leveraging the human feedback through RL for fine-tuning the supervised model trained on the labeler demonstrations of the desired model behaviour. 

The sequential tasks for employing an RL agent could be considered as pre-training the agent on the simulator, deploying the simulated agent in the real-world environment, and fine-tuning its parameters w.r.t. dynamics of the target domain. This sequence of actions, other than being time-consuming, suffers from being sample-inefficient. One way to deal with this limitation is through the multi-agent framework by splitting the sample observations between several agents. However, transferring the collected samples and the learned knowledge from an agent to a central model through a communication regime could be costly and prone to data privacy issues. Federated learning (FL) \cite{mcmahan2017communication,konevcny2016federated}, could be considered as one of the recently-emerged solutions to the aforementioned challenges, which could be efficiently augmented into the RL framework, known as federated RL (FRL) \cite{qi2021federated}. FRL is the collaborative training of a global agent through communication with multiple decentralized agents working on their local environments without concerning a centralized training data \cite{khodadadian2022federated}. Primary FRL results in addressing the sample-efficiency issue of RL are encouraging and there exists much room for improving and extending them to DRL \cite{luo2022federated,rezazadeh2022federated}, TRL \cite{liang2023federated}, and IRL frameworks \cite{banerjee2021identity}.

Another efficient learning scheme for RL applications is curriculum learning. Curriculum RL (CRL) is to gradually increase the difficulty of the learning task from the easiest to the most difficult one through the control of initial states of the environment or the environmental dynamics in order to speed up the learning process \cite{zhao2022self}. Even though encouraging frameworks have been proposed under the concept of CRL to improve RL through teacher-student learning \cite{schraner2022teacher}, domain adaptation \cite{huang2022curriculum}, HITL \cite{zeng2022human}, state dropout \cite{khaitan2022state}, and RS \cite{anca2022effects}, yet these models are mainly centred on the single-agent RL. However, problems such as how to control the number of agents and how the number of tasks could be changed in a multi-agent framework have remained unstudied. 

The learning process in RL could also be accelerated by prioritizing the replay buffer, aiming at prioritizing experience samples in the replay buffer, and choosing the most informative ones to improve the learning process. Some efforts have been devoted to prioritizing the replay buffer based on adaptively changing its size \cite{liu2018effects}, schemes to quickly forget less useful experiences \cite{nguyen2018experience}, and replacing mechanisms to replace useless transitions with other informative transition candidates \cite{bu2020double}. To choose proper experiences, most techniques rely on defining a score on a specific criterion using the TD error \cite{schaul2015prioritized}, entropy-based scaling \cite{ramicic2017entropy}, reward \cite{nguyen2019hindsight}, density estimation-based approach \cite{zhao2019curiosity}, and the likelihood of experience transitions \cite{sinha2022experience}. Besides the fact that most of the developed models are only concerned with a single criterion and a more efficient scheme could be built based on multiple criteria to combine the advantages of all the available criteria \cite{liu2022prioritized,9956746}, the prioritization concept could also be extended to the T-IRL to deal with the sample-efficiency. 

One of the critically-important aspects in the development of RL agents involves ensuring their safe and ethical conduct in real-world settings. This is of paramount importance for mitigating potential hazards associated with deploying RL agents in complex and ever-evolving environments. Given that current RL endeavors strive to construct autonomous systems capable of making instantaneous decisions, however, actions taken by these agents in unforeseen states could yield harmful outcomes, both in terms of safety and ethics. Therefore, a valuable research direction is to regulate the behavior of RL agents within acceptable boundaries. One promising approach could involve leveraging generative AI to expose vulnerabilities in the learned policy by an RL agent to better understand the consequences of its actions. Additionally, developing methodologies that incorporate constraint optimization, uncertainty estimation, and value alignment into the learning framework of an Rl could be beneficial for ensuring compliance with safety and ethical standards.

\section{Conclusion}\label{sec6}
Reinforcement learning, as a sub-domain of machine learning, refers to training an agent towards learning how to act in an environment based on the states of the environment with the hope of maximizing an expected long-term reward. Learning a reward function is, however, a laborious task and it is often difficult to end up with an explicit reward function that satisfies several desiderata of the under-study problem. Further to this, the trial-and-error search scheme in reinforcement learning involves a considerable amount of interactions between the agent and environment to experience various situations. Both of these aforementioned issues make reinforcement learning to be known for suffering from sample efficiency and generalization. To this end, in this survey, we reviewed the most recent advancements under the transfer and inverse reinforcement learning scheme to address the sample efficiency and generalization issues with reinforcement learning. We gave a brief introduction to reinforcement learning, which was followed by a comprehensive review of the fundamental methods of transfer and inverse reinforcement learning. We then delved into each category of methods and discussed what are the most recent advancements in each field of research following a comprehensive review of underlying methods. Our finding denoted that, under the transfer reinforcement learning scheme, a majority of works are concerned with sample efficiency by making use of human-in-the-loop and sim-to-real strategies to efficiently transfer the learned knowledge in the source domain(s) to the target domain. Under inverse reinforcement learning, the developed models are mainly centred on training when a low number of training samples are available and extending such frameworks to multi-intention situation for recovering multiple reward functions.

Even though the attained results of developed models under the transfer and inverse reinforcement learning framework are encouraging, however, there still exist several research gaps that could be considered as future research directions in this field of study. The problem of federated learning for multi-agent frameworks, improving the learning process through curriculum learning, and prioritizing the replay buffer could be of paramount importance to address the sample efficiency and time-consuming training problems in reinforcement learning.

\bibliographystyle{cas-model2-names}
\bibliography{cas-refs}

\begin{thebibliography}{208}
\expandafter\ifx\csname natexlab\endcsname\relax\def\natexlab#1{#1}\fi
\providecommand{\url}[1]{\texttt{#1}}
\providecommand{\href}[2]{#2}
\providecommand{\path}[1]{#1}
\providecommand{\DOIprefix}{doi:}
\providecommand{\ArXivprefix}{arXiv:}
\providecommand{\URLprefix}{URL: }
\providecommand{\Pubmedprefix}{pmid:}
\providecommand{\doi}[1]{\href{http://dx.doi.org/#1}{\path{#1}}}
\providecommand{\Pubmed}[1]{\href{pmid:#1}{\path{#1}}}
\providecommand{\bibinfo}[2]{#2}
\ifx\xfnm\relax \def\xfnm[#1]{\unskip,\space#1}\fi
%Type = Inproceedings
\bibitem[{Abbeel and Ng(2004)}]{abbeel2004apprenticeship}
\bibinfo{author}{Abbeel, P.}, \bibinfo{author}{Ng, A.Y.}, \bibinfo{year}{2004}.
\newblock \bibinfo{title}{Apprenticeship learning via inverse reinforcement
  learning}, in: \bibinfo{booktitle}{Proc. Twenty-First Int. Conf. Mach.
  Learn.}, p.~\bibinfo{pages}{1}.
%Type = Article
\bibitem[{Abel et~al.(2017)Abel, Salvatier, Stuhlm{\"u}ller and
  Evans}]{abel2017agent}
\bibinfo{author}{Abel, D.}, \bibinfo{author}{Salvatier, J.},
  \bibinfo{author}{Stuhlm{\"u}ller, A.}, \bibinfo{author}{Evans, O.},
  \bibinfo{year}{2017}.
\newblock \bibinfo{title}{Agent-agnostic human-in-the-loop reinforcement
  learning}.
\newblock \bibinfo{journal}{arXiv preprint arXiv:1701.04079} .
%Type = Article
\bibitem[{Adams et~al.(2022)Adams, Cody and Beling}]{adams2022survey}
\bibinfo{author}{Adams, S.}, \bibinfo{author}{Cody, T.},
  \bibinfo{author}{Beling, P.A.}, \bibinfo{year}{2022}.
\newblock \bibinfo{title}{A survey of inverse reinforcement learning}.
\newblock \bibinfo{journal}{Artif. Intell. Rev.} , \bibinfo{pages}{1--40}.
%Type = Article
\bibitem[{Alamdari et~al.(2020)Alamdari, Lobarinas and
  Kehtarnavaz}]{alamdari2020personalization}
\bibinfo{author}{Alamdari, N.}, \bibinfo{author}{Lobarinas, E.},
  \bibinfo{author}{Kehtarnavaz, N.}, \bibinfo{year}{2020}.
\newblock \bibinfo{title}{Personalization of hearing aid compression by
  human-in-the-loop deep reinforcement learning}.
\newblock \bibinfo{journal}{IEEE Access} \bibinfo{volume}{8},
  \bibinfo{pages}{203503--203515}.
%Type = Article
\bibitem[{Anca et~al.(2022)Anca, Studley, Hansen, Thomas and
  Pedamonti}]{anca2022effects}
\bibinfo{author}{Anca, M.}, \bibinfo{author}{Studley, M.},
  \bibinfo{author}{Hansen, M.}, \bibinfo{author}{Thomas, J.D.},
  \bibinfo{author}{Pedamonti, D.}, \bibinfo{year}{2022}.
\newblock \bibinfo{title}{Effects of reward shaping on curriculum learning in
  goal conditioned tasks}.
\newblock \bibinfo{journal}{arXiv preprint arXiv:2206.02462} .
%Type = Article
\bibitem[{Arakawa et~al.(2018)Arakawa, Kobayashi, Unno, Tsuboi and
  Maeda}]{arakawa2018dqn}
\bibinfo{author}{Arakawa, R.}, \bibinfo{author}{Kobayashi, S.},
  \bibinfo{author}{Unno, Y.}, \bibinfo{author}{Tsuboi, Y.},
  \bibinfo{author}{Maeda, S.i.}, \bibinfo{year}{2018}.
\newblock \bibinfo{title}{{DQN-TAMER}: Human-in-the-loop reinforcement learning
  with intractable feedback}.
\newblock \bibinfo{journal}{arXiv preprint arXiv:1810.11748} .
%Type = Inproceedings
\bibitem[{Arndt et~al.(2020)Arndt, Hazara, Ghadirzadeh and
  Kyrki}]{arndt2020meta}
\bibinfo{author}{Arndt, K.}, \bibinfo{author}{Hazara, M.},
  \bibinfo{author}{Ghadirzadeh, A.}, \bibinfo{author}{Kyrki, V.},
  \bibinfo{year}{2020}.
\newblock \bibinfo{title}{Meta reinforcement learning for sim-to-real domain
  adaptation}, in: \bibinfo{booktitle}{IEEE Int. Conf. Robot. Autom.},
  \bibinfo{organization}{IEEE}. pp. \bibinfo{pages}{2725--2731}.
%Type = Article
\bibitem[{Arnob(2020)}]{arnob2020off}
\bibinfo{author}{Arnob, S.Y.}, \bibinfo{year}{2020}.
\newblock \bibinfo{title}{Off-policy adversarial inverse reinforcement
  learning}.
\newblock \bibinfo{journal}{arXiv preprint arXiv:2005.01138} .
%Type = Article
\bibitem[{Arora and Doshi(2021)}]{ARORA2021103500}
\bibinfo{author}{Arora, S.}, \bibinfo{author}{Doshi, P.}, \bibinfo{year}{2021}.
\newblock \bibinfo{title}{A survey of inverse reinforcement learning:
  Challenges, methods and progress}.
\newblock \bibinfo{journal}{Artif. Intell.} \bibinfo{volume}{297},
  \bibinfo{pages}{103500}.
%Type = Article
\bibitem[{Arora et~al.(2021)Arora, Doshi and Banerjee}]{arora2021i2rl}
\bibinfo{author}{Arora, S.}, \bibinfo{author}{Doshi, P.},
  \bibinfo{author}{Banerjee, B.}, \bibinfo{year}{2021}.
\newblock \bibinfo{title}{{I2RL}: online inverse reinforcement learning under
  occlusion}.
\newblock \bibinfo{journal}{Auton. Agents Multi-Agent Syst.}
  \bibinfo{volume}{35}, \bibinfo{pages}{1--33}.
%Type = Article
\bibitem[{Auer et~al.(2008)Auer, Jaksch and Ortner}]{auer2008near}
\bibinfo{author}{Auer, P.}, \bibinfo{author}{Jaksch, T.},
  \bibinfo{author}{Ortner, R.}, \bibinfo{year}{2008}.
\newblock \bibinfo{title}{Near-optimal regret bounds for reinforcement
  learning}.
\newblock \bibinfo{journal}{NeurIPS} \bibinfo{volume}{21}.
%Type = Inproceedings
\bibitem[{Azar et~al.(2017)Azar, Osband and Munos}]{azar2017minimax}
\bibinfo{author}{Azar, M.G.}, \bibinfo{author}{Osband, I.},
  \bibinfo{author}{Munos, R.}, \bibinfo{year}{2017}.
\newblock \bibinfo{title}{Minimax regret bounds for reinforcement learning},
  in: \bibinfo{booktitle}{ICML}, \bibinfo{organization}{PMLR}. pp.
  \bibinfo{pages}{263--272}.
%Type = Inproceedings
\bibitem[{Babes et~al.(2011)Babes, Marivate, Subramanian and
  Littman}]{babes2011apprenticeship}
\bibinfo{author}{Babes, M.}, \bibinfo{author}{Marivate, V.},
  \bibinfo{author}{Subramanian, K.}, \bibinfo{author}{Littman, M.L.},
  \bibinfo{year}{2011}.
\newblock \bibinfo{title}{Apprenticeship learning about multiple intentions},
  in: \bibinfo{booktitle}{Proc. 28th Int. Conf. Mach. Learn.}, pp.
  \bibinfo{pages}{897--904}.
%Type = Inproceedings
\bibitem[{Badnava et~al.(2023)Badnava, Esmaeili, Mozayani and
  Zarkesh-Ha}]{badnava2023new}
\bibinfo{author}{Badnava, B.}, \bibinfo{author}{Esmaeili, M.},
  \bibinfo{author}{Mozayani, N.}, \bibinfo{author}{Zarkesh-Ha, P.},
  \bibinfo{year}{2023}.
\newblock \bibinfo{title}{A new potential-based reward shaping for
  reinforcement learning agent}, in: \bibinfo{booktitle}{CCWC},
  \bibinfo{organization}{IEEE}. pp. \bibinfo{pages}{01--06}.
%Type = Article
\bibitem[{Baghi and Dudek(2021)}]{baghi2021sample}
\bibinfo{author}{Baghi, B.H.}, \bibinfo{author}{Dudek, G.},
  \bibinfo{year}{2021}.
\newblock \bibinfo{title}{Sample efficient social navigation using inverse
  reinforcement learning}.
\newblock \bibinfo{journal}{arXiv preprint arXiv:2106.10318} .
%Type = Article
\bibitem[{Bahrami et~al.(2021)Bahrami, Dornaika and
  Bosaghzadeh}]{bahrami2021joint}
\bibinfo{author}{Bahrami, S.}, \bibinfo{author}{Dornaika, F.},
  \bibinfo{author}{Bosaghzadeh, A.}, \bibinfo{year}{2021}.
\newblock \bibinfo{title}{Joint auto-weighted graph fusion and scalable
  semi-supervised learning}.
\newblock \bibinfo{journal}{Inf. Fusion} \bibinfo{volume}{66},
  \bibinfo{pages}{213--228}.
%Type = Article
\bibitem[{Balakrishnan et~al.(2017)Balakrishnan, Wainwright and
  Yu}]{balakrishnan2017statistical}
\bibinfo{author}{Balakrishnan, S.}, \bibinfo{author}{Wainwright, M.J.},
  \bibinfo{author}{Yu, B.}, \bibinfo{year}{2017}.
\newblock \bibinfo{title}{Statistical guarantees for the em algorithm: From
  population to sample-based analysis}.
\newblock \bibinfo{journal}{Ann. Stat.} \bibinfo{volume}{45},
  \bibinfo{pages}{77--120}.
%Type = Inproceedings
\bibitem[{Banerjee et~al.(2021)Banerjee, Bouzefrane and
  Abane}]{banerjee2021identity}
\bibinfo{author}{Banerjee, S.}, \bibinfo{author}{Bouzefrane, S.},
  \bibinfo{author}{Abane, A.}, \bibinfo{year}{2021}.
\newblock \bibinfo{title}{Identity management with hybrid blockchain approach:
  A deliberate extension with federated-inverse-reinforcement learning}, in:
  \bibinfo{booktitle}{IEEE 22nd Int. Conf. High Perform. Switching Routing
  (HPSR)}, \bibinfo{organization}{IEEE}. pp. \bibinfo{pages}{1--6}.
%Type = Article
\bibitem[{Barekatain et~al.(2019)Barekatain, Yonetani and
  Hamaya}]{barekatain2019multipolar}
\bibinfo{author}{Barekatain, M.}, \bibinfo{author}{Yonetani, R.},
  \bibinfo{author}{Hamaya, M.}, \bibinfo{year}{2019}.
\newblock \bibinfo{title}{Multipolar: Multi-source policy aggregation for
  transfer reinforcement learning between diverse environmental dynamics}.
\newblock \bibinfo{journal}{arXiv preprint arXiv:1909.13111} .
%Type = Article
\bibitem[{Barto et~al.(1983)Barto, Sutton and Anderson}]{barto1983neuronlike}
\bibinfo{author}{Barto, A.G.}, \bibinfo{author}{Sutton, R.S.},
  \bibinfo{author}{Anderson, C.W.}, \bibinfo{year}{1983}.
\newblock \bibinfo{title}{Neuronlike adaptive elements that can solve difficult
  learning control problems}.
\newblock \bibinfo{journal}{IEEE Trans. Syst., Man, Cybern.}
  \bibinfo{volume}{SMC-13}, \bibinfo{pages}{834--846}.
%Type = Article
\bibitem[{Bergerson(2021)}]{bergerson2021multi}
\bibinfo{author}{Bergerson, S.}, \bibinfo{year}{2021}.
\newblock \bibinfo{title}{Multi-agent inverse reinforcement learning:
  Suboptimal demonstrations and alternative solution concepts}.
\newblock \bibinfo{journal}{arXiv preprint arXiv:2109.01178} .
%Type = Article
\bibitem[{Bi et~al.(2021)Bi, Sferrazza and D’Andrea}]{bi2021zero}
\bibinfo{author}{Bi, T.}, \bibinfo{author}{Sferrazza, C.},
  \bibinfo{author}{D’Andrea, R.}, \bibinfo{year}{2021}.
\newblock \bibinfo{title}{Zero-shot sim-to-real transfer of tactile control
  policies for aggressive swing-up manipulation}.
\newblock \bibinfo{journal}{IEEE Robot. Autom. Lett.} \bibinfo{volume}{6},
  \bibinfo{pages}{5761--5768}.
%Type = Inproceedings
\bibitem[{Bighashdel et~al.(2021)Bighashdel, Meletis, Jancura and
  Dubbelman}]{bighashdel2021deep}
\bibinfo{author}{Bighashdel, A.}, \bibinfo{author}{Meletis, P.},
  \bibinfo{author}{Jancura, P.}, \bibinfo{author}{Dubbelman, G.},
  \bibinfo{year}{2021}.
\newblock \bibinfo{title}{Deep adaptive multi-intention inverse reinforcement
  learning}, in: \bibinfo{booktitle}{Joint Eur. Conf. Mach. Learn. Knowl. Disc.
  Databases}, \bibinfo{organization}{Springer}. pp. \bibinfo{pages}{206--221}.
%Type = Inproceedings
\bibitem[{Bloem and Bambos(2014)}]{bloem2014infinite}
\bibinfo{author}{Bloem, M.}, \bibinfo{author}{Bambos, N.},
  \bibinfo{year}{2014}.
\newblock \bibinfo{title}{Infinite time horizon maximum causal entropy inverse
  reinforcement learning}, in: \bibinfo{booktitle}{53rd IEEE Conf. Dec.
  Control}, \bibinfo{organization}{IEEE}. pp. \bibinfo{pages}{4911--4916}.
%Type = Inproceedings
\bibitem[{Blond{\'e} and Kalousis(2019)}]{blonde2019sample}
\bibinfo{author}{Blond{\'e}, L.}, \bibinfo{author}{Kalousis, A.},
  \bibinfo{year}{2019}.
\newblock \bibinfo{title}{Sample-efficient imitation learning via generative
  adversarial nets}, in: \bibinfo{booktitle}{22nd Int. Conf. Artif. Intell.
  Stat.}, \bibinfo{organization}{PMLR}. pp. \bibinfo{pages}{3138--3148}.
%Type = Article
\bibitem[{Brafman and Tennenholtz(2002)}]{brafman2002r}
\bibinfo{author}{Brafman, R.I.}, \bibinfo{author}{Tennenholtz, M.},
  \bibinfo{year}{2002}.
\newblock \bibinfo{title}{R-max-a general polynomial time algorithm for
  near-optimal reinforcement learning}.
\newblock \bibinfo{journal}{J. Mach. Learn. Res.} \bibinfo{volume}{3},
  \bibinfo{pages}{213--231}.
%Type = Article
\bibitem[{Brown et~al.(2020)Brown, Mann, Ryder, Subbiah, Kaplan, Dhariwal,
  Neelakantan, Shyam, Sastry, Askell et~al.}]{brown2020language}
\bibinfo{author}{Brown, T.}, \bibinfo{author}{Mann, B.},
  \bibinfo{author}{Ryder, N.}, \bibinfo{author}{Subbiah, M.},
  \bibinfo{author}{Kaplan, J.D.}, \bibinfo{author}{Dhariwal, P.},
  \bibinfo{author}{Neelakantan, A.}, \bibinfo{author}{Shyam, P.},
  \bibinfo{author}{Sastry, G.}, \bibinfo{author}{Askell, A.}, et~al.,
  \bibinfo{year}{2020}.
\newblock \bibinfo{title}{Language models are few-shot learners}.
\newblock \bibinfo{journal}{Adv. Neural Inf. Process. Syst.}
  \bibinfo{volume}{33}, \bibinfo{pages}{1877--1901}.
%Type = Inproceedings
\bibitem[{Bu and Chang(2020)}]{bu2020double}
\bibinfo{author}{Bu, F.}, \bibinfo{author}{Chang, D.E.}, \bibinfo{year}{2020}.
\newblock \bibinfo{title}{Double prioritized state recycled experience replay},
  in: \bibinfo{booktitle}{IEEE Int. Conf. Consumer Electron.-Asia},
  \bibinfo{organization}{IEEE}. pp. \bibinfo{pages}{1--6}.
%Type = Inproceedings
\bibitem[{B{\"u}ning et~al.(2022)B{\"u}ning, George and
  Dimitrakakis}]{buning2022interactive}
\bibinfo{author}{B{\"u}ning, T.K.}, \bibinfo{author}{George, A.M.},
  \bibinfo{author}{Dimitrakakis, C.}, \bibinfo{year}{2022}.
\newblock \bibinfo{title}{Interactive inverse reinforcement learning for
  cooperative games}, in: \bibinfo{booktitle}{ICML},
  \bibinfo{organization}{PMLR}. pp. \bibinfo{pages}{2393--2413}.
%Type = Article
\bibitem[{Candela et~al.(2023)Candela, Doustaly, Parada, Feng, Demiris and
  Angeloudis}]{candela2023risk}
\bibinfo{author}{Candela, E.}, \bibinfo{author}{Doustaly, O.},
  \bibinfo{author}{Parada, L.}, \bibinfo{author}{Feng, F.},
  \bibinfo{author}{Demiris, Y.}, \bibinfo{author}{Angeloudis, P.},
  \bibinfo{year}{2023}.
\newblock \bibinfo{title}{Risk-aware controller for autonomous vehicles using
  model-based collision prediction and reinforcement learning}.
\newblock \bibinfo{journal}{Artif. Intell.} \bibinfo{volume}{320},
  \bibinfo{pages}{103923}.
%Type = Article
\bibitem[{Candela et~al.(2022)Candela, Parada, Marques, Georgescu, Demiris and
  Angeloudis}]{candela2022transferring}
\bibinfo{author}{Candela, E.}, \bibinfo{author}{Parada, L.},
  \bibinfo{author}{Marques, L.}, \bibinfo{author}{Georgescu, T.A.},
  \bibinfo{author}{Demiris, Y.}, \bibinfo{author}{Angeloudis, P.},
  \bibinfo{year}{2022}.
\newblock \bibinfo{title}{Transferring multi-agent reinforcement learning
  policies for autonomous driving using sim-to-real}.
\newblock \bibinfo{journal}{arXiv preprint arXiv:2203.11653} .
%Type = Article
\bibitem[{Cao and Xie(2022)}]{9715172}
\bibinfo{author}{Cao, K.}, \bibinfo{author}{Xie, L.}, \bibinfo{year}{2022}.
\newblock \bibinfo{title}{Game-theoretic inverse reinforcement learning: A
  differential pontryagin's maximum principle approach}.
\newblock \bibinfo{journal}{IEEE Trans. Neural Netw. Learn. Syst.} ,
  \bibinfo{pages}{1--8}\DOIprefix\doi{10.1109/TNNLS.2022.3148376}.
%Type = Article
\bibitem[{Chaffre et~al.(2020)Chaffre, Moras, Chan-Hon-Tong and
  Marzat}]{chaffre2020sim}
\bibinfo{author}{Chaffre, T.}, \bibinfo{author}{Moras, J.},
  \bibinfo{author}{Chan-Hon-Tong, A.}, \bibinfo{author}{Marzat, J.},
  \bibinfo{year}{2020}.
\newblock \bibinfo{title}{Sim-to-real transfer with incremental environment
  complexity for reinforcement learning of depth-based robot navigation}.
\newblock \bibinfo{journal}{arXiv preprint arXiv:2004.14684} .
%Type = Article
\bibitem[{Chai et~al.(2021)Chai, Li, Zhu, Zhao, Ma, Sun and Ding}]{9525046}
\bibinfo{author}{Chai, J.}, \bibinfo{author}{Li, W.}, \bibinfo{author}{Zhu,
  Y.}, \bibinfo{author}{Zhao, D.}, \bibinfo{author}{Ma, Z.},
  \bibinfo{author}{Sun, K.}, \bibinfo{author}{Ding, J.}, \bibinfo{year}{2021}.
\newblock \bibinfo{title}{Unmas: Multiagent reinforcement learning for unshaped
  cooperative scenarios}.
\newblock \bibinfo{journal}{IEEE Trans. Neural Netw. Learn. Syst.} ,
  \bibinfo{pages}{1--12}\DOIprefix\doi{10.1109/TNNLS.2021.3105869}.
%Type = Article
\bibitem[{Chen et~al.(2022a)Chen, Lan and Aggarwal}]{chen2022hierarchical}
\bibinfo{author}{Chen, J.}, \bibinfo{author}{Lan, T.},
  \bibinfo{author}{Aggarwal, V.}, \bibinfo{year}{2022}a.
\newblock \bibinfo{title}{Hierarchical adversarial inverse reinforcement
  learning}.
\newblock \bibinfo{journal}{arXiv preprint arXiv:2210.01969} .
%Type = Article
\bibitem[{Chen et~al.(2022b)Chen, Gao, Reddy, Berseth, Dragan and
  Levine}]{chen2022asha}
\bibinfo{author}{Chen, S.}, \bibinfo{author}{Gao, J.}, \bibinfo{author}{Reddy,
  S.}, \bibinfo{author}{Berseth, G.}, \bibinfo{author}{Dragan, A.D.},
  \bibinfo{author}{Levine, S.}, \bibinfo{year}{2022}b.
\newblock \bibinfo{title}{{ASHA}: Assistive teleoperation via human-in-the-loop
  reinforcement learning}.
\newblock \bibinfo{journal}{arXiv preprint arXiv:2202.02465} .
%Type = Inproceedings
\bibitem[{Chen et~al.(2021a)Chen, Xiang, Li, Tian, Tong, Niu, Liu, Li and
  Chen}]{chen2021protecting}
\bibinfo{author}{Chen, T.}, \bibinfo{author}{Xiang, Y.}, \bibinfo{author}{Li,
  Y.}, \bibinfo{author}{Tian, Y.}, \bibinfo{author}{Tong, E.},
  \bibinfo{author}{Niu, W.}, \bibinfo{author}{Liu, J.}, \bibinfo{author}{Li,
  G.}, \bibinfo{author}{Chen, Q.A.}, \bibinfo{year}{2021}a.
\newblock \bibinfo{title}{Protecting reward function of reinforcement learning
  via minimal and non-catastrophic adversarial trajectory}, in:
  \bibinfo{booktitle}{Int. Symp. Reliable Distributed Syst.},
  \bibinfo{organization}{IEEE}. pp. \bibinfo{pages}{299--309}.
%Type = Article
\bibitem[{Chen et~al.(2021b)Chen, Liu and Khoussainov}]{chen2021agent}
\bibinfo{author}{Chen, Y.}, \bibinfo{author}{Liu, J.},
  \bibinfo{author}{Khoussainov, B.}, \bibinfo{year}{2021}b.
\newblock \bibinfo{title}{Agent-level maximum entropy inverse reinforcement
  learning for mean field games}.
\newblock \bibinfo{journal}{arXiv preprint arXiv:2104.14654} .
%Type = Article
\bibitem[{Chen et~al.(2022c)Chen, Zeng, Wang, Lu and Yang}]{chen2022zero}
\bibinfo{author}{Chen, Y.}, \bibinfo{author}{Zeng, C.}, \bibinfo{author}{Wang,
  Z.}, \bibinfo{author}{Lu, P.}, \bibinfo{author}{Yang, C.},
  \bibinfo{year}{2022}c.
\newblock \bibinfo{title}{Zero-shot sim-to-real transfer of reinforcement
  learning framework for robotics manipulation with demonstration and force
  feedback}.
\newblock \bibinfo{journal}{Robotica} , \bibinfo{pages}{1--10}.
%Type = Article
\bibitem[{Choi and Kim(2012)}]{choi2012nonparametric}
\bibinfo{author}{Choi, J.}, \bibinfo{author}{Kim, K.E.}, \bibinfo{year}{2012}.
\newblock \bibinfo{title}{Nonparametric bayesian inverse reinforcement learning
  for multiple reward functions}.
\newblock \bibinfo{journal}{Adv. Neural Inf. Process Syst.}
  \bibinfo{volume}{25}.
%Type = Article
\bibitem[{Christiano et~al.(2017)Christiano, Leike, Brown, Martic, Legg and
  Amodei}]{christiano2017deep}
\bibinfo{author}{Christiano, P.F.}, \bibinfo{author}{Leike, J.},
  \bibinfo{author}{Brown, T.}, \bibinfo{author}{Martic, M.},
  \bibinfo{author}{Legg, S.}, \bibinfo{author}{Amodei, D.},
  \bibinfo{year}{2017}.
\newblock \bibinfo{title}{Deep reinforcement learning from human preferences}.
\newblock \bibinfo{journal}{Adv. Neural Inf. Process Syst.}
  \bibinfo{volume}{30}.
%Type = Article
\bibitem[{Cruz~Jr et~al.(2019)Cruz~Jr, Du and Taylor}]{cruz2019jointly}
\bibinfo{author}{Cruz~Jr, G.V.}, \bibinfo{author}{Du, Y.},
  \bibinfo{author}{Taylor, M.E.}, \bibinfo{year}{2019}.
\newblock \bibinfo{title}{Jointly pre-training with supervised, autoencoder,
  and value losses for deep reinforcement learning}.
\newblock \bibinfo{journal}{arXiv preprint arXiv:1904.02206} .
%Type = Article
\bibitem[{Dann and Brunskill(2015)}]{dann2015sample}
\bibinfo{author}{Dann, C.}, \bibinfo{author}{Brunskill, E.},
  \bibinfo{year}{2015}.
\newblock \bibinfo{title}{Sample complexity of episodic fixed-horizon
  reinforcement learning}.
\newblock \bibinfo{journal}{NeurIPS} \bibinfo{volume}{28}.
%Type = Inproceedings
\bibitem[{Dann et~al.(2022)Dann, Mansour, Mohri, Sekhari and
  Sridharan}]{dann2022guarantees}
\bibinfo{author}{Dann, C.}, \bibinfo{author}{Mansour, Y.},
  \bibinfo{author}{Mohri, M.}, \bibinfo{author}{Sekhari, A.},
  \bibinfo{author}{Sridharan, K.}, \bibinfo{year}{2022}.
\newblock \bibinfo{title}{Guarantees for epsilon-greedy reinforcement learning
  with function approximation}, in: \bibinfo{booktitle}{ICML},
  \bibinfo{organization}{PMLR}. pp. \bibinfo{pages}{4666--4689}.
%Type = Article
\bibitem[{{De Moor} et~al.(2022){De Moor}, Gijsbrechts and
  Boute}]{DEMOOR2022535}
\bibinfo{author}{{De Moor}, B.J.}, \bibinfo{author}{Gijsbrechts, J.},
  \bibinfo{author}{Boute, R.N.}, \bibinfo{year}{2022}.
\newblock \bibinfo{title}{Reward shaping to improve the performance of deep
  reinforcement learning in perishable inventory management}.
\newblock \bibinfo{journal}{Eur. J. Oper. Res.} \bibinfo{volume}{301},
  \bibinfo{pages}{535--545}.
%Type = Article
\bibitem[{Degris et~al.(2012)Degris, White and Sutton}]{degris2012off}
\bibinfo{author}{Degris, T.}, \bibinfo{author}{White, M.},
  \bibinfo{author}{Sutton, R.S.}, \bibinfo{year}{2012}.
\newblock \bibinfo{title}{Off-policy actor-critic}.
\newblock \bibinfo{journal}{arXiv preprint arXiv:1205.4839} .
%Type = Inproceedings
\bibitem[{Devlin and Kudenko(2012)}]{devlin2012dynamic}
\bibinfo{author}{Devlin, S.M.}, \bibinfo{author}{Kudenko, D.},
  \bibinfo{year}{2012}.
\newblock \bibinfo{title}{Dynamic potential-based reward shaping}, in:
  \bibinfo{booktitle}{AAMAS}, \bibinfo{organization}{IFAAMAS}. pp.
  \bibinfo{pages}{433--440}.
%Type = Article
\bibitem[{Dexter et~al.(2021)Dexter, Bello and Honorio}]{dexter2021inverse}
\bibinfo{author}{Dexter, G.}, \bibinfo{author}{Bello, K.},
  \bibinfo{author}{Honorio, J.}, \bibinfo{year}{2021}.
\newblock \bibinfo{title}{Inverse reinforcement learning in a continuous state
  space with formal guarantees}.
\newblock \bibinfo{journal}{NeurIPS} \bibinfo{volume}{34},
  \bibinfo{pages}{6972--6982}.
%Type = Inproceedings
\bibitem[{Dimitrakakis and Rothkopf(2011)}]{dimitrakakis2011bayesian}
\bibinfo{author}{Dimitrakakis, C.}, \bibinfo{author}{Rothkopf, C.A.},
  \bibinfo{year}{2011}.
\newblock \bibinfo{title}{Bayesian multitask inverse reinforcement learning},
  in: \bibinfo{booktitle}{Eur. Workshop Reinforcement Learn.},
  \bibinfo{organization}{Springer}. pp. \bibinfo{pages}{273--284}.
%Type = Article
\bibitem[{Ding et~al.(2019)Ding, Ding, Wei and Han}]{DING2019243}
\bibinfo{author}{Ding, D.}, \bibinfo{author}{Ding, Z.}, \bibinfo{author}{Wei,
  G.}, \bibinfo{author}{Han, F.}, \bibinfo{year}{2019}.
\newblock \bibinfo{title}{An improved reinforcement learning algorithm based on
  knowledge transfer and applications in autonomous vehicles}.
\newblock \bibinfo{journal}{Neurocomputing} \bibinfo{volume}{361},
  \bibinfo{pages}{243--255}.
%Type = Article
\bibitem[{Du and Ding(2021)}]{du2021survey}
\bibinfo{author}{Du, W.}, \bibinfo{author}{Ding, S.}, \bibinfo{year}{2021}.
\newblock \bibinfo{title}{A survey on multi-agent deep reinforcement learning:
  from the perspective of challenges and applications}.
\newblock \bibinfo{journal}{Artif. Intell. Rev.} \bibinfo{volume}{54},
  \bibinfo{pages}{3215--3238}.
%Type = Article
\bibitem[{Duan et~al.(2016)Duan, Schulman, Chen, Bartlett, Sutskever and
  Abbeel}]{duan2016rl}
\bibinfo{author}{Duan, Y.}, \bibinfo{author}{Schulman, J.},
  \bibinfo{author}{Chen, X.}, \bibinfo{author}{Bartlett, P.L.},
  \bibinfo{author}{Sutskever, I.}, \bibinfo{author}{Abbeel, P.},
  \bibinfo{year}{2016}.
\newblock \bibinfo{title}{Rl$^2$: Fast reinforcement learning via slow
  reinforcement learning}.
\newblock \bibinfo{journal}{arXiv preprint arXiv:1611.02779} .
%Type = Inproceedings
\bibitem[{Elnaggar and Bezzo(2018)}]{elnaggar2018irl}
\bibinfo{author}{Elnaggar, M.}, \bibinfo{author}{Bezzo, N.},
  \bibinfo{year}{2018}.
\newblock \bibinfo{title}{An irl approach for cyber-physical attack intention
  prediction and recovery}, in: \bibinfo{booktitle}{Ann. American Control
  Conf.}, \bibinfo{organization}{IEEE}. pp. \bibinfo{pages}{222--227}.
%Type = Article
\bibitem[{Even-Dar et~al.(2003)Even-Dar, Mansour and
  Bartlett}]{even2003learning}
\bibinfo{author}{Even-Dar, E.}, \bibinfo{author}{Mansour, Y.},
  \bibinfo{author}{Bartlett, P.}, \bibinfo{year}{2003}.
\newblock \bibinfo{title}{Learning rates for q-learning}.
\newblock \bibinfo{journal}{J. Mach. Learn. Res.} \bibinfo{volume}{5}.
%Type = Inproceedings
\bibitem[{Exarchos et~al.(2021)Exarchos, Jiang, Yu and
  Liu}]{exarchos2021policy}
\bibinfo{author}{Exarchos, I.}, \bibinfo{author}{Jiang, Y.},
  \bibinfo{author}{Yu, W.}, \bibinfo{author}{Liu, C.K.}, \bibinfo{year}{2021}.
\newblock \bibinfo{title}{Policy transfer via kinematic domain randomization
  and adaptation}, in: \bibinfo{booktitle}{IEEE Int. Conf. Robot. Autom.},
  \bibinfo{organization}{IEEE}. pp. \bibinfo{pages}{45--51}.
%Type = Inproceedings
\bibitem[{Fern{\'a}ndez et~al.(2021)Fern{\'a}ndez, Fern{\'a}ndez and
  Garc{\'\i}a}]{fernandez2021probabilistic}
\bibinfo{author}{Fern{\'a}ndez, D.}, \bibinfo{author}{Fern{\'a}ndez, F.},
  \bibinfo{author}{Garc{\'\i}a, J.}, \bibinfo{year}{2021}.
\newblock \bibinfo{title}{Probabilistic multi-knowledge transfer in
  reinforcement learning}, in: \bibinfo{booktitle}{20th IEEE Int. Conf. Mach.
  Learn. Appl.}, \bibinfo{organization}{IEEE}. pp. \bibinfo{pages}{471--476}.
%Type = Inproceedings
\bibitem[{Filos et~al.(2021)Filos, Lyle, Gal, Levine, Jaques and
  Farquhar}]{filos2021psiphi}
\bibinfo{author}{Filos, A.}, \bibinfo{author}{Lyle, C.}, \bibinfo{author}{Gal,
  Y.}, \bibinfo{author}{Levine, S.}, \bibinfo{author}{Jaques, N.},
  \bibinfo{author}{Farquhar, G.}, \bibinfo{year}{2021}.
\newblock \bibinfo{title}{Psiphi-learning: Reinforcement learning with
  demonstrations using successor features and inverse temporal difference
  learning}, in: \bibinfo{booktitle}{ICML}, \bibinfo{organization}{PMLR}. pp.
  \bibinfo{pages}{3305--3317}.
%Type = Inproceedings
\bibitem[{Finn et~al.(2017)Finn, Abbeel and Levine}]{finn2017model}
\bibinfo{author}{Finn, C.}, \bibinfo{author}{Abbeel, P.},
  \bibinfo{author}{Levine, S.}, \bibinfo{year}{2017}.
\newblock \bibinfo{title}{Model-agnostic meta-learning for fast adaptation of
  deep networks}, in: \bibinfo{booktitle}{ICML}, \bibinfo{organization}{PMLR}.
  pp. \bibinfo{pages}{1126--1135}.
%Type = Article
\bibitem[{Finn et~al.(2016)Finn, Christiano, Abbeel and
  Levine}]{finn2016connection}
\bibinfo{author}{Finn, C.}, \bibinfo{author}{Christiano, P.},
  \bibinfo{author}{Abbeel, P.}, \bibinfo{author}{Levine, S.},
  \bibinfo{year}{2016}.
\newblock \bibinfo{title}{A connection between generative adversarial networks,
  inverse reinforcement learning, and energy-based models}.
\newblock \bibinfo{journal}{arXiv preprint arXiv:1611.03852} .
%Type = Article
\bibitem[{Fu et~al.(2017)Fu, Luo and Levine}]{fu2017learning}
\bibinfo{author}{Fu, J.}, \bibinfo{author}{Luo, K.}, \bibinfo{author}{Levine,
  S.}, \bibinfo{year}{2017}.
\newblock \bibinfo{title}{Learning robust rewards with adversarial inverse
  reinforcement learning}.
\newblock \bibinfo{journal}{arXiv preprint arXiv:1710.11248} .
%Type = Inproceedings
\bibitem[{Fukumoto et~al.(2020)Fukumoto, Tadokoro and
  Takadama}]{fukumoto2020cooperative}
\bibinfo{author}{Fukumoto, Y.}, \bibinfo{author}{Tadokoro, M.},
  \bibinfo{author}{Takadama, K.}, \bibinfo{year}{2020}.
\newblock \bibinfo{title}{Cooperative multi-agent inverse reinforcement
  learning based on selfish expert and its behavior archives}, in:
  \bibinfo{booktitle}{IEEE Int. Symp. Comput. Intell. Inform.},
  \bibinfo{organization}{IEEE}. pp. \bibinfo{pages}{2202--2209}.
%Type = Article
\bibitem[{Galatolo et~al.(2021)Galatolo, Cimino and
  Vaglini}]{galatolo2021solving}
\bibinfo{author}{Galatolo, F.A.}, \bibinfo{author}{Cimino, M.G.},
  \bibinfo{author}{Vaglini, G.}, \bibinfo{year}{2021}.
\newblock \bibinfo{title}{Solving the scalarization issues of advantage-based
  reinforcement learning algorithms}.
\newblock \bibinfo{journal}{Comput. Electr. Eng.} \bibinfo{volume}{92},
  \bibinfo{pages}{107117}.
%Type = Article
\bibitem[{Gao et~al.(2018)Gao, Shi, Xie and Cheng}]{gao2018car}
\bibinfo{author}{Gao, H.}, \bibinfo{author}{Shi, G.}, \bibinfo{author}{Xie,
  G.}, \bibinfo{author}{Cheng, B.}, \bibinfo{year}{2018}.
\newblock \bibinfo{title}{Car-following method based on inverse reinforcement
  learning for autonomous vehicle decision-making}.
\newblock \bibinfo{journal}{Int. J. Adv. Robot. Syst.} \bibinfo{volume}{15},
  \bibinfo{pages}{1729881418817162}.
%Type = Article
\bibitem[{Gao et~al.(2022)Gao, Mynuddin, Wunsch and Jiang}]{9404328}
\bibinfo{author}{Gao, W.}, \bibinfo{author}{Mynuddin, M.},
  \bibinfo{author}{Wunsch, D.C.}, \bibinfo{author}{Jiang, Z.P.},
  \bibinfo{year}{2022}.
\newblock \bibinfo{title}{Reinforcement learning-based cooperative optimal
  output regulation via distributed adaptive internal model}.
\newblock \bibinfo{journal}{IEEE Trans. Neural Netw. Learn. Syst.}
  \bibinfo{volume}{33}, \bibinfo{pages}{5229--5240}.
%Type = Inproceedings
\bibitem[{Ghasemipour et~al.(2020)Ghasemipour, Zemel and
  Gu}]{ghasemipour2020divergence}
\bibinfo{author}{Ghasemipour, S.K.S.}, \bibinfo{author}{Zemel, R.},
  \bibinfo{author}{Gu, S.}, \bibinfo{year}{2020}.
\newblock \bibinfo{title}{A divergence minimization perspective on imitation
  learning methods}, in: \bibinfo{booktitle}{Conf. Robot Learn.},
  \bibinfo{organization}{PMLR}. pp. \bibinfo{pages}{1259--1277}.
%Type = Inproceedings
\bibitem[{Gimelfarb et~al.(2021)Gimelfarb, Sanner and
  Lee}]{gimelfarb2021contextual}
\bibinfo{author}{Gimelfarb, M.}, \bibinfo{author}{Sanner, S.},
  \bibinfo{author}{Lee, C.G.}, \bibinfo{year}{2021}.
\newblock \bibinfo{title}{Contextual policy transfer in reinforcement learning
  domains via deep mixtures-of-experts}, in: \bibinfo{booktitle}{Uncertainty
  Artif. Intell.}, \bibinfo{organization}{PMLR}. pp.
  \bibinfo{pages}{1787--1797}.
%Type = Article
\bibitem[{Goecks(2020)}]{goecks2020human}
\bibinfo{author}{Goecks, V.G.}, \bibinfo{year}{2020}.
\newblock \bibinfo{title}{Human-in-the-loop methods for data-driven and
  reinforcement learning systems}.
\newblock \bibinfo{journal}{arXiv preprint arXiv:2008.13221} .
%Type = Article
\bibitem[{Goodfellow et~al.(2020)Goodfellow, Pouget-Abadie, Mirza, Xu,
  Warde-Farley, Ozair, Courville and Bengio}]{goodfellow2020generative}
\bibinfo{author}{Goodfellow, I.}, \bibinfo{author}{Pouget-Abadie, J.},
  \bibinfo{author}{Mirza, M.}, \bibinfo{author}{Xu, B.},
  \bibinfo{author}{Warde-Farley, D.}, \bibinfo{author}{Ozair, S.},
  \bibinfo{author}{Courville, A.}, \bibinfo{author}{Bengio, Y.},
  \bibinfo{year}{2020}.
\newblock \bibinfo{title}{Generative adversarial networks}.
\newblock \bibinfo{journal}{Commun. ACM.} \bibinfo{volume}{63},
  \bibinfo{pages}{139--144}.
%Type = Article
\bibitem[{Gu et~al.(2023)Gu, Kuba, Chen, Du, Yang, Knoll and Yang}]{gu2023safe}
\bibinfo{author}{Gu, S.}, \bibinfo{author}{Kuba, J.G.}, \bibinfo{author}{Chen,
  Y.}, \bibinfo{author}{Du, Y.}, \bibinfo{author}{Yang, L.},
  \bibinfo{author}{Knoll, A.}, \bibinfo{author}{Yang, Y.},
  \bibinfo{year}{2023}.
\newblock \bibinfo{title}{Safe multi-agent reinforcement learning for
  multi-robot control}.
\newblock \bibinfo{journal}{Artif. Intell.} \bibinfo{volume}{319},
  \bibinfo{pages}{103905}.
%Type = Article
\bibitem[{Guan et~al.(2021)Guan, Verma, Guo, Zhang and
  Kambhampati}]{guan2021widening}
\bibinfo{author}{Guan, L.}, \bibinfo{author}{Verma, M.}, \bibinfo{author}{Guo,
  S.S.}, \bibinfo{author}{Zhang, R.}, \bibinfo{author}{Kambhampati, S.},
  \bibinfo{year}{2021}.
\newblock \bibinfo{title}{Widening the pipeline in human-guided reinforcement
  learning with explanation and context-aware data augmentation}.
\newblock \bibinfo{journal}{Adv. Neural Inf. Process Syst.}
  \bibinfo{volume}{34}, \bibinfo{pages}{21885--21897}.
%Type = Inproceedings
\bibitem[{Haarnoja et~al.(2018)Haarnoja, Zhou, Abbeel and
  Levine}]{haarnoja2018soft}
\bibinfo{author}{Haarnoja, T.}, \bibinfo{author}{Zhou, A.},
  \bibinfo{author}{Abbeel, P.}, \bibinfo{author}{Levine, S.},
  \bibinfo{year}{2018}.
\newblock \bibinfo{title}{Soft actor-critic: Off-policy maximum entropy deep
  reinforcement learning with a stochastic actor}, in:
  \bibinfo{booktitle}{ICML}, \bibinfo{organization}{PMLR}. pp.
  \bibinfo{pages}{1861--1870}.
%Type = Article
\bibitem[{Hasanbeig et~al.(2023)Hasanbeig, Kroening and
  Abate}]{hasanbeig2023certified}
\bibinfo{author}{Hasanbeig, H.}, \bibinfo{author}{Kroening, D.},
  \bibinfo{author}{Abate, A.}, \bibinfo{year}{2023}.
\newblock \bibinfo{title}{Certified reinforcement learning with logic
  guidance}.
\newblock \bibinfo{journal}{Artif. Intell.} \bibinfo{volume}{322},
  \bibinfo{pages}{103949}.
%Type = Article
\bibitem[{Hassani et~al.(2021)Hassani, Hallaji, Razavi-Far and
  Saif}]{hassani2021unsupervised}
\bibinfo{author}{Hassani, H.}, \bibinfo{author}{Hallaji, E.},
  \bibinfo{author}{Razavi-Far, R.}, \bibinfo{author}{Saif, M.},
  \bibinfo{year}{2021}.
\newblock \bibinfo{title}{Unsupervised concrete feature selection based on
  mutual information for diagnosing faults and cyber-attacks in power systems}.
\newblock \bibinfo{journal}{Eng. Appl. Artif. Intell.} \bibinfo{volume}{100},
  \bibinfo{pages}{104150}.
%Type = Article
\bibitem[{Hassani et~al.(2024)Hassani, Nikan and Shami}]{HASSANI2024109147}
\bibinfo{author}{Hassani, H.}, \bibinfo{author}{Nikan, S.},
  \bibinfo{author}{Shami, A.}, \bibinfo{year}{2024}.
\newblock \bibinfo{title}{Traffic navigation via reinforcement learning with
  episodic-guided prioritized experience replay}.
\newblock \bibinfo{journal}{Eng. Appl. Artif. Intell.} \bibinfo{volume}{137},
  \bibinfo{pages}{109147}.
%Type = Article
\bibitem[{Hassani et~al.(2025)Hassani, Nikan and Shami}]{HASSANI2025128836}
\bibinfo{author}{Hassani, H.}, \bibinfo{author}{Nikan, S.},
  \bibinfo{author}{Shami, A.}, \bibinfo{year}{2025}.
\newblock \bibinfo{title}{Improved exploration–exploitation trade-off through
  adaptive prioritized experience replay}.
\newblock \bibinfo{journal}{Neurocomputing} \bibinfo{volume}{614},
  \bibinfo{pages}{128836}.
%Type = Article
\bibitem[{Hassani et~al.(2022)Hassani, Razavi-Far and Saif}]{hassani2022real}
\bibinfo{author}{Hassani, H.}, \bibinfo{author}{Razavi-Far, R.},
  \bibinfo{author}{Saif, M.}, \bibinfo{year}{2022}.
\newblock \bibinfo{title}{Real-time out-of-step prediction control to prevent
  emerging blackouts in power systems: A reinforcement learning approach}.
\newblock \bibinfo{journal}{Appl. Energy} \bibinfo{volume}{314},
  \bibinfo{pages}{118861}.
%Type = Article
\bibitem[{Hassani et~al.(2023)Hassani, Razavi-Far, Saif and
  Herrera-Viedma}]{9928222}
\bibinfo{author}{Hassani, H.}, \bibinfo{author}{Razavi-Far, R.},
  \bibinfo{author}{Saif, M.}, \bibinfo{author}{Herrera-Viedma, E.},
  \bibinfo{year}{2023}.
\newblock \bibinfo{title}{Reinforcement learning-based feedback and
  weight-adjustment mechanisms for consensus reaching in group decision
  making}.
\newblock \bibinfo{journal}{IEEE Trans. Syst., Man, Cybern.: Syst.}
  \bibinfo{volume}{53}, \bibinfo{pages}{2456--2468}.
%Type = Article
\bibitem[{He et~al.(2022)He, Chen, Xu, Yang, Liu and Yang}]{he2022irlsot}
\bibinfo{author}{He, C.}, \bibinfo{author}{Chen, L.}, \bibinfo{author}{Xu, L.},
  \bibinfo{author}{Yang, C.}, \bibinfo{author}{Liu, X.}, \bibinfo{author}{Yang,
  B.}, \bibinfo{year}{2022}.
\newblock \bibinfo{title}{Irlsot: Inverse reinforcement learning for
  scene-oriented trajectory prediction}.
\newblock \bibinfo{journal}{IET Intell. Transp. Syst.} \bibinfo{volume}{16},
  \bibinfo{pages}{769--781}.
%Type = Inproceedings
\bibitem[{Henderson et~al.(2018)Henderson, Chang, Bacon, Meger, Pineau and
  Precup}]{henderson2018optiongan}
\bibinfo{author}{Henderson, P.}, \bibinfo{author}{Chang, W.D.},
  \bibinfo{author}{Bacon, P.L.}, \bibinfo{author}{Meger, D.},
  \bibinfo{author}{Pineau, J.}, \bibinfo{author}{Precup, D.},
  \bibinfo{year}{2018}.
\newblock \bibinfo{title}{Option{GAN}: Learning joint reward-policy options
  using generative adversarial inverse reinforcement learning}, in:
  \bibinfo{booktitle}{Proc. AAAI Conf. Artif. Intell.}
%Type = Inproceedings
\bibitem[{Hester et~al.(2018)Hester, Vecerik, Pietquin, Lanctot, Schaul, Piot,
  Horgan, Quan, Sendonaris, Osband et~al.}]{hester2018deep}
\bibinfo{author}{Hester, T.}, \bibinfo{author}{Vecerik, M.},
  \bibinfo{author}{Pietquin, O.}, \bibinfo{author}{Lanctot, M.},
  \bibinfo{author}{Schaul, T.}, \bibinfo{author}{Piot, B.},
  \bibinfo{author}{Horgan, D.}, \bibinfo{author}{Quan, J.},
  \bibinfo{author}{Sendonaris, A.}, \bibinfo{author}{Osband, I.}, et~al.,
  \bibinfo{year}{2018}.
\newblock \bibinfo{title}{Deep {Q}-learning from demonstrations}, in:
  \bibinfo{booktitle}{Proc. AAAI Conf. Artif. Intell.}
%Type = Article
\bibitem[{Hill et~al.(2020)Hill, Mokra, Wong and Harley}]{hill2020human}
\bibinfo{author}{Hill, F.}, \bibinfo{author}{Mokra, S.}, \bibinfo{author}{Wong,
  N.}, \bibinfo{author}{Harley, T.}, \bibinfo{year}{2020}.
\newblock \bibinfo{title}{Human instruction-following with deep reinforcement
  learning via transfer-learning from text}.
\newblock \bibinfo{journal}{arXiv preprint arXiv:2005.09382} .
%Type = Article
\bibitem[{Ho and Ermon(2016)}]{ho2016generative}
\bibinfo{author}{Ho, J.}, \bibinfo{author}{Ermon, S.}, \bibinfo{year}{2016}.
\newblock \bibinfo{title}{Generative adversarial imitation learning}.
\newblock \bibinfo{journal}{Adv. Neural Inf. Process Syst.}
  \bibinfo{volume}{29}.
%Type = Inproceedings
\bibitem[{Holmberg et~al.(2020)Holmberg, Davidsson and
  Linde}]{holmberg2020feature}
\bibinfo{author}{Holmberg, L.}, \bibinfo{author}{Davidsson, P.},
  \bibinfo{author}{Linde, P.}, \bibinfo{year}{2020}.
\newblock \bibinfo{title}{A feature space focus in machine teaching}, in:
  \bibinfo{booktitle}{IEEE Int. Conf. Pervasive Comput. Commun. Workshops},
  \bibinfo{organization}{IEEE}. pp. \bibinfo{pages}{1--2}.
%Type = Article
\bibitem[{Horgan et~al.(2018)Horgan, Quan, Budden, Barth-Maron, Hessel,
  Van~Hasselt and Silver}]{horgan2018distributed}
\bibinfo{author}{Horgan, D.}, \bibinfo{author}{Quan, J.},
  \bibinfo{author}{Budden, D.}, \bibinfo{author}{Barth-Maron, G.},
  \bibinfo{author}{Hessel, M.}, \bibinfo{author}{Van~Hasselt, H.},
  \bibinfo{author}{Silver, D.}, \bibinfo{year}{2018}.
\newblock \bibinfo{title}{Distributed prioritized experience replay}.
\newblock \bibinfo{journal}{arXiv preprint arXiv:1803.00933} .
%Type = Inproceedings
\bibitem[{Hoshino et~al.(2022)Hoshino, Ota, Kanezaki and
  Yokota}]{hoshino2022opirl}
\bibinfo{author}{Hoshino, H.}, \bibinfo{author}{Ota, K.},
  \bibinfo{author}{Kanezaki, A.}, \bibinfo{author}{Yokota, R.},
  \bibinfo{year}{2022}.
\newblock \bibinfo{title}{{OPIRL}: Sample efficient off-policy inverse
  reinforcement learning via distribution matching}, in:
  \bibinfo{booktitle}{Int. Conf. Robotics Autom.},
  \bibinfo{organization}{IEEE}. pp. \bibinfo{pages}{448--454}.
%Type = Article
\bibitem[{Hu and Sun(2021)}]{HU2021106967}
\bibinfo{author}{Hu, Y.}, \bibinfo{author}{Sun, S.}, \bibinfo{year}{2021}.
\newblock \bibinfo{title}{Rl-vaegan: Adversarial defense for reinforcement
  learning agents via style transfer}.
\newblock \bibinfo{journal}{Knowl.-Based Syst.} \bibinfo{volume}{221},
  \bibinfo{pages}{106967}.
%Type = Article
\bibitem[{Huang et~al.(2023)Huang, Wang, Zhou, Zhang and Lin}]{9956746}
\bibinfo{author}{Huang, C.}, \bibinfo{author}{Wang, G.}, \bibinfo{author}{Zhou,
  Z.}, \bibinfo{author}{Zhang, R.}, \bibinfo{author}{Lin, L.},
  \bibinfo{year}{2023}.
\newblock \bibinfo{title}{Reward-adaptive reinforcement learning: Dynamic
  policy gradient optimization for bipedal locomotion}.
\newblock \bibinfo{journal}{IEEE Trans. Pattern Anal. Mach. Intell.}
  \bibinfo{volume}{45}, \bibinfo{pages}{7686--7695}.
%Type = Article
\bibitem[{Huang et~al.(2021)Huang, Zhang, Ouyang, Wei, Lin, Su and
  Lin}]{9537641}
\bibinfo{author}{Huang, C.}, \bibinfo{author}{Zhang, R.},
  \bibinfo{author}{Ouyang, M.}, \bibinfo{author}{Wei, P.},
  \bibinfo{author}{Lin, J.}, \bibinfo{author}{Su, J.}, \bibinfo{author}{Lin,
  L.}, \bibinfo{year}{2021}.
\newblock \bibinfo{title}{Deductive reinforcement learning for visual
  autonomous urban driving navigation}.
\newblock \bibinfo{journal}{IEEE Trans. Neural Netw. Learn. Syst.}
  \bibinfo{volume}{32}, \bibinfo{pages}{5379--5391}.
%Type = Article
\bibitem[{Huang et~al.(2022)Huang, Xu, Zhu, Shi, Fang and
  Zhao}]{huang2022curriculum}
\bibinfo{author}{Huang, P.}, \bibinfo{author}{Xu, M.}, \bibinfo{author}{Zhu,
  J.}, \bibinfo{author}{Shi, L.}, \bibinfo{author}{Fang, F.},
  \bibinfo{author}{Zhao, D.}, \bibinfo{year}{2022}.
\newblock \bibinfo{title}{Curriculum reinforcement learning using optimal
  transport via gradual domain adaptation}.
\newblock \bibinfo{journal}{arXiv preprint arXiv:2210.10195} .
%Type = Article
\bibitem[{Imani and Ghoreishi(2022)}]{9334410}
\bibinfo{author}{Imani, M.}, \bibinfo{author}{Ghoreishi, S.F.},
  \bibinfo{year}{2022}.
\newblock \bibinfo{title}{Scalable inverse reinforcement learning through
  multifidelity bayesian optimization}.
\newblock \bibinfo{journal}{IEEE Trans. Neural Netw. Learn. Syst.}
  \bibinfo{volume}{33}, \bibinfo{pages}{4125--4132}.
%Type = Article
\bibitem[{Jaynes(1957)}]{jaynes1957information}
\bibinfo{author}{Jaynes, E.T.}, \bibinfo{year}{1957}.
\newblock \bibinfo{title}{Information theory and statistical mechanics}.
\newblock \bibinfo{journal}{Physical Rev.} \bibinfo{volume}{106},
  \bibinfo{pages}{620}.
%Type = Article
\bibitem[{Jeon et~al.(2020)Jeon, Barde, Nowrouzezahrai and
  Pineau}]{jeon2020scalable}
\bibinfo{author}{Jeon, W.}, \bibinfo{author}{Barde, P.},
  \bibinfo{author}{Nowrouzezahrai, D.}, \bibinfo{author}{Pineau, J.},
  \bibinfo{year}{2020}.
\newblock \bibinfo{title}{Scalable multi-agent inverse reinforcement learning
  via actor-attention-critic}.
\newblock \bibinfo{journal}{arXiv preprint arXiv:2002.10525} .
%Type = Inproceedings
\bibitem[{Jeong et~al.(2020)Jeong, Aytar, Khosid, Zhou, Kay, Lampe, Bousmalis
  and Nori}]{jeong2020self}
\bibinfo{author}{Jeong, R.}, \bibinfo{author}{Aytar, Y.},
  \bibinfo{author}{Khosid, D.}, \bibinfo{author}{Zhou, Y.},
  \bibinfo{author}{Kay, J.}, \bibinfo{author}{Lampe, T.},
  \bibinfo{author}{Bousmalis, K.}, \bibinfo{author}{Nori, F.},
  \bibinfo{year}{2020}.
\newblock \bibinfo{title}{Self-supervised sim-to-real adaptation for visual
  robotic manipulation}, in: \bibinfo{booktitle}{IEEE Int. Conf. Robot.
  Autom.}, \bibinfo{organization}{IEEE}. pp. \bibinfo{pages}{2718--2724}.
%Type = Inproceedings
\bibitem[{Jiang and Agarwal(2018)}]{jiang2018open}
\bibinfo{author}{Jiang, N.}, \bibinfo{author}{Agarwal, A.},
  \bibinfo{year}{2018}.
\newblock \bibinfo{title}{Open problem: The dependence of sample complexity
  lower bounds on planning horizon}, in: \bibinfo{booktitle}{Conf. Learn.
  Theory}, \bibinfo{organization}{PMLR}. pp. \bibinfo{pages}{3395--3398}.
%Type = Inproceedings
\bibitem[{Jiang et~al.(2021)Jiang, Zhang, Ho, Bai, Liu, Levine and
  Tan}]{jiang2021simgan}
\bibinfo{author}{Jiang, Y.}, \bibinfo{author}{Zhang, T.}, \bibinfo{author}{Ho,
  D.}, \bibinfo{author}{Bai, Y.}, \bibinfo{author}{Liu, C.K.},
  \bibinfo{author}{Levine, S.}, \bibinfo{author}{Tan, J.},
  \bibinfo{year}{2021}.
\newblock \bibinfo{title}{{SimGAN}: Hybrid simulator identification for domain
  adaptation via adversarial reinforcement learning}, in:
  \bibinfo{booktitle}{IEEE Int. Conf. Robot. Autom.},
  \bibinfo{organization}{IEEE}. pp. \bibinfo{pages}{2884--2890}.
%Type = Article
\bibitem[{Joshi and Chowdhary(2021)}]{joshi2021adaptive}
\bibinfo{author}{Joshi, G.}, \bibinfo{author}{Chowdhary, G.},
  \bibinfo{year}{2021}.
\newblock \bibinfo{title}{Adaptive policy transfer in reinforcement learning}.
\newblock \bibinfo{journal}{arXiv preprint arXiv:2105.04699} .
%Type = Article
\bibitem[{Kadokawa et~al.(2022)Kadokawa, Zhu, Tsurumine and
  Matsubara}]{kadokawa2022cyclic}
\bibinfo{author}{Kadokawa, Y.}, \bibinfo{author}{Zhu, L.},
  \bibinfo{author}{Tsurumine, Y.}, \bibinfo{author}{Matsubara, T.},
  \bibinfo{year}{2022}.
\newblock \bibinfo{title}{Cyclic policy distillation: Sample-efficient
  sim-to-real reinforcement learning with domain randomization}.
\newblock \bibinfo{journal}{arXiv preprint arXiv:2207.14561} .
%Type = Phdthesis
\bibitem[{Kakade(2003)}]{kakade2003sample}
\bibinfo{author}{Kakade, S.M.}, \bibinfo{year}{2003}.
\newblock \bibinfo{title}{On the sample complexity of reinforcement learning}.
\newblock Ph.D. thesis. UCL (University College London).
%Type = Article
\bibitem[{Kearns and Singh(1998)}]{kearns1998finite}
\bibinfo{author}{Kearns, M.}, \bibinfo{author}{Singh, S.},
  \bibinfo{year}{1998}.
\newblock \bibinfo{title}{Finite-sample convergence rates for q-learning and
  indirect algorithms}.
\newblock \bibinfo{journal}{NeurIPS} \bibinfo{volume}{11}.
%Type = Inproceedings
\bibitem[{Keramati and Brunskill(2019)}]{keramati2019value}
\bibinfo{author}{Keramati, R.}, \bibinfo{author}{Brunskill, E.},
  \bibinfo{year}{2019}.
\newblock \bibinfo{title}{Value driven representation for human-in-the-loop
  reinforcement learning}, in: \bibinfo{booktitle}{Proc. 27th ACM Conf. User
  Model., Adapt. Personal.}, pp. \bibinfo{pages}{176--180}.
%Type = Article
\bibitem[{Khaitan and Dolan(2022)}]{khaitan2022state}
\bibinfo{author}{Khaitan, S.}, \bibinfo{author}{Dolan, J.M.},
  \bibinfo{year}{2022}.
\newblock \bibinfo{title}{State dropout-based curriculum reinforcement learning
  for self-driving at unsignalized intersections}.
\newblock \bibinfo{journal}{arXiv preprint arXiv:2207.04361} .
%Type = Inproceedings
\bibitem[{Khodadadian et~al.(2022)Khodadadian, Sharma, Joshi and
  Maguluri}]{khodadadian2022federated}
\bibinfo{author}{Khodadadian, S.}, \bibinfo{author}{Sharma, P.},
  \bibinfo{author}{Joshi, G.}, \bibinfo{author}{Maguluri, S.T.},
  \bibinfo{year}{2022}.
\newblock \bibinfo{title}{Federated reinforcement learning: Linear speedup
  under markovian sampling}, in: \bibinfo{booktitle}{ICML},
  \bibinfo{organization}{PMLR}. pp. \bibinfo{pages}{10997--11057}.
%Type = Inproceedings
\bibitem[{Knox and Stone(2009)}]{knox2009interactively}
\bibinfo{author}{Knox, W.B.}, \bibinfo{author}{Stone, P.},
  \bibinfo{year}{2009}.
\newblock \bibinfo{title}{Interactively shaping agents via human reinforcement:
  The {TAMER} framework}, in: \bibinfo{booktitle}{Proc. fifth Int. Conf. Knowl.
  Capture}, pp. \bibinfo{pages}{9--16}.
%Type = Article
\bibitem[{Kobayashi et~al.(2019)Kobayashi, Horii, Iwaki, Nagai and
  Asada}]{kobayashi2019situated}
\bibinfo{author}{Kobayashi, K.}, \bibinfo{author}{Horii, T.},
  \bibinfo{author}{Iwaki, R.}, \bibinfo{author}{Nagai, Y.},
  \bibinfo{author}{Asada, M.}, \bibinfo{year}{2019}.
\newblock \bibinfo{title}{Situated {GAIL}: Multitask imitation using
  task-conditioned adversarial inverse reinforcement learning}.
\newblock \bibinfo{journal}{arXiv preprint arXiv:1911.00238} .
%Type = Article
\bibitem[{Komanduru and Honorio(2019)}]{komanduru2019correctness}
\bibinfo{author}{Komanduru, A.}, \bibinfo{author}{Honorio, J.},
  \bibinfo{year}{2019}.
\newblock \bibinfo{title}{On the correctness and sample complexity of inverse
  reinforcement learning}.
\newblock \bibinfo{journal}{NeurIPS} \bibinfo{volume}{32}.
%Type = Inproceedings
\bibitem[{Komanduru and Honorio(2021)}]{komanduru2021lower}
\bibinfo{author}{Komanduru, A.}, \bibinfo{author}{Honorio, J.},
  \bibinfo{year}{2021}.
\newblock \bibinfo{title}{A lower bound for the sample complexity of inverse
  reinforcement learning}, in: \bibinfo{booktitle}{ICML},
  \bibinfo{organization}{PMLR}. pp. \bibinfo{pages}{5676--5685}.
%Type = Article
\bibitem[{Kone{\v{c}}n{\`y} et~al.(2016)Kone{\v{c}}n{\`y}, McMahan, Yu,
  Richt{\'a}rik, Suresh and Bacon}]{konevcny2016federated}
\bibinfo{author}{Kone{\v{c}}n{\`y}, J.}, \bibinfo{author}{McMahan, H.B.},
  \bibinfo{author}{Yu, F.X.}, \bibinfo{author}{Richt{\'a}rik, P.},
  \bibinfo{author}{Suresh, A.T.}, \bibinfo{author}{Bacon, D.},
  \bibinfo{year}{2016}.
\newblock \bibinfo{title}{Federated learning: Strategies for improving
  communication efficiency}.
\newblock \bibinfo{journal}{arXiv preprint arXiv:1610.05492} .
%Type = Article
\bibitem[{Kostrikov et~al.(2018)Kostrikov, Agrawal, Dwibedi, Levine and
  Tompson}]{kostrikov2018discriminator}
\bibinfo{author}{Kostrikov, I.}, \bibinfo{author}{Agrawal, K.K.},
  \bibinfo{author}{Dwibedi, D.}, \bibinfo{author}{Levine, S.},
  \bibinfo{author}{Tompson, J.}, \bibinfo{year}{2018}.
\newblock \bibinfo{title}{Discriminator-actor-critic: Addressing sample
  inefficiency and reward bias in adversarial imitation learning}.
\newblock \bibinfo{journal}{arXiv preprint arXiv:1809.02925} .
%Type = Article
\bibitem[{Li et~al.(2017)Li, Modares, Chai, Lewis and Xie}]{7902130}
\bibinfo{author}{Li, J.}, \bibinfo{author}{Modares, H.}, \bibinfo{author}{Chai,
  T.}, \bibinfo{author}{Lewis, F.L.}, \bibinfo{author}{Xie, L.},
  \bibinfo{year}{2017}.
\newblock \bibinfo{title}{Off-policy reinforcement learning for synchronization
  in multiagent graphical games}.
\newblock \bibinfo{journal}{IEEE Trans. Neural Netw. Learn. Syst.}
  \bibinfo{volume}{28}, \bibinfo{pages}{2434--2445}.
%Type = Inproceedings
\bibitem[{Li and Zhu(2019)}]{li2019convergence}
\bibinfo{author}{Li, T.}, \bibinfo{author}{Zhu, Q.}, \bibinfo{year}{2019}.
\newblock \bibinfo{title}{On convergence rate of adaptive multiscale value
  function approximation for reinforcement learning}, in:
  \bibinfo{booktitle}{MLSP}, \bibinfo{organization}{IEEE}. pp.
  \bibinfo{pages}{1--6}.
%Type = Inproceedings
\bibitem[{Li et~al.(2022)Li, Wang and Yang}]{li2022settling}
\bibinfo{author}{Li, Y.}, \bibinfo{author}{Wang, R.}, \bibinfo{author}{Yang,
  L.F.}, \bibinfo{year}{2022}.
\newblock \bibinfo{title}{Settling the horizon-dependence of sample complexity
  in reinforcement learning}, in: \bibinfo{booktitle}{FOCS},
  \bibinfo{organization}{IEEE}. pp. \bibinfo{pages}{965--976}.
%Type = Article
\bibitem[{Lian et~al.(2021)Lian, Xue, Lewis and Chai}]{9565156}
\bibinfo{author}{Lian, B.}, \bibinfo{author}{Xue, W.}, \bibinfo{author}{Lewis,
  F.L.}, \bibinfo{author}{Chai, T.}, \bibinfo{year}{2021}.
\newblock \bibinfo{title}{Inverse reinforcement learning for adversarial
  apprentice games}.
\newblock \bibinfo{journal}{IEEE Trans. Neural Netw. Learn. Syst.} .
%Type = Incollection
\bibitem[{Liang et~al.(2023)Liang, Liu, Chen, Liu and
  Yang}]{liang2023federated}
\bibinfo{author}{Liang, X.}, \bibinfo{author}{Liu, Y.}, \bibinfo{author}{Chen,
  T.}, \bibinfo{author}{Liu, M.}, \bibinfo{author}{Yang, Q.},
  \bibinfo{year}{2023}.
\newblock \bibinfo{title}{Federated transfer reinforcement learning for
  autonomous driving}, in: \bibinfo{booktitle}{Federated and Transfer
  Learning}. \bibinfo{publisher}{Springer}, pp. \bibinfo{pages}{357--371}.
%Type = Article
\bibitem[{Likmeta et~al.(2021)Likmeta, Metelli, Ramponi, Tirinzoni, Giuliani
  and Restelli}]{likmeta2021dealing}
\bibinfo{author}{Likmeta, A.}, \bibinfo{author}{Metelli, A.M.},
  \bibinfo{author}{Ramponi, G.}, \bibinfo{author}{Tirinzoni, A.},
  \bibinfo{author}{Giuliani, M.}, \bibinfo{author}{Restelli, M.},
  \bibinfo{year}{2021}.
\newblock \bibinfo{title}{Dealing with multiple experts and non-stationarity in
  inverse reinforcement learning: an application to real-life problems}.
\newblock \bibinfo{journal}{Mach. Learn.} \bibinfo{volume}{110},
  \bibinfo{pages}{2541--2576}.
%Type = Article
\bibitem[{Lillicrap et~al.(2015)Lillicrap, Hunt, Pritzel, Heess, Erez, Tassa,
  Silver and Wierstra}]{lillicrap2015continuous}
\bibinfo{author}{Lillicrap, T.P.}, \bibinfo{author}{Hunt, J.J.},
  \bibinfo{author}{Pritzel, A.}, \bibinfo{author}{Heess, N.},
  \bibinfo{author}{Erez, T.}, \bibinfo{author}{Tassa, Y.},
  \bibinfo{author}{Silver, D.}, \bibinfo{author}{Wierstra, D.},
  \bibinfo{year}{2015}.
\newblock \bibinfo{title}{Continuous control with deep reinforcement learning}.
\newblock \bibinfo{journal}{arXiv preprint arXiv:1509.02971} .
%Type = Inproceedings
\bibitem[{Lin and Zhang(2018)}]{lin2018acgail}
\bibinfo{author}{Lin, J.}, \bibinfo{author}{Zhang, Z.}, \bibinfo{year}{2018}.
\newblock \bibinfo{title}{{ACGAIL}: Imitation learning about multiple
  intentions with auxiliary classifier gans}, in: \bibinfo{booktitle}{Pacific
  Rim Int. Conf. Artif. Intell.}, \bibinfo{organization}{Springer}. pp.
  \bibinfo{pages}{321--334}.
%Type = Article
\bibitem[{Lindner and El-Assady(2022)}]{lindner2022humans}
\bibinfo{author}{Lindner, D.}, \bibinfo{author}{El-Assady, M.},
  \bibinfo{year}{2022}.
\newblock \bibinfo{title}{Humans are not boltzmann distributions: Challenges
  and opportunities for modelling human feedback and interaction in
  reinforcement learning}.
\newblock \bibinfo{journal}{arXiv preprint arXiv:2206.13316} .
%Type = Inproceedings
\bibitem[{Liu and Zou(2018)}]{liu2018effects}
\bibinfo{author}{Liu, R.}, \bibinfo{author}{Zou, J.}, \bibinfo{year}{2018}.
\newblock \bibinfo{title}{The effects of memory replay in reinforcement
  learning}, in: \bibinfo{booktitle}{56th Annu. Allerton Conf. Commun.,
  Control, Comput.}, \bibinfo{organization}{IEEE}. pp.
  \bibinfo{pages}{478--485}.
%Type = Article
\bibitem[{Liu et~al.(2022)Liu, Zhu, Jiang, Ye and Zhao}]{liu2022prioritized}
\bibinfo{author}{Liu, X.}, \bibinfo{author}{Zhu, T.}, \bibinfo{author}{Jiang,
  C.}, \bibinfo{author}{Ye, D.}, \bibinfo{author}{Zhao, F.},
  \bibinfo{year}{2022}.
\newblock \bibinfo{title}{Prioritized experience replay based on multi-armed
  bandit}.
\newblock \bibinfo{journal}{Expert Syst. Appl.} \bibinfo{volume}{189},
  \bibinfo{pages}{116023}.
%Type = Article
\bibitem[{Liu et~al.(2019)Liu, Nie, Gao, Gao, Han and Shao}]{liu2019flexible}
\bibinfo{author}{Liu, Y.}, \bibinfo{author}{Nie, F.}, \bibinfo{author}{Gao,
  Q.}, \bibinfo{author}{Gao, X.}, \bibinfo{author}{Han, J.},
  \bibinfo{author}{Shao, L.}, \bibinfo{year}{2019}.
\newblock \bibinfo{title}{Flexible unsupervised feature extraction for image
  classification}.
\newblock \bibinfo{journal}{Neural Netw.} \bibinfo{volume}{115},
  \bibinfo{pages}{65--71}.
%Type = Inproceedings
\bibitem[{Lobos-Tsunekawa and Harada(2020)}]{lobos2020point}
\bibinfo{author}{Lobos-Tsunekawa, K.}, \bibinfo{author}{Harada, T.},
  \bibinfo{year}{2020}.
\newblock \bibinfo{title}{Point cloud based reinforcement learning for
  sim-to-real and partial observability in visual navigation}, in:
  \bibinfo{booktitle}{IEEE/RSJ Int. Conf. Intell. Robots Syst.},
  \bibinfo{organization}{IEEE}. pp. \bibinfo{pages}{5871--5878}.
%Type = Inproceedings
\bibitem[{Lon{\v{c}}arcvi{\'c} et~al.(2021)Lon{\v{c}}arcvi{\'c}, Ude and
  Gams}]{lonvcarcvic2021accelerated}
\bibinfo{author}{Lon{\v{c}}arcvi{\'c}, Z.}, \bibinfo{author}{Ude, A.},
  \bibinfo{author}{Gams, A.}, \bibinfo{year}{2021}.
\newblock \bibinfo{title}{Accelerated robot skill acquisition by reinforcement
  learning-aided sim-to-real domain adaptation}, in: \bibinfo{booktitle}{20th
  Int. Conf. Adv. Robot.}, \bibinfo{organization}{IEEE}. pp.
  \bibinfo{pages}{269--274}.
%Type = Article
\bibitem[{Lu et~al.(2020)Lu, Xiao, Dai and Dai}]{9170648}
\bibinfo{author}{Lu, X.}, \bibinfo{author}{Xiao, L.}, \bibinfo{author}{Dai,
  C.}, \bibinfo{author}{Dai, H.}, \bibinfo{year}{2020}.
\newblock \bibinfo{title}{Uav-aided cellular communications with deep
  reinforcement learning against jamming}.
\newblock \bibinfo{journal}{IEEE Trans. Wirel. Commun.} \bibinfo{volume}{27},
  \bibinfo{pages}{48--53}.
%Type = Article
\bibitem[{Lu et~al.(2022)Lu, Xiao, Niu, Ji and Wang}]{9705557}
\bibinfo{author}{Lu, X.}, \bibinfo{author}{Xiao, L.}, \bibinfo{author}{Niu,
  G.}, \bibinfo{author}{Ji, X.}, \bibinfo{author}{Wang, Q.},
  \bibinfo{year}{2022}.
\newblock \bibinfo{title}{Safe exploration in wireless security: A safe
  reinforcement learning algorithm with hierarchical structure}.
\newblock \bibinfo{journal}{IEEE Trans. Inf. Forensics Secur.}
  \bibinfo{volume}{17}, \bibinfo{pages}{732--743}.
%Type = Article
\bibitem[{Luo et~al.(2022)Luo, Ni, Tian and Cheng}]{luo2022federated}
\bibinfo{author}{Luo, R.}, \bibinfo{author}{Ni, W.}, \bibinfo{author}{Tian,
  H.}, \bibinfo{author}{Cheng, J.}, \bibinfo{year}{2022}.
\newblock \bibinfo{title}{Federated deep reinforcement learning for
  {RIS}-assisted indoor multi-robot communication systems}.
\newblock \bibinfo{journal}{IEEE Trans. Veh. Technol.} .
%Type = Inproceedings
\bibitem[{Matas et~al.(2018)Matas, James and Davison}]{matas2018sim}
\bibinfo{author}{Matas, J.}, \bibinfo{author}{James, S.},
  \bibinfo{author}{Davison, A.J.}, \bibinfo{year}{2018}.
\newblock \bibinfo{title}{Sim-to-real reinforcement learning for deformable
  object manipulation}, in: \bibinfo{booktitle}{Conf. Robot Learn.},
  \bibinfo{organization}{PMLR}. pp. \bibinfo{pages}{734--743}.
%Type = Inproceedings
\bibitem[{McMahan et~al.(2017)McMahan, Moore, Ramage, Hampson and
  y~Arcas}]{mcmahan2017communication}
\bibinfo{author}{McMahan, B.}, \bibinfo{author}{Moore, E.},
  \bibinfo{author}{Ramage, D.}, \bibinfo{author}{Hampson, S.},
  \bibinfo{author}{y~Arcas, B.A.}, \bibinfo{year}{2017}.
\newblock \bibinfo{title}{Communication-efficient learning of deep networks
  from decentralized data}, in: \bibinfo{booktitle}{Artif. Intell. Stat.},
  \bibinfo{organization}{PMLR}. pp. \bibinfo{pages}{1273--1282}.
%Type = Article
\bibitem[{Melo and Lopes(2021)}]{melo2021teaching}
\bibinfo{author}{Melo, F.S.}, \bibinfo{author}{Lopes, M.},
  \bibinfo{year}{2021}.
\newblock \bibinfo{title}{Teaching multiple inverse reinforcement learners}.
\newblock \bibinfo{journal}{Front. Artif. Intell.} , \bibinfo{pages}{27}.
%Type = Article
\bibitem[{Mnih et~al.(2015)Mnih, Kavukcuoglu, Silver, Rusu, Veness, Bellemare,
  Graves, Riedmiller, Fidjeland, Ostrovski et~al.}]{mnih2015human}
\bibinfo{author}{Mnih, V.}, \bibinfo{author}{Kavukcuoglu, K.},
  \bibinfo{author}{Silver, D.}, \bibinfo{author}{Rusu, A.A.},
  \bibinfo{author}{Veness, J.}, \bibinfo{author}{Bellemare, M.G.},
  \bibinfo{author}{Graves, A.}, \bibinfo{author}{Riedmiller, M.},
  \bibinfo{author}{Fidjeland, A.K.}, \bibinfo{author}{Ostrovski, G.}, et~al.,
  \bibinfo{year}{2015}.
\newblock \bibinfo{title}{Human-level control through deep reinforcement
  learning}.
\newblock \bibinfo{journal}{nature} \bibinfo{volume}{518},
  \bibinfo{pages}{529--533}.
%Type = Article
\bibitem[{Mosqueira-Rey et~al.(2022)Mosqueira-Rey, Hern{\'a}ndez-Pereira,
  Alonso-R{\'\i}os, Bobes-Bascar{\'a}n and
  Fern{\'a}ndez-Leal}]{mosqueira2022human}
\bibinfo{author}{Mosqueira-Rey, E.}, \bibinfo{author}{Hern{\'a}ndez-Pereira,
  E.}, \bibinfo{author}{Alonso-R{\'\i}os, D.},
  \bibinfo{author}{Bobes-Bascar{\'a}n, J.},
  \bibinfo{author}{Fern{\'a}ndez-Leal, {\'A}.}, \bibinfo{year}{2022}.
\newblock \bibinfo{title}{Human-in-the-loop machine learning: A state of the
  art}.
\newblock \bibinfo{journal}{Artif. Intell. Rev.} , \bibinfo{pages}{1--50}.
%Type = Article
\bibitem[{Mou et~al.(2020)Mou, Wen and Chen}]{mou2020sample}
\bibinfo{author}{Mou, W.}, \bibinfo{author}{Wen, Z.}, \bibinfo{author}{Chen,
  X.}, \bibinfo{year}{2020}.
\newblock \bibinfo{title}{On the sample complexity of reinforcement learning
  with policy space generalization}.
\newblock \bibinfo{journal}{arXiv preprint arXiv:2008.07353} .
%Type = Article
\bibitem[{Muratore et~al.(2021)Muratore, Eilers, Gienger and
  Peters}]{muratore2021data}
\bibinfo{author}{Muratore, F.}, \bibinfo{author}{Eilers, C.},
  \bibinfo{author}{Gienger, M.}, \bibinfo{author}{Peters, J.},
  \bibinfo{year}{2021}.
\newblock \bibinfo{title}{Data-efficient domain randomization with bayesian
  optimization}.
\newblock \bibinfo{journal}{IEEE Robot. Autom. Lett.} \bibinfo{volume}{6},
  \bibinfo{pages}{911--918}.
%Type = Article
\bibitem[{Nagabandi et~al.(2018)Nagabandi, Clavera, Liu, Fearing, Abbeel,
  Levine and Finn}]{nagabandi2018learning}
\bibinfo{author}{Nagabandi, A.}, \bibinfo{author}{Clavera, I.},
  \bibinfo{author}{Liu, S.}, \bibinfo{author}{Fearing, R.S.},
  \bibinfo{author}{Abbeel, P.}, \bibinfo{author}{Levine, S.},
  \bibinfo{author}{Finn, C.}, \bibinfo{year}{2018}.
\newblock \bibinfo{title}{Learning to adapt in dynamic, real-world environments
  through meta-reinforcement learning}.
\newblock \bibinfo{journal}{arXiv preprint arXiv:1803.11347} .
%Type = Article
\bibitem[{Najar and Chetouani(2021)}]{najar2021reinforcement}
\bibinfo{author}{Najar, A.}, \bibinfo{author}{Chetouani, M.},
  \bibinfo{year}{2021}.
\newblock \bibinfo{title}{Reinforcement learning with human advice: a survey}.
\newblock \bibinfo{journal}{Front. Robot. AI} \bibinfo{volume}{8},
  \bibinfo{pages}{584075}.
%Type = Article
\bibitem[{Navidi and Landry~Jr(2021)}]{navidi2021new}
\bibinfo{author}{Navidi, N.}, \bibinfo{author}{Landry~Jr, R.},
  \bibinfo{year}{2021}.
\newblock \bibinfo{title}{New approach in human-{AI} interaction by
  reinforcement-imitation learning}.
\newblock \bibinfo{journal}{Appl. Sci.} \bibinfo{volume}{11},
  \bibinfo{pages}{3068}.
%Type = Article
\bibitem[{Neumeyer et~al.(2021)Neumeyer, Oliehoek and
  Gavrila}]{neumeyer2021general}
\bibinfo{author}{Neumeyer, C.}, \bibinfo{author}{Oliehoek, F.A.},
  \bibinfo{author}{Gavrila, D.M.}, \bibinfo{year}{2021}.
\newblock \bibinfo{title}{General-sum multi-agent continuous inverse optimal
  control}.
\newblock \bibinfo{journal}{IEEE Robot. Autom. Lett.} \bibinfo{volume}{6},
  \bibinfo{pages}{3429--3436}.
%Type = Inproceedings
\bibitem[{Ng et~al.(1999)Ng, Harada and Russell}]{ng1999policy}
\bibinfo{author}{Ng, A.Y.}, \bibinfo{author}{Harada, D.},
  \bibinfo{author}{Russell, S.}, \bibinfo{year}{1999}.
\newblock \bibinfo{title}{Policy invariance under reward transformations:
  Theory and application to reward shaping}, in: \bibinfo{booktitle}{ICML}, pp.
  \bibinfo{pages}{278--287}.
%Type = Inproceedings
\bibitem[{Ng et~al.(2000)Ng, Russell et~al.}]{ng2000algorithms}
\bibinfo{author}{Ng, A.Y.}, \bibinfo{author}{Russell, S.}, et~al.,
  \bibinfo{year}{2000}.
\newblock \bibinfo{title}{Algorithms for inverse reinforcement learning.}, in:
  \bibinfo{booktitle}{ICML}, p.~\bibinfo{pages}{2}.
%Type = Inproceedings
\bibitem[{Nguyen et~al.(2019)Nguyen, La and Deans}]{nguyen2019hindsight}
\bibinfo{author}{Nguyen, H.}, \bibinfo{author}{La, H.M.},
  \bibinfo{author}{Deans, M.}, \bibinfo{year}{2019}.
\newblock \bibinfo{title}{Hindsight experience replay with experience ranking},
  in: \bibinfo{booktitle}{Joint IEEE 9th Int. Conf. Develop. Learn. Epigenetic
  Robot.}, \bibinfo{organization}{IEEE}. pp. \bibinfo{pages}{1--6}.
%Type = Inproceedings
\bibitem[{Nguyen et~al.(2018)Nguyen, Akiyama and Ohashi}]{nguyen2018experience}
\bibinfo{author}{Nguyen, P.X.}, \bibinfo{author}{Akiyama, T.},
  \bibinfo{author}{Ohashi, H.}, \bibinfo{year}{2018}.
\newblock \bibinfo{title}{Experience filtering for robot navigation using deep
  reinforcement learning.}, in: \bibinfo{booktitle}{ICAART (2)}, pp.
  \bibinfo{pages}{243--249}.
%Type = Article
\bibitem[{Ni et~al.(2020)Ni, Sikchi, Wang, Gupta, Lee and Eysenbach}]{ni2020f}
\bibinfo{author}{Ni, T.}, \bibinfo{author}{Sikchi, H.}, \bibinfo{author}{Wang,
  Y.}, \bibinfo{author}{Gupta, T.}, \bibinfo{author}{Lee, L.},
  \bibinfo{author}{Eysenbach, B.}, \bibinfo{year}{2020}.
\newblock \bibinfo{title}{f-{IRL}: Inverse reinforcement learning via state
  marginal matching}.
\newblock \bibinfo{journal}{arXiv preprint arXiv:2011.04709} .
%Type = Article
\bibitem[{Nian et~al.(2020)Nian, Liu and Huang}]{NIAN2020106886}
\bibinfo{author}{Nian, R.}, \bibinfo{author}{Liu, J.}, \bibinfo{author}{Huang,
  B.}, \bibinfo{year}{2020}.
\newblock \bibinfo{title}{A review on reinforcement learning: Introduction and
  applications in industrial process control}.
\newblock \bibinfo{journal}{Comput. Chem. Eng.} \bibinfo{volume}{139},
  \bibinfo{pages}{106886}.
%Type = Inproceedings
\bibitem[{Nishi and Shimosaka(2020)}]{nishi2020fine}
\bibinfo{author}{Nishi, K.}, \bibinfo{author}{Shimosaka, M.},
  \bibinfo{year}{2020}.
\newblock \bibinfo{title}{Fine-grained driving behavior prediction via
  context-aware multi-task inverse reinforcement learning}, in:
  \bibinfo{booktitle}{IEEE Int. Conf. Robot. Autom.},
  \bibinfo{organization}{IEEE}. pp. \bibinfo{pages}{2281--2287}.
%Type = Article
\bibitem[{Olson et~al.(2021)Olson, Khanna, Neal, Li and
  Wong}]{olson2021counterfactual}
\bibinfo{author}{Olson, M.L.}, \bibinfo{author}{Khanna, R.},
  \bibinfo{author}{Neal, L.}, \bibinfo{author}{Li, F.}, \bibinfo{author}{Wong,
  W.K.}, \bibinfo{year}{2021}.
\newblock \bibinfo{title}{Counterfactual state explanations for reinforcement
  learning agents via generative deep learning}.
\newblock \bibinfo{journal}{Artif. Intell.} \bibinfo{volume}{295},
  \bibinfo{pages}{103455}.
%Type = Article
\bibitem[{Ouyang et~al.(2022)Ouyang, Wu, Jiang, Almeida, Wainwright, Mishkin,
  Zhang, Agarwal, Slama, Ray et~al.}]{ouyang2022training}
\bibinfo{author}{Ouyang, L.}, \bibinfo{author}{Wu, J.}, \bibinfo{author}{Jiang,
  X.}, \bibinfo{author}{Almeida, D.}, \bibinfo{author}{Wainwright, C.L.},
  \bibinfo{author}{Mishkin, P.}, \bibinfo{author}{Zhang, C.},
  \bibinfo{author}{Agarwal, S.}, \bibinfo{author}{Slama, K.},
  \bibinfo{author}{Ray, A.}, et~al., \bibinfo{year}{2022}.
\newblock \bibinfo{title}{Training language models to follow instructions with
  human feedback}.
\newblock \bibinfo{journal}{arXiv preprint arXiv:2203.02155} .
%Type = Inproceedings
\bibitem[{Park et~al.(2021)Park, Lee and Suh}]{park2021sim}
\bibinfo{author}{Park, Y.}, \bibinfo{author}{Lee, S.H.}, \bibinfo{author}{Suh,
  I.H.}, \bibinfo{year}{2021}.
\newblock \bibinfo{title}{Sim-to-real visual grasping via state representation
  learning based on combining pixel-level and feature-level domain adaptation},
  in: \bibinfo{booktitle}{IEEE Int. Conf. Robot. Autom.},
  \bibinfo{organization}{IEEE}. pp. \bibinfo{pages}{6300--6307}.
%Type = Article
\bibitem[{Parras et~al.(2022)Parras, Almodóvar, Apellániz and Zazo}]{9844128}
\bibinfo{author}{Parras, J.}, \bibinfo{author}{Almodóvar, A.},
  \bibinfo{author}{Apellániz, P.A.}, \bibinfo{author}{Zazo, S.},
  \bibinfo{year}{2022}.
\newblock \bibinfo{title}{Inverse reinforcement learning: A new framework to
  mitigate an intelligent backoff attack}.
\newblock \bibinfo{journal}{IEEE Internet Things J.} \bibinfo{volume}{9},
  \bibinfo{pages}{24790--24799}.
%Type = Inproceedings
\bibitem[{Pathak et~al.(2017)Pathak, Agrawal, Efros and
  Darrell}]{pathak2017curiosity}
\bibinfo{author}{Pathak, D.}, \bibinfo{author}{Agrawal, P.},
  \bibinfo{author}{Efros, A.A.}, \bibinfo{author}{Darrell, T.},
  \bibinfo{year}{2017}.
\newblock \bibinfo{title}{Curiosity-driven exploration by self-supervised
  prediction}, in: \bibinfo{booktitle}{ICML}, \bibinfo{organization}{PMLR}. pp.
  \bibinfo{pages}{2778--2787}.
%Type = Article
\bibitem[{Pohlen et~al.(2018)Pohlen, Piot, Hester, Azar, Horgan, Budden,
  Barth-Maron, Van~Hasselt, Quan, Ve{\v{c}}er{\'\i}k
  et~al.}]{pohlen2018observe}
\bibinfo{author}{Pohlen, T.}, \bibinfo{author}{Piot, B.},
  \bibinfo{author}{Hester, T.}, \bibinfo{author}{Azar, M.G.},
  \bibinfo{author}{Horgan, D.}, \bibinfo{author}{Budden, D.},
  \bibinfo{author}{Barth-Maron, G.}, \bibinfo{author}{Van~Hasselt, H.},
  \bibinfo{author}{Quan, J.}, \bibinfo{author}{Ve{\v{c}}er{\'\i}k, M.}, et~al.,
  \bibinfo{year}{2018}.
\newblock \bibinfo{title}{Observe and look further: Achieving consistent
  performance on atari}.
\newblock \bibinfo{journal}{arXiv preprint arXiv:1805.11593} .
%Type = Article
\bibitem[{Qi et~al.(2021)Qi, Zhou, Lei and Zheng}]{qi2021federated}
\bibinfo{author}{Qi, J.}, \bibinfo{author}{Zhou, Q.}, \bibinfo{author}{Lei,
  L.}, \bibinfo{author}{Zheng, K.}, \bibinfo{year}{2021}.
\newblock \bibinfo{title}{Federated reinforcement learning: techniques,
  applications, and open challenges}.
\newblock \bibinfo{journal}{arXiv preprint arXiv:2108.11887} .
%Type = Article
\bibitem[{Qiang et~al.(2023)Qiang, Zhu, Li, Zhu, Yuan and
  Wu}]{qiang2023natural}
\bibinfo{author}{Qiang, J.}, \bibinfo{author}{Zhu, S.}, \bibinfo{author}{Li,
  Y.}, \bibinfo{author}{Zhu, Y.}, \bibinfo{author}{Yuan, Y.},
  \bibinfo{author}{Wu, X.}, \bibinfo{year}{2023}.
\newblock \bibinfo{title}{Natural language watermarking via paraphraser-based
  lexical substitution}.
\newblock \bibinfo{journal}{Artif. Intell.} \bibinfo{volume}{317},
  \bibinfo{pages}{103859}.
%Type = Inproceedings
\bibitem[{Ramachandran and Amir(2007)}]{ramachandran2007bayesian}
\bibinfo{author}{Ramachandran, D.}, \bibinfo{author}{Amir, E.},
  \bibinfo{year}{2007}.
\newblock \bibinfo{title}{Bayesian inverse reinforcement learning}, in:
  \bibinfo{booktitle}{IJCAI}, pp. \bibinfo{pages}{2586--2591}.
%Type = Inproceedings
\bibitem[{Ramicic and Bonarini(2017)}]{ramicic2017entropy}
\bibinfo{author}{Ramicic, M.}, \bibinfo{author}{Bonarini, A.},
  \bibinfo{year}{2017}.
\newblock \bibinfo{title}{Entropy-based prioritized sampling in deep
  {Q}-learning}, in: \bibinfo{booktitle}{2nd Int. Conf. Image, Vision Comput.},
  \bibinfo{organization}{IEEE}. pp. \bibinfo{pages}{1068--1072}.
%Type = Inproceedings
\bibitem[{Ramponi et~al.(2020)Ramponi, Likmeta, Metelli, Tirinzoni and
  Restelli}]{ramponi2020truly}
\bibinfo{author}{Ramponi, G.}, \bibinfo{author}{Likmeta, A.},
  \bibinfo{author}{Metelli, A.M.}, \bibinfo{author}{Tirinzoni, A.},
  \bibinfo{author}{Restelli, M.}, \bibinfo{year}{2020}.
\newblock \bibinfo{title}{Truly batch model-free inverse reinforcement learning
  about multiple intentions}, in: \bibinfo{booktitle}{Int. Conf. Artif. Intell.
  Stat.}, \bibinfo{organization}{PMLR}. pp. \bibinfo{pages}{2359--2369}.
%Type = Article
\bibitem[{Rezazadeh and Bartzoudis(2022)}]{rezazadeh2022federated}
\bibinfo{author}{Rezazadeh, F.}, \bibinfo{author}{Bartzoudis, N.},
  \bibinfo{year}{2022}.
\newblock \bibinfo{title}{A federated drl approach for smart micro-grid energy
  control with distributed energy resources}.
\newblock \bibinfo{journal}{arXiv preprint arXiv:2211.03430} .
%Type = Inproceedings
\bibitem[{Russell(1998)}]{russell1998learning}
\bibinfo{author}{Russell, S.}, \bibinfo{year}{1998}.
\newblock \bibinfo{title}{Learning agents for uncertain environments}, in:
  \bibinfo{booktitle}{Proc. Eleventh Ann. Conf. Comput. Learn. Theory}, pp.
  \bibinfo{pages}{101--103}.
%Type = Book
\bibitem[{Russell(2010)}]{russell2010artificial}
\bibinfo{author}{Russell, S.J.}, \bibinfo{year}{2010}.
\newblock \bibinfo{title}{Artificial intelligence: A modern approach}.
\newblock \bibinfo{publisher}{Pearson Education, Inc.}
%Type = Inproceedings
\bibitem[{Sasaki et~al.(2018)Sasaki, Yohira and Kawaguchi}]{sasaki2018sample}
\bibinfo{author}{Sasaki, F.}, \bibinfo{author}{Yohira, T.},
  \bibinfo{author}{Kawaguchi, A.}, \bibinfo{year}{2018}.
\newblock \bibinfo{title}{Sample efficient imitation learning for continuous
  control}, in: \bibinfo{booktitle}{Int. Conf. Learn. Rep.}
%Type = Article
\bibitem[{Schafer et~al.(2022)Schafer, Wikle and Hooten}]{schafer2022bayesian}
\bibinfo{author}{Schafer, T.L.}, \bibinfo{author}{Wikle, C.K.},
  \bibinfo{author}{Hooten, M.B.}, \bibinfo{year}{2022}.
\newblock \bibinfo{title}{Bayesian inverse reinforcement learning for
  collective animal movement}.
\newblock \bibinfo{journal}{Ann. Appl. Stat.} \bibinfo{volume}{16},
  \bibinfo{pages}{999--1013}.
%Type = Article
\bibitem[{Schaul et~al.(2015)Schaul, Quan, Antonoglou and
  Silver}]{schaul2015prioritized}
\bibinfo{author}{Schaul, T.}, \bibinfo{author}{Quan, J.},
  \bibinfo{author}{Antonoglou, I.}, \bibinfo{author}{Silver, D.},
  \bibinfo{year}{2015}.
\newblock \bibinfo{title}{Prioritized experience replay}.
\newblock \bibinfo{journal}{arXiv preprint arXiv:1511.05952} .
%Type = Article
\bibitem[{Schraner(2022)}]{schraner2022teacher}
\bibinfo{author}{Schraner, Y.}, \bibinfo{year}{2022}.
\newblock \bibinfo{title}{Teacher-student curriculum learning for reinforcement
  learning}.
\newblock \bibinfo{journal}{arXiv preprint arXiv:2210.17368} .
%Type = Inproceedings
\bibitem[{Schulman et~al.(2015)Schulman, Levine, Abbeel, Jordan and
  Moritz}]{schulman2015trust}
\bibinfo{author}{Schulman, J.}, \bibinfo{author}{Levine, S.},
  \bibinfo{author}{Abbeel, P.}, \bibinfo{author}{Jordan, M.},
  \bibinfo{author}{Moritz, P.}, \bibinfo{year}{2015}.
\newblock \bibinfo{title}{Trust region policy optimization}, in:
  \bibinfo{booktitle}{ICML}, \bibinfo{organization}{PMLR}. pp.
  \bibinfo{pages}{1889--1897}.
%Type = Article
\bibitem[{Schulman et~al.(2017)Schulman, Wolski, Dhariwal, Radford and
  Klimov}]{schulman2017proximal}
\bibinfo{author}{Schulman, J.}, \bibinfo{author}{Wolski, F.},
  \bibinfo{author}{Dhariwal, P.}, \bibinfo{author}{Radford, A.},
  \bibinfo{author}{Klimov, O.}, \bibinfo{year}{2017}.
\newblock \bibinfo{title}{Proximal policy optimization algorithms}.
\newblock \bibinfo{journal}{arXiv preprint arXiv:1707.06347} .
%Type = Article
\bibitem[{Shi et~al.(2022)Shi, Tong, Liu and Fan}]{shi2022knowledge}
\bibinfo{author}{Shi, D.}, \bibinfo{author}{Tong, J.}, \bibinfo{author}{Liu,
  Y.}, \bibinfo{author}{Fan, W.}, \bibinfo{year}{2022}.
\newblock \bibinfo{title}{Knowledge reuse of multi-agent reinforcement learning
  in cooperative tasks}.
\newblock \bibinfo{journal}{Entropy} \bibinfo{volume}{24},
  \bibinfo{pages}{470}.
%Type = Article
\bibitem[{Shi et~al.(2021)Shi, Li, Mao and Hwang}]{shi2021lateral}
\bibinfo{author}{Shi, H.}, \bibinfo{author}{Li, J.}, \bibinfo{author}{Mao, J.},
  \bibinfo{author}{Hwang, K.S.}, \bibinfo{year}{2021}.
\newblock \bibinfo{title}{Lateral transfer learning for multiagent
  reinforcement learning}.
\newblock \bibinfo{journal}{IEEE Trans. Cybern.} .
%Type = Article
\bibitem[{Shu et~al.(2022)Shu, Liu, Mu and Cao}]{9583858}
\bibinfo{author}{Shu, H.}, \bibinfo{author}{Liu, T.}, \bibinfo{author}{Mu, X.},
  \bibinfo{author}{Cao, D.}, \bibinfo{year}{2022}.
\newblock \bibinfo{title}{Driving tasks transfer using deep reinforcement
  learning for decision-making of autonomous vehicles in unsignalized
  intersection}.
\newblock \bibinfo{journal}{IEEE Trans. Veh. Technol.} \bibinfo{volume}{71},
  \bibinfo{pages}{41--52}.
%Type = Inproceedings
\bibitem[{Sinha et~al.(2022)Sinha, Song, Garg and Ermon}]{sinha2022experience}
\bibinfo{author}{Sinha, S.}, \bibinfo{author}{Song, J.}, \bibinfo{author}{Garg,
  A.}, \bibinfo{author}{Ermon, S.}, \bibinfo{year}{2022}.
\newblock \bibinfo{title}{Experience replay with likelihood-free importance
  weights}, in: \bibinfo{booktitle}{Learn. Dynamics Control Conf.},
  \bibinfo{organization}{PMLR}. pp. \bibinfo{pages}{110--123}.
%Type = Article
\bibitem[{Snoswell et~al.(2021)Snoswell, Singh and Ye}]{snoswell2021limiirl}
\bibinfo{author}{Snoswell, A.J.}, \bibinfo{author}{Singh, S.P.},
  \bibinfo{author}{Ye, N.}, \bibinfo{year}{2021}.
\newblock \bibinfo{title}{{LiMIIRL}: Lightweight multiple-intent inverse
  reinforcement learning}.
\newblock \bibinfo{journal}{arXiv preprint arXiv:2106.01777} .
%Type = Inproceedings
\bibitem[{Strehl et~al.(2006)Strehl, Li, Wiewiora, Langford and
  Littman}]{strehl2006pac}
\bibinfo{author}{Strehl, A.L.}, \bibinfo{author}{Li, L.},
  \bibinfo{author}{Wiewiora, E.}, \bibinfo{author}{Langford, J.},
  \bibinfo{author}{Littman, M.L.}, \bibinfo{year}{2006}.
\newblock \bibinfo{title}{{PAC} model-free reinforcement learning}, in:
  \bibinfo{booktitle}{ICML}, pp. \bibinfo{pages}{881--888}.
%Type = Article
\bibitem[{Su et~al.(2021)Su, Tang, Jiang, Lu, Ge, Song, Xiong, Sun and
  Luo}]{su2021enhanced}
\bibinfo{author}{Su, J.}, \bibinfo{author}{Tang, J.}, \bibinfo{author}{Jiang,
  H.}, \bibinfo{author}{Lu, Z.}, \bibinfo{author}{Ge, Y.},
  \bibinfo{author}{Song, L.}, \bibinfo{author}{Xiong, D.},
  \bibinfo{author}{Sun, L.}, \bibinfo{author}{Luo, J.}, \bibinfo{year}{2021}.
\newblock \bibinfo{title}{Enhanced aspect-based sentiment analysis models with
  progressive self-supervised attention learning}.
\newblock \bibinfo{journal}{Artif. Intell.} \bibinfo{volume}{296},
  \bibinfo{pages}{103477}.
%Type = Article
\bibitem[{Sun and Yang(2022)}]{sun2022low}
\bibinfo{author}{Sun, P.}, \bibinfo{author}{Yang, L.}, \bibinfo{year}{2022}.
\newblock \bibinfo{title}{Low-rank supervised and semi-supervised multi-metric
  learning for classification}.
\newblock \bibinfo{journal}{Knowl-Based Syst.} \bibinfo{volume}{236},
  \bibinfo{pages}{107787}.
%Type = Article
\bibitem[{Sutton(1995)}]{sutton1995generalization}
\bibinfo{author}{Sutton, R.S.}, \bibinfo{year}{1995}.
\newblock \bibinfo{title}{Generalization in reinforcement learning: Successful
  examples using sparse coarse coding}.
\newblock \bibinfo{journal}{Adv. Neural Inf. Process Syst.}
  \bibinfo{volume}{8}.
%Type = Book
\bibitem[{Sutton and Barto(2018)}]{sutton2018reinforcement}
\bibinfo{author}{Sutton, R.S.}, \bibinfo{author}{Barto, A.G.},
  \bibinfo{year}{2018}.
\newblock \bibinfo{title}{Reinforcement learning: An introduction}.
\newblock \bibinfo{publisher}{MIT press}.
%Type = Inproceedings
\bibitem[{Tao et~al.(2021)Tao, Genc, Chung, Sun and Mallya}]{tao2021repaint}
\bibinfo{author}{Tao, Y.}, \bibinfo{author}{Genc, S.}, \bibinfo{author}{Chung,
  J.}, \bibinfo{author}{Sun, T.}, \bibinfo{author}{Mallya, S.},
  \bibinfo{year}{2021}.
\newblock \bibinfo{title}{Repaint: Knowledge transfer in deep reinforcement
  learning}, in: \bibinfo{booktitle}{ICML}, \bibinfo{organization}{PMLR}. pp.
  \bibinfo{pages}{10141--10152}.
%Type = Article
\bibitem[{Taylor et~al.(2021)Taylor, Nissen, Wang and
  Navidi}]{taylor2021improving}
\bibinfo{author}{Taylor, M.E.}, \bibinfo{author}{Nissen, N.},
  \bibinfo{author}{Wang, Y.}, \bibinfo{author}{Navidi, N.},
  \bibinfo{year}{2021}.
\newblock \bibinfo{title}{Improving reinforcement learning with human
  assistance: an argument for human subject studies with {HIPPO} gym}.
\newblock \bibinfo{journal}{Neural Comput. Appl.} , \bibinfo{pages}{1--11}.
%Type = Article
\bibitem[{Tiddi and Schlobach(2022)}]{tiddi2022knowledge}
\bibinfo{author}{Tiddi, I.}, \bibinfo{author}{Schlobach, S.},
  \bibinfo{year}{2022}.
\newblock \bibinfo{title}{Knowledge graphs as tools for explainable machine
  learning: A survey}.
\newblock \bibinfo{journal}{Artif. Intell.} \bibinfo{volume}{302},
  \bibinfo{pages}{103627}.
%Type = Book
\bibitem[{Tobin(2019)}]{tobin2019real}
\bibinfo{author}{Tobin, J.P.}, \bibinfo{year}{2019}.
\newblock \bibinfo{title}{Real-World Robotic Perception and Control Using
  Synthetic Data}.
\newblock \bibinfo{publisher}{University of California, Berkeley}.
%Type = Inproceedings
\bibitem[{Tosatto et~al.(2017)Tosatto, Pirotta, d’Eramo and
  Restelli}]{tosatto2017boosted}
\bibinfo{author}{Tosatto, S.}, \bibinfo{author}{Pirotta, M.},
  \bibinfo{author}{d’Eramo, C.}, \bibinfo{author}{Restelli, M.},
  \bibinfo{year}{2017}.
\newblock \bibinfo{title}{Boosted fitted q-iteration}, in:
  \bibinfo{booktitle}{ICML}, \bibinfo{organization}{PMLR}. pp.
  \bibinfo{pages}{3434--3443}.
%Type = Inproceedings
\bibitem[{Van~Hasselt et~al.(2016)Van~Hasselt, Guez and Silver}]{van2016deep}
\bibinfo{author}{Van~Hasselt, H.}, \bibinfo{author}{Guez, A.},
  \bibinfo{author}{Silver, D.}, \bibinfo{year}{2016}.
\newblock \bibinfo{title}{Deep reinforcement learning with double q-learning},
  in: \bibinfo{booktitle}{Proc. AAAI Conf. Artif. Intell.}
%Type = Article
\bibitem[{Vecerik et~al.(2017)Vecerik, Hester, Scholz, Wang, Pietquin, Piot,
  Heess, Roth{\"o}rl, Lampe and Riedmiller}]{vecerik2017leveraging}
\bibinfo{author}{Vecerik, M.}, \bibinfo{author}{Hester, T.},
  \bibinfo{author}{Scholz, J.}, \bibinfo{author}{Wang, F.},
  \bibinfo{author}{Pietquin, O.}, \bibinfo{author}{Piot, B.},
  \bibinfo{author}{Heess, N.}, \bibinfo{author}{Roth{\"o}rl, T.},
  \bibinfo{author}{Lampe, T.}, \bibinfo{author}{Riedmiller, M.},
  \bibinfo{year}{2017}.
\newblock \bibinfo{title}{Leveraging demonstrations for deep reinforcement
  learning on robotics problems with sparse rewards}.
\newblock \bibinfo{journal}{arXiv preprint arXiv:1707.08817} .
%Type = Book
\bibitem[{Venuto(2020)}]{venuto2020robust}
\bibinfo{author}{Venuto, D.}, \bibinfo{year}{2020}.
\newblock \bibinfo{title}{Robust Adversarial Inverse Reinforcement Learning
  with Temporally Extended Actions}.
\newblock \bibinfo{publisher}{McGill University (Canada)}.
%Type = Article
\bibitem[{Verma et~al.(2022)Verma, Kharkwal and Kambhampati}]{verma2022advice}
\bibinfo{author}{Verma, M.}, \bibinfo{author}{Kharkwal, A.},
  \bibinfo{author}{Kambhampati, S.}, \bibinfo{year}{2022}.
\newblock \bibinfo{title}{Advice conformance verification by reinforcement
  learning agents for human-in-the-loop}.
\newblock \bibinfo{journal}{arXiv preprint arXiv:2210.03455} .
%Type = Inproceedings
\bibitem[{Wang et~al.(2021)Wang, Li and Chan}]{wang2021meta}
\bibinfo{author}{Wang, P.}, \bibinfo{author}{Li, H.}, \bibinfo{author}{Chan,
  C.Y.}, \bibinfo{year}{2021}.
\newblock \bibinfo{title}{Meta-adversarial inverse reinforcement learning for
  decision-making tasks}, in: \bibinfo{booktitle}{IEEE Int. Conf. Robotics
  Autom.}, \bibinfo{organization}{IEEE}. pp. \bibinfo{pages}{12632--12638}.
%Type = Article
\bibitem[{Wang et~al.(2023)Wang, Yang, Li and Kan}]{9679819}
\bibinfo{author}{Wang, S.}, \bibinfo{author}{Yang, R.}, \bibinfo{author}{Li,
  B.}, \bibinfo{author}{Kan, Z.}, \bibinfo{year}{2023}.
\newblock \bibinfo{title}{Structural parameter space exploration for
  reinforcement learning via a matrix variate distribution}.
\newblock \bibinfo{journal}{IEEE Trans. Emerg. Top. Comput. Intell.}
  \bibinfo{volume}{7}, \bibinfo{pages}{1025--1035}.
%Type = Article
\bibitem[{Wang et~al.(2016a)Wang, Bapst, Heess, Mnih, Munos, Kavukcuoglu and
  de~Freitas}]{wang2016sample}
\bibinfo{author}{Wang, Z.}, \bibinfo{author}{Bapst, V.},
  \bibinfo{author}{Heess, N.}, \bibinfo{author}{Mnih, V.},
  \bibinfo{author}{Munos, R.}, \bibinfo{author}{Kavukcuoglu, K.},
  \bibinfo{author}{de~Freitas, N.}, \bibinfo{year}{2016}a.
\newblock \bibinfo{title}{Sample efficient actor-critic with experience
  replay}.
\newblock \bibinfo{journal}{arXiv preprint arXiv:1611.01224} .
%Type = Inproceedings
\bibitem[{Wang et~al.(2016b)Wang, Schaul, Hessel, Hasselt, Lanctot and
  Freitas}]{wang2016dueling}
\bibinfo{author}{Wang, Z.}, \bibinfo{author}{Schaul, T.},
  \bibinfo{author}{Hessel, M.}, \bibinfo{author}{Hasselt, H.},
  \bibinfo{author}{Lanctot, M.}, \bibinfo{author}{Freitas, N.},
  \bibinfo{year}{2016}b.
\newblock \bibinfo{title}{Dueling network architectures for deep reinforcement
  learning}, in: \bibinfo{booktitle}{ICML}, \bibinfo{organization}{PMLR}. pp.
  \bibinfo{pages}{1995--2003}.
%Type = Article
\bibitem[{Wang and Taylor(2018)}]{wang2018interactive}
\bibinfo{author}{Wang, Z.}, \bibinfo{author}{Taylor, M.E.},
  \bibinfo{year}{2018}.
\newblock \bibinfo{title}{Interactive reinforcement learning with dynamic reuse
  of prior knowledge from human/agent's demonstration}.
\newblock \bibinfo{journal}{arXiv preprint arXiv:1805.04493} .
%Type = Article
\bibitem[{Watkins and Dayan(1992)}]{watkins1992q}
\bibinfo{author}{Watkins, C.J.}, \bibinfo{author}{Dayan, P.},
  \bibinfo{year}{1992}.
\newblock \bibinfo{title}{Q-learning}.
\newblock \bibinfo{journal}{Mach. Learn.} \bibinfo{volume}{8},
  \bibinfo{pages}{279--292}.
%Type = Inproceedings
\bibitem[{Wiewiora et~al.(2003)Wiewiora, Cottrell and
  Elkan}]{wiewiora2003principled}
\bibinfo{author}{Wiewiora, E.}, \bibinfo{author}{Cottrell, G.W.},
  \bibinfo{author}{Elkan, C.}, \bibinfo{year}{2003}.
\newblock \bibinfo{title}{Principled methods for advising reinforcement
  learning agents}, in: \bibinfo{booktitle}{Proc. 20th Int. Conf. Mach.
  Learn.}, pp. \bibinfo{pages}{792--799}.
%Type = Article
\bibitem[{Williams(1992)}]{williams1992simple}
\bibinfo{author}{Williams, R.J.}, \bibinfo{year}{1992}.
\newblock \bibinfo{title}{Simple statistical gradient-following algorithms for
  connectionist reinforcement learning}.
\newblock \bibinfo{journal}{Mach. learn.} \bibinfo{volume}{8},
  \bibinfo{pages}{229--256}.
%Type = Inproceedings
\bibitem[{Wu et~al.(2020a)Wu, Xu, He, Gupta and Allen}]{wu2020squirl}
\bibinfo{author}{Wu, B.}, \bibinfo{author}{Xu, F.}, \bibinfo{author}{He, Z.},
  \bibinfo{author}{Gupta, A.}, \bibinfo{author}{Allen, P.K.},
  \bibinfo{year}{2020}a.
\newblock \bibinfo{title}{{SQUIRL}: Robust and efficient learning from video
  demonstration of long-horizon robotic manipulation tasks}, in:
  \bibinfo{booktitle}{IEEE/RSJ Int. Conf. Intell. Robots Syst.},
  \bibinfo{organization}{IEEE}. pp. \bibinfo{pages}{9720--9727}.
%Type = Article
\bibitem[{Wu et~al.(2020b)Wu, Sun, Zhan, Yang and Tomizuka}]{wu2020efficient}
\bibinfo{author}{Wu, Z.}, \bibinfo{author}{Sun, L.}, \bibinfo{author}{Zhan,
  W.}, \bibinfo{author}{Yang, C.}, \bibinfo{author}{Tomizuka, M.},
  \bibinfo{year}{2020}b.
\newblock \bibinfo{title}{Efficient sampling-based maximum entropy inverse
  reinforcement learning with application to autonomous driving}.
\newblock \bibinfo{journal}{IEEE Robotics Autom. Lett.} \bibinfo{volume}{5},
  \bibinfo{pages}{5355--5362}.
%Type = Article
\bibitem[{Xia et~al.(2019)Xia, Zhang, Bai, Zhou and Pan}]{8742593}
\bibinfo{author}{Xia, S.M.}, \bibinfo{author}{Zhang, L.}, \bibinfo{author}{Bai,
  W.}, \bibinfo{author}{Zhou, X.Y.}, \bibinfo{author}{Pan, Z.S.},
  \bibinfo{year}{2019}.
\newblock \bibinfo{title}{Ddos traffic control using transfer learning dqn with
  structure information}.
\newblock \bibinfo{journal}{IEEE Access} \bibinfo{volume}{7},
  \bibinfo{pages}{81481--81493}.
%Type = Article
\bibitem[{Xue et~al.(2023)Xue, Lian, Fan, Kolaric, Chai and Lewis}]{9537731}
\bibinfo{author}{Xue, W.}, \bibinfo{author}{Lian, B.}, \bibinfo{author}{Fan,
  J.}, \bibinfo{author}{Kolaric, P.}, \bibinfo{author}{Chai, T.},
  \bibinfo{author}{Lewis, F.L.}, \bibinfo{year}{2023}.
\newblock \bibinfo{title}{Inverse reinforcement q-learning through expert
  imitation for discrete-time systems}.
\newblock \bibinfo{journal}{IEEE Trans. Neural Netw. Learn. Syst.}
  \bibinfo{volume}{34}, \bibinfo{pages}{2386--2399}.
%Type = Article
\bibitem[{Yang et~al.(2019)Yang, Petersen, Zha and Faissol}]{yang2019single}
\bibinfo{author}{Yang, J.}, \bibinfo{author}{Petersen, B.},
  \bibinfo{author}{Zha, H.}, \bibinfo{author}{Faissol, D.},
  \bibinfo{year}{2019}.
\newblock \bibinfo{title}{Single episode policy transfer in reinforcement
  learning}.
\newblock \bibinfo{journal}{arXiv preprint arXiv:1910.07719} .
%Type = Article
\bibitem[{Yang et~al.(2020)Yang, Hao, Meng, Zhang, Hu, Cheng, Fan, Wang, Liu,
  Wang et~al.}]{yang2020efficient}
\bibinfo{author}{Yang, T.}, \bibinfo{author}{Hao, J.}, \bibinfo{author}{Meng,
  Z.}, \bibinfo{author}{Zhang, Z.}, \bibinfo{author}{Hu, Y.},
  \bibinfo{author}{Cheng, Y.}, \bibinfo{author}{Fan, C.},
  \bibinfo{author}{Wang, W.}, \bibinfo{author}{Liu, W.}, \bibinfo{author}{Wang,
  Z.}, et~al., \bibinfo{year}{2020}.
\newblock \bibinfo{title}{Efficient deep reinforcement learning via adaptive
  policy transfer}.
\newblock \bibinfo{journal}{arXiv preprint arXiv:2002.08037} .
%Type = Article
\bibitem[{Yao et~al.(2022)Yao, Bing, Zhuang, Chen, Zhou, Huang and
  Knoll}]{yao2022learning}
\bibinfo{author}{Yao, X.}, \bibinfo{author}{Bing, Z.}, \bibinfo{author}{Zhuang,
  G.}, \bibinfo{author}{Chen, K.}, \bibinfo{author}{Zhou, H.},
  \bibinfo{author}{Huang, K.}, \bibinfo{author}{Knoll, A.},
  \bibinfo{year}{2022}.
\newblock \bibinfo{title}{Learning from symmetry: Meta-reinforcement learning
  with symmetric data and language instructions}.
\newblock \bibinfo{journal}{arXiv preprint arXiv:2209.10656} .
%Type = Article
\bibitem[{You et~al.(2019)You, Lu, Filev and Tsiotras}]{YOU20191}
\bibinfo{author}{You, C.}, \bibinfo{author}{Lu, J.}, \bibinfo{author}{Filev,
  D.}, \bibinfo{author}{Tsiotras, P.}, \bibinfo{year}{2019}.
\newblock \bibinfo{title}{Advanced planning for autonomous vehicles using
  reinforcement learning and deep inverse reinforcement learning}.
\newblock \bibinfo{journal}{Rob. Auton. Syst.} \bibinfo{volume}{114},
  \bibinfo{pages}{1--18}.
%Type = Inproceedings
\bibitem[{Yu et~al.(2019)Yu, Song and Ermon}]{yu2019multi}
\bibinfo{author}{Yu, L.}, \bibinfo{author}{Song, J.}, \bibinfo{author}{Ermon,
  S.}, \bibinfo{year}{2019}.
\newblock \bibinfo{title}{Multi-agent adversarial inverse reinforcement
  learning}, in: \bibinfo{booktitle}{ICML}, \bibinfo{organization}{PMLR}. pp.
  \bibinfo{pages}{7194--7201}.
%Type = Article
\bibitem[{Zeng et~al.(2022)Zeng, Duan, Li, Ferrara, Pinto, Kuo and
  Nikolaidis}]{zeng2022human}
\bibinfo{author}{Zeng, Y.}, \bibinfo{author}{Duan, J.}, \bibinfo{author}{Li,
  Y.}, \bibinfo{author}{Ferrara, E.}, \bibinfo{author}{Pinto, L.},
  \bibinfo{author}{Kuo, C.C.J.}, \bibinfo{author}{Nikolaidis, S.},
  \bibinfo{year}{2022}.
\newblock \bibinfo{title}{Human decision makings on curriculum reinforcement
  learning with difficulty adjustment}.
\newblock \bibinfo{journal}{arXiv preprint arXiv:2208.02932} .
%Type = Article
\bibitem[{Zhang et~al.(2019)Zhang, Tai, Yun, Xiong, Liu, Boedecker and
  Burgard}]{zhang2019vr}
\bibinfo{author}{Zhang, J.}, \bibinfo{author}{Tai, L.}, \bibinfo{author}{Yun,
  P.}, \bibinfo{author}{Xiong, Y.}, \bibinfo{author}{Liu, M.},
  \bibinfo{author}{Boedecker, J.}, \bibinfo{author}{Burgard, W.},
  \bibinfo{year}{2019}.
\newblock \bibinfo{title}{{VR-GoGGLES} for robots: Real-to-sim domain
  adaptation for visual control}.
\newblock \bibinfo{journal}{IEEE Robot. Autom. Lett.} \bibinfo{volume}{4},
  \bibinfo{pages}{1148--1155}.
%Type = Article
\bibitem[{Zhang et~al.(2022)Zhang, Gao, Zhang, Guo, Ding, Wang, Sun and
  Zhao}]{9889241}
\bibinfo{author}{Zhang, Q.}, \bibinfo{author}{Gao, Y.}, \bibinfo{author}{Zhang,
  Y.}, \bibinfo{author}{Guo, Y.}, \bibinfo{author}{Ding, D.},
  \bibinfo{author}{Wang, Y.}, \bibinfo{author}{Sun, P.}, \bibinfo{author}{Zhao,
  D.}, \bibinfo{year}{2022}.
\newblock \bibinfo{title}{Trajgen: Generating realistic and diverse
  trajectories with reactive and feasible agent behaviors for autonomous
  driving}.
\newblock \bibinfo{journal}{IEEE Trans. Intell. Transp. Syst.}
  \bibinfo{volume}{23}, \bibinfo{pages}{24474--24487}.
%Type = Article
\bibitem[{Zhao and Tresp(2019)}]{zhao2019curiosity}
\bibinfo{author}{Zhao, R.}, \bibinfo{author}{Tresp, V.}, \bibinfo{year}{2019}.
\newblock \bibinfo{title}{Curiosity-driven experience prioritization via
  density estimation}.
\newblock \bibinfo{journal}{arXiv preprint arXiv:1902.08039} .
%Type = Article
\bibitem[{Zhao and Pajarinen(2022)}]{zhao2022self}
\bibinfo{author}{Zhao, W.}, \bibinfo{author}{Pajarinen, J.},
  \bibinfo{year}{2022}.
\newblock \bibinfo{title}{Self-paced multi-agent reinforcement learning}.
\newblock \bibinfo{journal}{arXiv preprint arXiv:2205.10016} .
%Type = Inproceedings
\bibitem[{Zhao et~al.(2020)Zhao, Queralta and Westerlund}]{zhao2020sim}
\bibinfo{author}{Zhao, W.}, \bibinfo{author}{Queralta, J.P.},
  \bibinfo{author}{Westerlund, T.}, \bibinfo{year}{2020}.
\newblock \bibinfo{title}{Sim-to-real transfer in deep reinforcement learning
  for robotics: a survey}, in: \bibinfo{booktitle}{IEEE Symp. Series Comput.
  Intell.}, \bibinfo{organization}{IEEE}. pp. \bibinfo{pages}{737--744}.
%Type = Article
\bibitem[{Zhou et~al.(2022)Zhou, Wang, Du and Li}]{zhou2022clustering}
\bibinfo{author}{Zhou, P.}, \bibinfo{author}{Wang, X.}, \bibinfo{author}{Du,
  L.}, \bibinfo{author}{Li, X.}, \bibinfo{year}{2022}.
\newblock \bibinfo{title}{Clustering ensemble via structured hypergraph
  learning}.
\newblock \bibinfo{journal}{Inf. Fusion} \bibinfo{volume}{78},
  \bibinfo{pages}{171--179}.
%Type = Article
\bibitem[{Zhu et~al.(2022)Zhu, Wu, Li, Lv and Xu}]{zhu2022context}
\bibinfo{author}{Zhu, R.}, \bibinfo{author}{Wu, S.}, \bibinfo{author}{Li, L.},
  \bibinfo{author}{Lv, P.}, \bibinfo{author}{Xu, M.}, \bibinfo{year}{2022}.
\newblock \bibinfo{title}{Context-aware multi-agent broad reinforcement
  learning for mixed pedestrian-vehicle adaptive traffic light control}.
\newblock \bibinfo{journal}{IEEE Internet Things J.} .
%Type = Article
\bibitem[{Zhu et~al.(2020)Zhu, Lin and Zhou}]{zhu2020transfer}
\bibinfo{author}{Zhu, Z.}, \bibinfo{author}{Lin, K.}, \bibinfo{author}{Zhou,
  J.}, \bibinfo{year}{2020}.
\newblock \bibinfo{title}{Transfer learning in deep reinforcement learning: A
  survey}.
\newblock \bibinfo{journal}{arXiv preprint arXiv:2009.07888} .

\end{thebibliography}

\end{document}